\documentclass[twoside]{article}

\usepackage{tex_macros}
\usepackage{subcaption}
\usepackage{bm}
\usepackage{framed}
\usepackage{physics}
\usepackage{wrapfig}
\usepackage{enumitem}
\usepackage{xfrac}
\theoremstyle{plain}
\newtheorem{theorem}{Theorem}[section]

\newtheorem{lemma}[theorem]{Lemma}

\theoremstyle{definition}

\newtheorem{assumption}[theorem]{Assumption}
\theoremstyle{remark}
\newtheorem{remark}[theorem]{Remark}

\usepackage[accepted]{aistats2026}
\usepackage[round]{natbib}

\PassOptionsToPackage{numbers,compress}{natbib}

\begin{document}

\runningtitle{DSM with Random Features:
Insights on Diffusion Models from
Precise Learning Curves}


\twocolumn[

\aistatstitle{Denoising Score Matching with Random Features:
Insights on Diffusion Models From
Precise Learning Curves}

\aistatsauthor{ Anand Jerry George \And Rodrigo Veiga \And  Nicolas Macris }

\aistatsaddress{ École Polytechnique Fédérale de Lausanne (EPFL),\\
Lab for Statistical Mechanics of Inference in Large Systems (SMILS),\\ CH-1015 Lausanne,\\ Switzerland} 
% \aistatsaddress{ SMILS, EPFL \And  SMILS, EPFL \And SMILS, EPFL } 
]
\begin{abstract}
We theoretically investigate the phenomena of generalization and memorization in diffusion models. Empirical studies suggest that these phenomena are influenced by model complexity and the size of the training dataset. In our experiments, we further observe that the number of noise samples per data sample ($m$) used during Denoising Score Matching (DSM) plays a significant and non-trivial role. We capture these behaviors and shed insights into their mechanisms by deriving asymptotically precise expressions for test and train errors of DSM under a simple theoretical setting. The score function is parameterized by random features neural networks, with the target distribution being $d$-dimensional Gaussian. We operate in a regime where the dimension $d$, number of data samples $n$, and number of features $p$ tend to infinity while keeping the ratios $\psi_n=\frac{n}{d}$ and $\psi_p=\frac{p}{d}$ fixed. By characterizing the test and train errors, we identify regimes of generalization and memorization as a function of $\psi_n,\psi_p$, and $m$. Our theoretical findings are consistent with the empirical observations.
\end{abstract}

\section{INTRODUCTION}
Generative models are at the heart of the ongoing revolution in artificial intelligence. Formally, they aim to generate new samples from an unknown probability distribution, given
$n$ i.i.d. samples drawn from it.
Commercial generative models demonstrate remarkable capabilities across various modalities, including text, speech, images, and videos, with new advancements being reported regularly. Their impressive capabilities are mainly driven by two key architectures: \textit{transformers} and \textit{diffusion models}. While transformers excel primarily on text data, diffusion models show exceptional proficiency in generating natural-looking images from text prompts. Despite their success, the scientific community remains divided on whether these models are truly creative or merely imitate based on the examples that they have seen during training~\citep{ukpaka_creative_2024,kamb_analytic_2024}. An even more pressing concern is that of \textit{memorization}, where the model's response partially or fully resembles training data. The memorization phenomenon raises serious implications, particularly regarding the privacy of data for training~\citep{carlini_extracting_2021, carlini_extracting_2023}. The limited theoretical understanding of these models prevents addressing such questions effectively.

In this study, we focus on the generalization and memorization behaviors of diffusion models. Our work is motivated by empirical observations of the factors affecting these properties in practical scenarios. Several prior works investigated the impact of model complexity and size of the training dataset on these behaviors \citep{somepalli_diffusion_2023,somepalli_understanding_2024,carlini_extracting_2023,zhang_emergence_2024,yoon_diffusion_2023,gu_memorization_2023,ross_geometric_2024,pham_memorization_2024}. In addition to this, our experiments also reveal that the number of noise samples per data sample ($m$) used in Denoising Score Matching (DSM) is another key factor impacting generalization and memorization. Details and results of our experiments with real and synthetic datasets can be found in Section~\ref{sec:numerical_exps} and in Appendix~\ref{appndx:real_data_exp}. In a nutshell, the experiments suggest the following: 1) Memorization increases as the size of the training dataset decreases relative to model complexity (equivalently, it increases as model complexity increases relative to the number of data points), 2) Memorization increases as the number of noise samples per data sample increases in the overparametrized regime. Despite these observations, a theoretical study of memorization aspects of diffusion models has been done only with the \textit{empirical optimal score} function \citep{biroli_dynamical_2024, achilli_losing_2024,achilli_memorization_2025, george_analysis_2025} (see Section~\ref{sec:empirical_opt_score}). Moreover, these analyses look at a regime in which the size of the training dataset scales exponentially with dimension, and do not shed much light on other regimes, e.g., a proportional one. This gap in theory and practice raises an important question: can the phenomenon of memorization be theoretically shown when using a parametric class of functions for the score function? This is precisely the context of our study. By analyzing the learning process of a specific diffusion model instance, we provide insights into generalization and memorization in diffusion models.

\subsection{Main Contributions}\label{sec:main_contributions}
This section provides a brief overview of our key contributions and findings. Our study focuses on the learning aspect of diffusion models, i.e., learning the score function of perturbed versions of a target distribution $P_0$ (see Eq.~\eqref{eqn:marginal_prob_forward}). The task is to minimize a loss function known as \textit{denoising score matching} (DSM) objective (see Eq.~\eqref{eqn:dsm_mfinite}) over a parametric class of functions, for which we consider random features neural networks (RFNNs). The target distribution is the $d$-dimensional Gaussian distribution, with $n$ denoting the number of samples and $p$ the number of features of the RFNN. We operate in a regime where $d,n,p\rightarrow\infty$, while the ratios $\psi_n = \frac{n}{d}$ and $\psi_p=\frac{p}{d}$ are kept fixed. The DSM objective involves an additional parameter $m$: the number of noise samples per data sample used in forming the loss function, Eq.~\eqref{eqn:dsm_mfinite}.   
% 1) We analytically compute the  test and train errors of the minimizer of denoising score matching loss.
% 2) Using the obtained test and train errors, we study the generalization and memorization behavior in diffusion models.
% 3) We show that a crossover transition between generalization and memorization behaviors occur when the number of features equals the number of samples. .
% 4) We demonstration that increasing the value of $m$ enhances generalization when $p<n$, while it intensifies memorization when $p>n$.
\begin{figure}[ht]
    \centering    \includegraphics[width=0.65\linewidth]{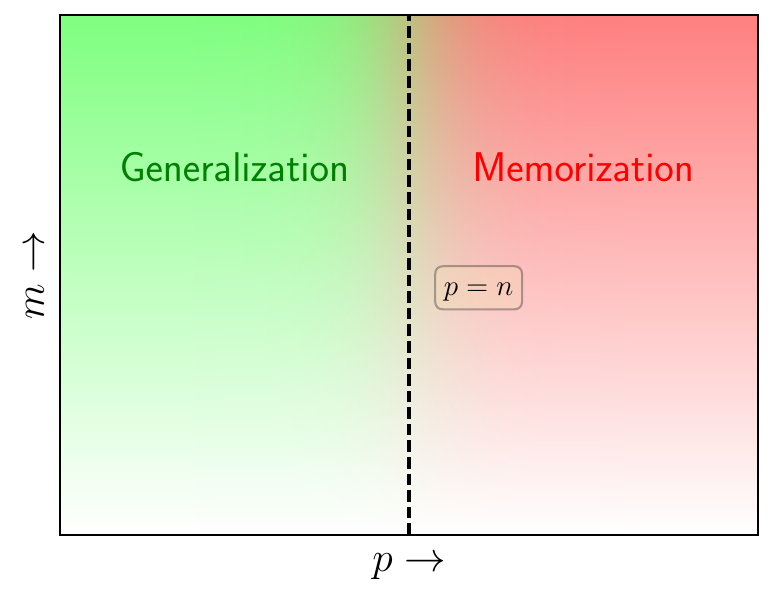}
    \caption{Phase diagram showing regimes of generalization and memorization. The gradient in color with $m$ indicates the change in strength of the phenomenon.}
    \label{fig:phase_diagram}
    %\vspace{-40pt}
\end{figure}
In this theoretical setting, we make the following contributions, illustrated on the schematic phase diagram in Fig.~\ref{fig:phase_diagram}:
\begin{itemize}[leftmargin=*]
    \item We analytically compute the test and train errors of the minimizer of the DSM loss.
    \item Using the obtained test and train errors, we study the generalization and memorization behavior in diffusion models.
    %\item We show that a crossover transition between generalization and memorization behaviors occurs when the number of features equals the number of samples. %This is in contrast to the $\bigO{e^d}$ samples required for generalization when the empirical optimal score function is used.
    \item We show that a crossover transition between generalization and memorization occurs when $p=n$. 
    \item We demonstrate that increasing the value of $m$ enhances generalization when $p<n$, while it intensifies memorization when $p>n$.
\end{itemize}
%We consider the simple isotropic Gaussian as the target distribution for two primary reasons. First, when studying the memorization phenomenon, the exact form of the distribution is less critical, as the focus lies on the specific samples used during training. Second, choosing a Gaussian target distribution provides an analytically tractable setting, allowing for a more precise theoretical analysis.

The choice of a Gaussian target distribution $P_0$ provides an analytically tractable setting as long as we assume some simple structure for the covariance. It allows a precise theoretical analysis and captures phenomena observed in practice. We fully solve the case where $P_0$ is isotropic on a $D$-dimensional linear subspace, with $\psi_D = \frac{D}{d}\le 1$ held fixed as $D,d\to\infty$. 
\subsection{Related Works}
\paragraph{Diffusion models} Diffusion models \citep{sohl-dickstein_deep_2015,song_generative_2019,ho_denoising_2020,song_score-based_2020} rely on the non-equilibrium dynamics of a diffusion process for generative modeling. Since their introduction, there were several improvements \citep{dhariwal_diffusion_2021,rombach_high-resolution_2022,ho_classifier-free_2021,nichol_glide_2022} that led them to become the state-of-the-art in generative modeling of images.  Further, the design aspects of diffusion models were studied in \cite{karras_elucidating_2022}.
\paragraph{Sampling accuracy, generalization, and memorization} Several works have investigated the theoretical aspects of diffusion models. Sampling accuracy of the generative process in terms of distance from the target distribution was derived in \cite{chen_sampling_2022,benton_nearly_2023,chen_improved_2023,bortoli_convergence_2022} for various settings. These works assume that the score function has been learned \emph{a priori} with a certain level of accuracy. The score learning process was studied in \cite{cui_analysis_2023,shah_learning_2023,han_neural_2023,zeno_when_2025}. The works~\cite{kadkhodaie_generalization_2023,chen_score_2023,li_generalization_2023,wang_evaluating_2024,cui_precise_2025} performed an end-to-end study of diffusion models giving a better understanding of their generalization properties. Specifically,~\cite{li_generalization_2023} and \cite{saha_generalization_2025} considers the score function parametrized by RFNNs and provides bounds on the KL divergence between the learned and target distributions. Further, the critical nature of feature emergence in diffusion models was studied in \cite{li_critical_2024,sclocchi_phase_2025,favero_how_2025}. The manifolds learned by diffusion models were theoretically studied in \cite{pidstrigach_score-based_2022}.
Statistical physics tools were employed to study memorization in high-dimensions using the empirical optimal score function in~\cite{biroli_dynamical_2024,raya_spontaneous_2023,ambrogioni_search_2024,achilli_losing_2024,achilli_memorization_2025}. A geometric framework to understand memorization was proposed in \cite{ross_geometric_2024}. More recently, \cite{bonnaire_why_2025} investigated how early stopping in training helps mitigate memorization. In particular, their theoretical analysis uses the same setting as ours (see Appendix~\ref{appendix:disc_biroli} for further discussion).  
%\paragraph{Sampling accuracy, generalization, and memorization} Several works have investigated the theoretical aspects of diffusion models. Sampling accuracy of the generative process in diffusion models in terms of distance from the target distribution was derived in \cite{chen_sampling_2022,benton_nearly_2023,chen_improved_2023,bortoli_convergence_2022} for various settings. These works assume that the score function has been learned a \emph{priori} with a certain level of accuracy. The score learning process was studied in \cite{cui_analysis_2024,shah_learning_2023,han_neural_2023}. The works \cite{kadkhodaie_generalization_2024,chen_score_2023,wang_evaluating_2024} performed an end-to-end study of diffusion models giving a better understanding of their generalization properties.
%More recently, statistical physics tools were employed to study the memorization phenomenon in diffusion models that use empirical optimal score function, in the large data regime \cite{biroli_dynamical_2024,raya_spontaneous_2023,ambrogioni_search_2023,achilli_losing_2024}. 

\subsection{Our Techniques}
In the setting briefly described in Section~\ref{sec:main_contributions} and to be detailed in Section~\ref{sec:main_results}, we analytically compute the test and train errors of the minimizer of the DSM loss. Our theoretical results are summarized in Theorems~\ref{thm:lc_minf} and~\ref{thm:lc_m1}. Two main ingredients allowing the computation of exact asymptotic learning curves are the \textit{Gaussian equivalence principle} (GEP) \citep{gerace_generalisation_2020,goldt_gaussian_2022,hu_universality_2023} and the theory of \textit{linear pencils} \citep{far_spectra_2006,helton_applications_2018,adlam_neural_2020,bodin_random_2024}. We briefly describe them here.
\paragraph{Gaussian Equivalence Principle}
Suppose $W\in\R^{p\times d}$, and $X\in\R^{d\times n}$ are random matrices with i.i.d. Gaussian entries. Let $\varrho$ be a non-linear activation function and let $F=\varrho\left(\rat{W}{d}X\right)$, where $\varrho$ acts element-wise on a matrix. The GEP states that in the calculation of test and train errors, it is asymptotically equivalent to replace $F$ by the matrix $\hat{F} = \mu_0\bone_p\bone_n^T+\mu_1\rat{W}{d}X+v\Omega$, where $\Omega\in\R^{p\times n}$ is a random matrix with i.i.d. Gaussian entries, independent of $W, X$. The coefficients are given by $\mu_0 = \bE{g}{\varrho(g)},\; \mu_1 = \bE{g}{\varrho(g)g},\; v^2 = \bE{g}{\varrho(g)^2}-\mu_0^2-\mu_1^2$, where $g\sim\cN{0,1}$ \;.
%\paragraph{Linear Pencils} The theory of linear pencils is a powerful technique that allows us  to compute traces of rational functions involving random matrices with Gaussian entries. It amounts to constructing an appropriate block matrix called \textit{linear pencil}, whose inverse gives the desired rational functions. For further details on the linear pencils method, we refer the reader to Chapter~3 of \cite{bodin_random_2024} .
\paragraph{Linear Pencils} The theory of linear pencils is a powerful technique to compute traces of rational functions involving random matrices with Gaussian entries. It entails constructing an appropriate block matrix called \textit{linear pencil}, whose inverse gives the desired rational functions. We refer the reader to Chapter~3 of \cite{bodin_random_2024} for further details on the method.

%The theory of linear pencils is a powerful technique allowing to compute traces of rational functions involving random matrices with Gaussian entries. It amounts to constructing an appropriate block matrix called \textit{linear pencil}, whose inverse gives the desired rational functions. For further details on the linear pencils method, we refer the reader to Chapter~3 of \cite{bodin_random_2024} .

%A short exposition of the linear pencils method can be found in Appendix~\ref{appndx:linear_pencils}.

 %In the present context, the test and train errors can be expressed as a sum of traces of rational functions of random matrices. However, some of these random matrices have non-Gaussian entries due to the presence of an activation function $\varrho$. We circumvent this using the Gaussian equivalence principle. Subsequently, we use the linear pencils method to compute the traces.

In our context, the test and train errors can be expressed as a sum of traces of rational functions of random matrices. Some of them have non-Gaussian entries due to $\varrho$. We circumvent this by using the GEP and, subsequently, the linear pencils method to compute the traces.
 
 %albeit non-linear due to the presence of an activation function $\varrho$. We linearize the terms using Gaussian equivalence principle. Subsequently, we use the linear pencils method to compute the traces.
% \paragraph{Notations}
% We denote the $d$-dimensional Gaussian distribution with mean $\mu$ and covariance $\Sigma$ by $\cN{\mu,\Sigma}$. The $d$-dimensional identity matrix is denoted by $I_d$. $\norm{\cdot}$ denotes the $l_2$ norm of a vector, while $\norm{\cdot}_F$ denotes the Frobenius norm of a matrix. The operator $\nabla$ represents the gradient of a scalar function. 
\paragraph{Notations}
The $d$-dimensional Gaussian distribution with mean $\mu$ and covariance $\Sigma$ is denoted by $\cN{\mu,\Sigma}$. The $d$-dimensional identity matrix is denoted by $I_d$, and $\bone_d$ stands for the $d$-dimensional all-ones vector. $\norm{\cdot}$ denotes the $l_2$ norm of a vector, while $\norm{\cdot}_F$ denotes the Frobenius norm of a matrix. The operator $\nabla$ represents the gradient of a scalar function. 
\section{PRELIMINARIES}

\subsection{Diffusion Models}
Consider a set of $n$ i.i.d. samples $\cS=\{x_1,x_2,\cdots,x_n\}$ from an unknown distribution $P_0$ on $\R^d$. Diffusion models address the problem of leveraging the information in $\cS$ to draw new samples from $P_0$ by time reversing a stochastic process transporting $P_0$ to a known distribution, such as the isotropic Gaussian. Letting this forward diffusion be an Ornstein-Uhlenbeck (OU) process~\citep{gardiner_stochastic_2009}, we have:
%Consider a set of $n$ i.i.d. samples $\cS=\{x_1,x_2,\cdots,x_n\}$ from an unknown distribution $P_0$ on $\R^d$. Generative modeling aims to leverage the information in $\cS$ to draw new samples from $P_0$. Diffusion models address the problem by time reversing a diffusion process that transports $P_0$ to a known distribution such as the isotropic Gaussian. 
%In this work, we let the forward process to be an Ornstein-Uhlenbeck (OU) process~\cite{gardiner_stochastic_2009} governed by the following stochastic differential equation (SDE):
\begin{equation}\label{eqn:forward_ou}
    \dd X_t = - X_t \;  \dd t +\sqrt{2} \; \dd B_t\;,\quad X_0\sim P_0\;,
\end{equation}
where $B_t$ is a standard Brownian motion in $\R^d$. The distribution of $X_t$ given $X_0$ can be computed in closed form from Eq.~\eqref{eqn:forward_ou} and is simply $\cN{a_tX_0,h_tI_d}$, where $a_t=e^{-t}$ and $h_t=1-e^{-2t}$ (hence, $a_t^2+h_t=1$). As $t\to\infty$, the distribution of $X_t$ converges to the $d$-dimensional standard Gaussian, regardless of $X_0$, since $a_t\to 0$ and $h_t\to 1$. 
% The probability distribution of $X_t$ is given by $P_t(x) = {(2\pi h_t)^{-d/2}}\int_{\R^d}  \dd P_0(x_0) \; e^{-\frac{\norm{x-a_tx_0}^2}{2h_t}}$.
Let $P_t$ denote the probability distribution of $X_t$:
\begin{equation}\label{eqn:marginal_prob_forward}
    P_t(x) = {(2\pi h_t)^{-d/2}}\int_{\R^d}  \dd P_0(x_0) \; e^{-\frac{\norm{x-a_tx_0}^2}{2h_t}}\;.
\end{equation}
Then, for a fixed $T>0$ and $Y_T\sim P_T$, we define the \textit{time reversal} of the forward process (\ref{eqn:forward_ou}) as 
\begin{equation}\label{eqn:backward_sde}
    -\dd Y_t = \left(Y_t+2\nabla\log P_{t}(Y_t)\right) \;  \dd t+\sqrt{2} \; \dd \tilde{B}_t\;,
\end{equation}
where the process runs backward in time starting from $Y_T$, and $\tilde{B}_t$ is a different instance of standard Brownian motion. 
The term \emph{time reversal} here means that the distributions of $Y_t$ and $X_t$ are identical for every $t$ \citep{anderson_reverse-time_1982, haussmann_time_1986}.
% When we say that $Y_t$ is a time reversal \cite{anderson_reverse-time_1982, haussmann_time_1986} of $X_t$, we mean that their distributions are identical for every $t$. 
%This can be verified by writing the Fokker-Planck equations satisfied by the time marginals of processes $X_t$ and $Y_t$. 
If the backward process is initiated with $Y_T\sim P_T$, the distribution of $Y_0$ will be $P_0$. However, since $P_T$ is unknown due to the lack of knowledge of $P_0$, we instead start the reverse process with $Y_T\sim\cN{0,I_d}$. This approximation incurs minimal error due to the OU process’s exponential convergence to $\cN{0,I_d}$. 

The implementation of the backward process, Eq.~\eqref{eqn:backward_sde}, requires the so-called \textit{score function} $\nabla\log P_t$, which is unknown as $P_t$ is unknown. The learning task is to estimate the score function using the dataset $\cS$. We consider minimizing the following score matching \citep{hyvarinen_estimation_2005} objective for this: $\cL_{\text{SM}}(s) = \int_0^T  \dd t \; \bE{x_t\sim P_t}{\norm{s(t,x_t)-\nabla\log P_t(x_t)}^2}  \;.
$
% \begin{equation*}
%     \cL_{\text{SM}}(s) = \int_0^T  \dd t \; \bE{x_t\sim P_t}{\norm{s(t,x_t)-\nabla\log P_t(x_t)}^2}  \;.
% \end{equation*}
The $\cL_{\text{SM}}$ loss function is not practical, as $\nabla\log P_t(x)$ is unknown. However, it is possible to construct an equivalent objective, the denoising score matching (DSM) loss \citep{vincent_connection_2011}: $\cL_{\text{DSM}}(s) = \int_0^T  \dd t \; w(t)\shortexpect\norm{s(t,x_t)-\nabla\log P_t(x_t|x_0)}^2 ,$
% \begin{align*}
%     \cL_{\text{DSM}}(s) &= \int_0^T  \dd t \; w(t)\shortexpect\norm{s(t,x_t)-\nabla\log P_t(x_t|x_0)}^2 ,
% \end{align*}
where $w$ is a weighting function and the expectation is with respect to $x_0$ and $x_t$. 
%Here, by an abuse of notation, $P_t$ also denotes the joint {\color{blue} conditional ?} probability distribution of the process $X_t$.
In Appendix~\ref{appndx:dsm_sm_equiv}, we show that $\cL_{\text{DSM}}$ is equal to $\cL_{\text{SM}}$ up to a time-dependent scaling factor and offset.
%Hence, the minimizer of $\cL_{\text{DSM}}$ is same as the minimizer of $\cL_{\text{SM}}$.
Following \cite{song_score-based_2020}, we choose $w(t) = (\shortexpect_{x_0,x_t}{\norm{\nabla\log P_t(x_t|x_0)}^2})^{-1}$.
For OU process, we can compute $\nabla\log P_t(x_t|x_0)$ in closed form. We can write $x_t\sim P_t$ as  $x_t = a_t x_0+\sqrt{h_t}z$, where $x_0\sim P_0,\; z\sim\cN{0,I_d}$ are independent rvs and $a_t=e^{-t}$,\; $h_t=1-e^{-2t}$. Consequently, $\nabla\log P_t(x_t|x_0) = -\frac{(x_t-a_tx_0)}{h_t}=-\frac{z}{\sqrt{h_t}}$. The weight function is given by $w(t)=\frac{h_t}{d}$. Substituting these, we can write $\cL_{\text{DSM}}(s) = \int_0^T  \dd t  \frac{1}{d}\shortexpect\norm{\sqrt{h_t}s(t,a_tx_0+\sqrt{h_t}z)+z}^2,$
% \begin{align*}
%     \cL_{\text{DSM}}(s) &= \int_0^T  \dd t  \frac{1}{d}\shortexpect\norm{\sqrt{h_t}s(t,a_tx_0+\sqrt{h_t}z)+z}^2,
% \end{align*}
where the expectation is with respect to $x_0$ and $z$. Since $P_0$ is unknown and only samples from it are available, we use an empirical estimate for the expectation with respect to $x_0$. Setting $ y_i^t (z) = a_tx_i+\sqrt{h_t}z$, we define
\begin{align}\label{eqn:dsm_minf}
    \cL_{\text{DSM}}^\infty(s) &= \int_0^T \dd t \; \frac{1}{dn}\sum_{i=1}^{n}{\shortexpect_{z}{\norm{\sqrt{h_t}s(t, y_i^t (z))+z}^2}} .
\end{align}
%We define
%\begin{align}\label{eqn:dsm_minf}
%    \cL_{\text{DSM}}^\infty(s) &= \int_0^T \dd t \; \frac{1}{dn}\sum_{i=1}^{n}{\shortexpect_{z}{\norm{\sqrt{h_t}s(t,a_tx_i+\sqrt{h_t}z)+z}^2}} .
%\end{align}
When $s$ is a complicated function such as a neural network, which is the typical practical setting, computing the expectation over $z$ might be intractable. In this case, one consideres $m$ samples for $z$ and uses an additional empirical estimate, obtaining the following loss function:
%\begin{align}\label{eqn:dsm_mfinite}
%    \cL_{\text{DSM}}^m(s) &= \int_0^T \dd t \; \frac{1}{dnm}{\sum_{i,j=1}^{n,m}\norm{\sqrt{h_t}s(t,a_tx_i+\sqrt{h_t}z^t_{ij})+z^t_{ij}}^2}  \;.
%\end{align}
\begin{equation}\label{eqn:dsm_mfinite}
    \cL_{\text{DSM}}^m(s) = \int_0^T \dd t \; \frac{1}{dnm}{\sum_{i,j=1}^{n,m}\norm{\sqrt{h_t}s(t,y_{ij}^t)+z^t_{ij}}^2} 
\end{equation}
where $ y_{ij}^t (z) = a_t x_i+\sqrt{h_t} z_{ij}^t$.
\subsection{Empirical Optimal Score and Memorization}\label{sec:empirical_opt_score}
%Consider the loss function given in (\ref{eqn:dsm_minf}). There is a unique minimizer for this loss function, and it is given by 
Consider the loss function given in (\ref{eqn:dsm_minf}). It has an unique minimizer: $s^e(t,x) = \nabla\log P^e_t(x) $, 
%\begin{equation}\label{eqn:empirical_opt_score}
%    s^e(t,x) = \nabla\log P^e_t(x) \;,
%\end{equation}
with
\begin{equation}\label{eqn:marginal_prob_empirical}
    P_t^e(x) = \frac{1}{n(2\pi h_t)^{d/2}}\sum_{i=1}^ne^{-\frac{\norm{x-a_tx_i}^2}{2h_t}} \;.
\end{equation}
The score $s^e$ is often referred to as the \textit{empirical optimal score}. A backward process using $s^e$ converges in distribution to the empirical distribution of the dataset $\cS$ as $t\rightarrow 0$. That is, the backward process collapses to one of the data samples as $t\rightarrow 0$. This is related to the memorization phenomenon as studied in \cite{biroli_dynamical_2024}, although their study focuses on the regime where $n=\bigO{e^{\Theta(d)}}$. 

%\textit{Inspired by this connection, we define memorization in diffusion models as occurring when the score function learned using denoising score matching closely approximates the empirical optimal score function instead of the exact score.} 
\textit{Inspired by this connection, we define memorization as occurring when the score function learned using DSM closely approximates the empirical optimal score function instead of the exact score.}

\subsection{Random Features Model}
In practice, the score function $s$ is typically chosen from a parametric class of functions, and the DSM objective (\ref{eqn:dsm_mfinite}) is minimized within this class, with appropriate regularization. In this work, we represent the score function using a \textit{random features} neural network (RFNN) \citep{rahimi_random_2007}. A RFNN is a two-layer neural network in which the first layer weights are randomly chosen and fixed, while the second layer weights are learned during training. It is a function from $\R^d$ to $\R^d$ of the form $s_A(x|W) = \rat{A}{p}\act{\rat{W}{d}x},$ where $W\in\R^{p\times d}$ is a random matrix with its elements chosen i.i.d. from $\cN{0,1}$, $\varrho$ is an activation function acting element-wise and $A\in\R^{d\times p}$ are the second layer weights that need to be learned. The RFNN is a simple neural network amenable to theoretical analysis. It is able to capture attributes  frequently observed in more complex models, such as the double descent curve related to overparametrized regimes~\citep{mei_generalization_2022,bodin_model_2021}.
%In practice, the score function $s$ is typically selected from a parametric class of functions, and the DSM objective (\ref{eqn:dsm_mfinite}) is minimized within this class, with appropriate regularization. We choose a \textit{random features} neural network (RFNN) \cite{rahimi_random_2007}, which is a two-layer neural network in which the first layer weights are randomly chosen and fixed, while the second layer weights are learned during training. Formally, it is a function from $\R^d$ to $\R^d$ of the form $s_A(x|W) = \rat{A}{p}\act{\rat{W}{d}x},$, where $W\in\R^{p\times d}$ is a random matrix with its elements chosen i.i.d. from $\cN{0,1}$, $\varrho$ is an activation function acting element-wise and $A\in\R^{d\times p}$ are the second layer weights that need to be learned. The RFNN is a simple neural network amenable to theoretical analysis and is able to capture interesting characteristics observed in more complicated neural network models, such as the double descent curve related to overparametrized regimes~\cite{mei_generalization_2022,bodin_model_2021}.
\section{MAIN RESULTS}\label{sec:main_results}
We study the DSM loss $\cL_{\text{DSM}}^m$ given in (\ref{eqn:dsm_mfinite}) when the score function is modeled using a RFNN and $P_0\equiv\cN{0,\cC}$ with $\frac{1}{d}{\rm tr}\,\mathcal{C}$ tending to a finite limit as $d\to \infty$. % is a $d$-dimensional standard Gaussian distribution. 
Our results characterize the asymptotic learning curves (test and train errors) of the minimizer of $\cL_{\text{DSM}}^m$ (\ref{eqn:dsm_mfinite}) in this setting. We obtain closed form expressions for learning curves 
when $\cC$ has certain structure. Specifically, we assume that $\cC = MM^T$, where $M\in\R^{d\times D}$ has orthonormal columns. This corresponds to the situation where data lies in a $D$-dimensional subspace of $\R^d$, inspired by the well known manifold hypothesis. We consider two values of $m$: $m=1$ and $m=\infty$, which correspond to the extremes of the number of noise samples per data sample used during score learning. Based on the derived learning curves, we discuss the generalization and memorization behaviors in diffusion models. For intermediate values of $m$, we obtain the learning curves numerically in Appendix~\ref{appndx:lc_numerical_comparison}.%, a phenomenon observed in several previous works (see, e.g., \cite{yoon_diffusion_2023}).

We assume that at each time instant $t$, a different instance of RFNN is used to learn the score function specific to that time. Although this is a simplification compared to the methods employed in practice, it has been adopted in prior theoretical studies, e.g.,~\cite{cui_analysis_2023}. When using a different instance of RFNN at each $t$, note that minimizing the DSM loss (\ref{eqn:dsm_mfinite}) is equivalent to minimizing the integrand therein at each time instant. Henceforth, we focus on minimizing the DSM loss for a fixed $t$. Introducing a regularization parameter $\lambda>0$, the loss function becomes:
\begin{align}\label{eqn:dsm_loss_rfm}
    \cL_t^m(A_t) =& \frac{1}{dnm}\sum_{i,j=1}^{n,m}{\norm{\sqrt{h_t}\rat{A_t}{p}\act{\rat{W_t}{d}y^t_{ij}}+z^t_{ij}}^2} \nonumber\\
&+ \frac{h_t\lambda}{dp}\norm{A_t}_F^2 \;,
\end{align}
%\begin{equation}\label{eqn:dsm_loss_rfm}
%     \cL_t^m(A_t) = \frac{1}{dnm}\sum_{i,j=1}^{n,m}{\norm{\sqrt{h_t}\rat{A_t}{p}\act{\rat{W_t}{d}(a_tx_i+\sqrt{h_t}z^t_{ij})}+z^t_{ij}}^2} +\frac{h_t\lambda}{dp}\norm{A_t}_F^2 \;,
%\end{equation}
with $z^t_{ij}\sim\cN{0,I_d}$. We emphasize that $W_t$ is a different and independent random matrix for each $t$, and $A_t$ is learned separately at each $t$.
%Denote $\hat{A}_t$ as the second layer weights learned by some learning algorithm minimizing (\ref{eqn:dsm_loss_rfm}). The respective performance is evaluated by the test and train errors defined as follows:
Denote $\hat{A}_t$ as the output of some algorithm minimizing~\eqref{eqn:dsm_loss_rfm}. The respective performance is evaluated by the test and train errors defined as follows:
%\begin{align}
%    \cE^m_{\text{test}}(\hat{A}_t) &= \frac{1}{d}\shortexpect_{x\sim P_t}{\norm{\rat{\hat{A}_t}{p}\act{\rat{W_t}{d}x}-\nabla\log P_t(x)}^2} \label{eq:At_test_inf}\;,\\
%    \cE^m_{\text{train}}(\hat{A}_t) &= \frac{1}{dnm}\sum_{i=1}^{n}{\sum_{j=1}^{m}\norm{\sqrt{h_t}\rat{\hat{A}_t}{p}\act{\rat{W_t}{d}(a_tx_i+\sqrt{h_t}z^t_{ij})}+z^t_{ij}}^2} \;. \label{eq:At_train_inf}
%\end{align}
\begin{align}
    \cE^m_{\text{test}}(\hat{A}_t) &= \frac{1}{d}\shortexpect_{x\sim P_t}{\norm{\rat{\hat{A}_t}{p}\act{\rat{W_t}{d}x}-\nabla\log P_t(x)}^2} \label{eq:At_test_inf}\;,\\
    \cE^m_{\text{train}}(\hat{A}_t) &= \frac{1}{dnm}\sum_{i,j=1}^{n,m}{\norm{\sqrt{h_t}\rat{\hat{A}_t}{p}\act{\rat{W_t}{d}y_{ij}^t}+z^t_{ij}}^2}.\label{eq:At_train_inf}
\end{align}
The quantities $\cE^m_{\text{test}}$ and $\cE^m_{\text{train}}$ are random, due to $W_t$, $\{x_i\}_{i=1}^n$, and $\{z^t_{ij}\}_{i,j=1}^{n,m}$. We expect them to concentrate on their expectations as $d\to\infty$.  

The rationale behind studying the test and training errors of DSM to assess the performance of diffusion models stems from their direct connection to the model's generative accuracy. Specifically, as described in \cite{song_maximum_2021}, the error in diffusion models can be upper bounded using the test error. Suppose $\hat{P}_t$ denotes the probability distribution of the backward process when we use the learned score function instead of the true score, and assume that $\hat{P}_T = P_T$. Then, the Kullback-Leibler (KL) divergence between $P_0$ and $\hat{P}_0$ can be upper bounded as
    $D_{KL}(P_0||\hat{P}_0)\le \frac{d}{2}\int_0^T\dd t\; \cE^m_{\text{test}}(\hat{A}_t).$

\subsection{Infinite Noise Samples per Data Sample}
For $m=\infty$, the average over $z^t_{ij}$ in the DSM loss becomes an expectation. We write the loss as
\begin{align}\label{eqn:dsm_loss_rfm_minf}
    \cL_t^\infty(A_t) &= \frac{1}{dn}\sum_{i=1}^{n}{\shortexpect_{z}{\norm{\sqrt{h_t}\rat{A_t}{p}\act{\rat{W_t}{d}y^t_i(z)}+z}^2}}\nonumber\\
    &\qquad +\frac{h_t\lambda}{dp}\norm{A_t}_F^2 \;,
\end{align}
%\begin{equation}\label{eqn:dsm_loss_rfm_minf}
%    \cL_t^\infty(A_t) = \frac{1}{dn}\sum_{i=1}^{n}{\shortexpect_{z}{\norm{\sqrt{h_t}\rat{A_t}{p}\act{\rat{W_t}{d}(a_tx_i+\sqrt{h_t}z)}+z}^2}} +\frac{h_t\lambda}{dp}\norm{A_t}_F^2\;, 
%\end{equation}
and characterize the minimizer in the asymptotic regime $d,n,p\rightarrow\infty$, with $\psi_n = \frac{n}{d}$ and $\psi_p = \frac{p}{d}$ fixed. We make the following assumption on $\varrho$.

\begin{assumption}\label{assmptn:activation_fn}
   The activation function $\varrho$ has a Hermite polynomial expansion given by $\varrho(y) = \sum_{l=0}^\infty\mu_l \text{He}_l(y)$, where $\text{He}_l$ is the $l^{th}$ Hermite polynomial. For ease of presentation, assume that $\mu_0=0$. The $L_2$ norm of $\varrho$ with respect to the standard Gaussian measure is denoted by $\norm{\varrho}$. The function $c$ is defined as $c(\gamma) = \bE{u,v\sim P^\gamma}{\varrho(u)\varrho(v)}$, with $P^\gamma$ denoting the bivariate standard Gaussian distribution with correlation coefficient $\gamma$ (see Eq.~\eqref{eqn:bivariate_gaussian_pdf} in the Appendix~\ref{appndx:lc_minf_proof}).  
\end{assumption}
 
%With these notations and assumptions, the following theorem characterizes the test and train errors of the denoising score matching loss for $m=\infty$ case.
%Within these notations and assumptions, we characterize the test and train error of the DSM loss for $m=\infty$ through the following theorem.
Theorem~\ref{appendix:thm:lc_minf_man} in Appendix~\ref{appndx:lc_minf_proof} characterizes the test and train errors when $P_0$ is a Gaussian supported on a $D$-dimensional subspace in $\R^d$. Here, for simplicity, we present the theorem for $\cC=I_d$, which already captures the main phenomena. %We then characterize the test and train error of the DSM loss for $m=\infty$ through the following theorem.
\begin{theorem}\label{thm:lc_minf}
    Suppose $P_0\equiv\cN{0,I_d}$ and $\varrho$ satisfies Assumption~\ref{assmptn:activation_fn}. Let $s^2 = \norm{\varrho}^2-c(a_t^2)-h_t\mu_1^2$,\; $v_0^2=c(a_t^2)-a_t^2\mu_1^2$, and $v^2 = \norm{\varrho}^2-\mu_1^2$. Let $\psi_n = \frac{n}{d}$, and $\psi_p = \frac{p}{d}$. Let $\zeta_1,\zeta_2,\zeta_3,\zeta_4$ be the solution of the following set of algebraic equations as a function of $q$ and $z$:
\begin{equation}
    \begin{aligned}
        \zeta_2(\psi_n+v_0^2\psi_p\zeta_1-a_t\mu_1\zeta_3)-\psi_n &=0 \;,\nonumber \\
        a_t^2\mu_1^2\psi_p\zeta_1\zeta_2\zeta_4+(1+(h_t\mu_1^2+q)\psi_p\zeta_1)\zeta_4 -1 &=0 \;,\nonumber \\
        \zeta_1(s^2-z+a_t^2\mu_1^2\zeta_2\zeta_4+v_0^2\zeta_2+(h_t\mu_1^2+q)\zeta_4) -1 &=0 \;,\nonumber \\
        a_t^2\mu_1^2\psi_p\zeta_1\zeta_2\zeta_4+(1+(h_t\mu_1^2+q)\psi_p\zeta_1)\zeta_4 -1 &=0 \;.\nonumber \\
    \end{aligned}
\end{equation}
%    \begin{equation*}
%        \begin{aligned}[c]
%            \zeta_2(\psi_n+v_0^2\psi_p\zeta_1-a_t\mu_1\zeta_3)-\psi_n =0 \;,
%        \end{aligned}
%        \quad
%        \begin{aligned}[c]
%            a_t^2\mu_1^2\psi_p\zeta_1\zeta_2\zeta_4+(1+(h_t\mu_1^2+q)\psi_p\zeta_1)\zeta_4 -1= 0 \;,
%        \end{aligned}
%    \end{equation*}
  
%    \begin{align*}
%    \zeta_1(s^2-z+a_t^2\mu_1^2\zeta_2\zeta_4+v_0^2\zeta_2+(h_t\mu_1^2+q)\zeta_4) -1= 0 \;,\nonumber\\
    %a_t^2\mu_1^2\psi_p\zeta_1\zeta_2\zeta_4+(1+(h_t\mu_1^2+q)\psi_p\zeta_1)\zeta_4 -1= 0 \;.\nonumber\\
    %\end{align*}
    Define the function $\cK(q,z) = -\frac{\zeta_3(q,z)}{a_t\mu_1}$. 
    Let $\varepsilon^\infty_{\text{test}} = 1-2\mu_1^2\cK(0,-\lambda)-\mu_1^4\frac{\partial \cK}{\partial q}(0,-\lambda)+\mu_1^2v^2\frac{\partial \cK}{\partial z}(0,-\lambda)$, and $\varepsilon^\infty_{\text{train}} = 1-\mu_1^2h_t\cK(0,-\lambda)-\mu_1^2\lambda h_t\frac{\partial \cK}{\partial z}(0,-\lambda)$. Then, for the minimizer of (\ref{eqn:dsm_loss_rfm_minf}) $\hat{A}_t$, as $d,n,p\to\infty$:
    %\begin{align*}
    %    \lim_{d,n,p\to\infty}\bE{}{\cE^\infty_{\text{test}}(\hat{A}_t)} =  \varepsilon^\infty_{\text{test}}\;,\quad
    %\lim_{d,n,p\to\infty}\bE{}{\cE^\infty_{\text{train}}(\hat{A}_t)} &= \varepsilon^\infty_{\text{train}} \;.
    %\end{align*}
    \begin{equation}
        \begin{aligned}
            \lim_{d,n,p\to\infty}\bE{}{\cE^\infty_{\text{test}}(\hat{A}_t)}  &=  \varepsilon^\infty_{\text{test}} \nonumber \;, \\
             \lim_{d,n,p\to\infty}\bE{}{\cE^\infty_{\text{train}}(\hat{A}_t)} &= \varepsilon^\infty_{\text{train}} \nonumber \;.
        \end{aligned}
    \end{equation}

    %Let $e_1 = \cK(0,-\lambda),\; e_2 = -\frac{\partial \cK}{\partial q}(0,-\lambda),\; e_3 = \frac{\partial \cK}{\partial z}(0,-\lambda)$. Then, for the minimizer of (\ref{eqn:dsm_loss_rfm_minf}) $\hat{A}_t$, as $d,n,p\to\infty$: 
    % \begin{align*}
    %     \lim_{d,n,p\to\infty}\bE{}{\cE^\infty_{\text{test}}(\hat{A}_t)} &= 1-\mu_1^2(2e_1-\mu_1^2e_2-v^2e_3) \;,\\
    % \lim_{d,n,p\to\infty}\bE{}{\cE^\infty_{\text{train}}(\hat{A}_t)} &= 1-\mu_1^2(h_te_1+\lambda h_te_3) \;.
    % \end{align*}
\end{theorem}
We defer the proof to Appendix~\ref{appndx:lc_minf_proof}. 
\begin{remark}\label{rem:concentration}
    We expect $\cE^\infty_{\text{test}}(\hat{A}_t)$ and $\cE^\infty_{\text{train}}(\hat{A}_t)$ to concentrate on their expectations as $d\to\infty$, while a rigorous proof is beyond the scope of the current work. Henceforth, we use the term test (train) error for both $\cE^\infty_{\text{test}}$ ($\cE^\infty_{\text{train}}$) and $\bE{}{\cE^\infty_{\text{test}}}$ ($\bE{}{\cE^\infty_{\text{train}}}$), interchangeably.
\end{remark}
\begin{remark}\label{rem:covariance}
    In Appendix \ref{appndx:lc_minf_proof}, Lemma \ref{lemma:lc_minf}, we first obtain expressions for non-averaged errors when $P_0=\mathcal{N}(0, \mathcal{C})$. In order to derive closed-form formulas for their expected values, one has to assume some structure for the covariance $\mathcal{C}$. In Theorem~\ref{appendix:thm:lc_minf_man} we present the scenario where $P_0$ is a Gaussian distribution supported on a $D$-dimensional subspace in $\R^d$; and analytically obtained curves are illustrated for $\frac{D}{d}=0.2$ in Appendix~\ref{appendix:lowdim}%, whose analysis is already non-trivial. 
\end{remark}
%\begin{figure}
%    \centering    \includegraphics[width=0.85\textwidth]{figs/rfm_err_analytical_lam0p001_relu_nfixed_minf.pdf}
%    \caption{Learning curves for $m=\infty$, with $\psi_n=20.0,\lambda=0.001$. The activation function is ReLU.}
%    \label{fig:lc_minf_rp}
%\end{figure}
\begin{figure}[ht]
    \centering
\includegraphics[width=0.8\linewidth]{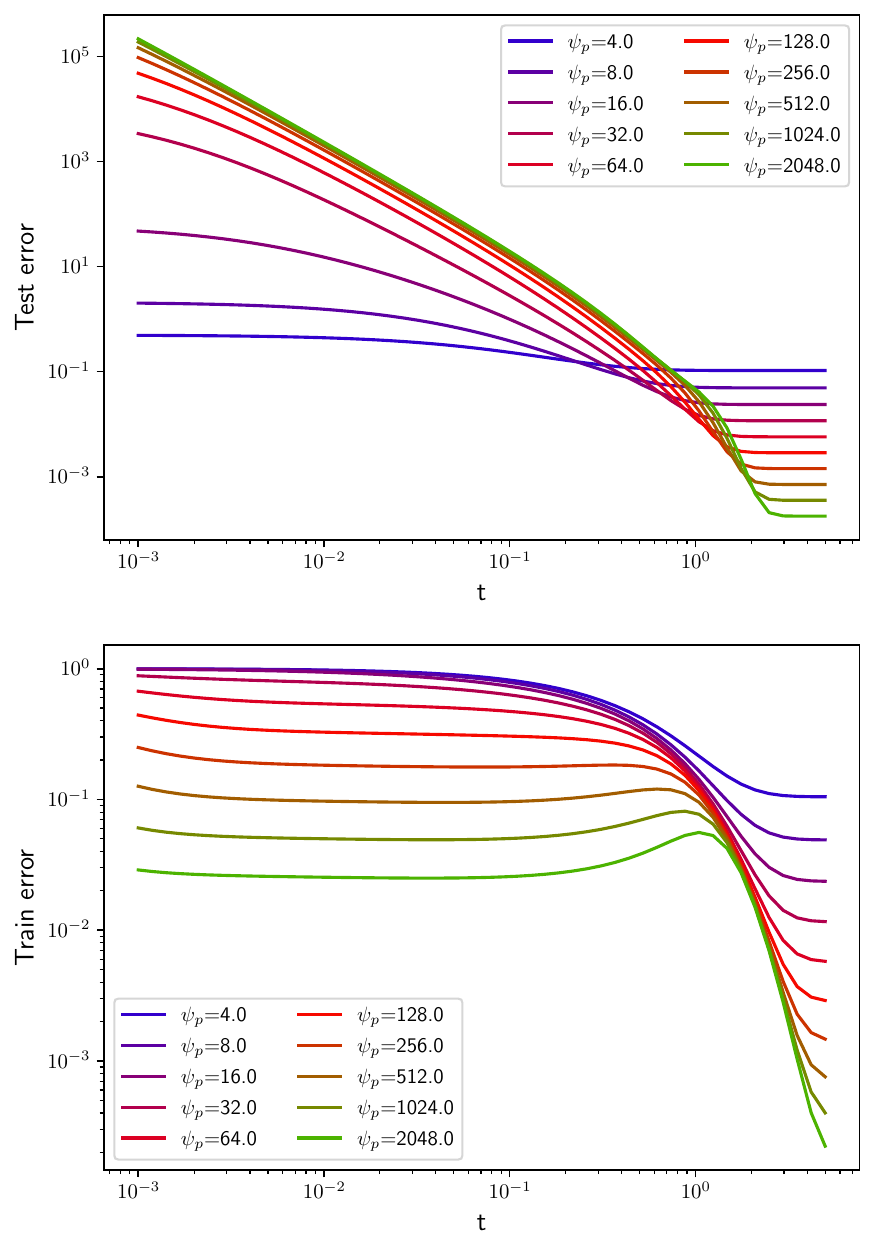}
    \caption{Learning curves for $m=\infty$, with $\psi_n=20.0,\lambda=0.001$. The activation function is ReLU.}
    \label{fig:lc_minf_rp}
\end{figure}

Theorem~\ref{thm:lc_minf} allows the computation of test and train errors for different values of $t,\psi_n,\psi_p$. Fig.~\ref{fig:lc_minf_rp} shows the results as a function of $t$ for various $\psi_p$ with fixed $\psi_n$. We can comprehend them by decomposing $\cE^\infty_{\text{train}}(\hat{A}_t)$ into bias and variance components through the following Lemma, which is proved in Appendix~\ref{appndx:bias_var_proof}.
\begin{lemma}\label{lem:bias_variance_minf}
    Suppose $\cM_t$ and $\cV_t$ are given by
%\begin{align*}
%    \cM_t &= \frac{1}{dn}\sum_{i=1}^{n}{\shortexpect_{z}{\norm{\rat{\hat{A}_t}{p}\act{\rat{W_t}{d}(a_tx_i+\sqrt{h_t}z)}-s^e(t,a_tx_i+\sqrt{h_t}z)}^2}} \;, \\
%    \cV_t &= \frac{1}{dn}\sum_{i=1}^{n}{\shortexpect_{z}{\norm{\sqrt{h_t}s^e(t,a_tx_i+\sqrt{h_t}z)+z}^2}}\;,
%\end{align*}
\begin{align*}
    \cM_t &= \frac{1}{dn}\sum_{i=1}^{n}{\shortexpect_{z}{\norm{\rat{\hat{A}_t}{p}\act{\rat{W_t}{d} y_i^t (z)}-s^e(t,y_i^t (z))}^2}} \;, \\
    \cV_t &= \frac{1}{dn}\sum_{i=1}^{n}{\shortexpect_{z}{\norm{\sqrt{h_t}s^e(t, y_i^t (z))+z}^2}}\;,
\end{align*}
where $s^e(t,x) = \nabla\log P^e_t(x) $ with $P^e_t$ given by Eq.~\eqref{eqn:marginal_prob_empirical}. Then, $\cE^\infty_{\text{train}}(\hat{A}_t) = \cV_t+h_t\cM_t.$
\end{lemma}
%A proof of Lemma~\ref{lem:bias_variance_minf} can be found in Appendix~\ref{appndx:bias_var_proof}.
%Lemma~\ref{lem:bias_variance_minf} is proved in Appendix~\ref{appndx:bias_var_proof}.
We discuss Fig.~\ref{fig:lc_minf_rp} by exploring the bias-variance decomposition above for different values of $t$. A general observation is that, since $\cV_t$ is independent of $\hat{A}_t$, any change in $\cE^\infty_{\text{train}}(\hat{A}_t)$ as the relative number of features $\psi_p$ varies must be attributed to changes in $\cM_t$. A smaller $\cM_t$ indicates that the learned score is closer to the empirical optimal score. Such closeness can harm the generalization performance of diffusion models. This can be understood as follows: 
%By discussing the behaviours of $\cM_t$ and $\cV_t$ for different values of $t$, we explain the learning curves in Fig.~\ref{fig:lc_minf_rp}. The parameter $\psi_p$ is used to control the approximation power of RFNNs. A general observation is that, since $\cV_t$ is independent of $\hat{A}_t$, any change in $\cE^\infty_{\text{train}}(\hat{A}_t)$ as $\psi_p$ varies must be attributed to the changes in $\cM_t$. A smaller value of $\cM_t$ indicates that the learned score function is closer to the empirical optimal score function, as its definition suggests. A score function close to the empirical optimal score function can negatively affect the generalization performance of diffusion models. This can be understood as follows: 

%We discuss the behavior of $\cM_t$ and $\cV_t$ for different values of $t$, thereby explaining the learning curves in Fig.~\ref{fig:lc_minf_rp}. The parameter $\psi_p$ is used to control the approximation power of RFNNs. A general observation is that, since $\cV_t$ is independent of $\hat{A}_t$, any change in $\cE^\infty_{\text{train}}(\hat{A}_t)$ as $\psi_p$ varies must be attributed to the changes in $\cM_t$. A smaller value of $\cM_t$ indicates that the learned score function is closer to the empirical optimal score function, as its definition suggests. A score function close to the empirical optimal score function can negatively affect the generalization performance of diffusion models. This can be understood as follows: 

 %For large $t$, we have $a_t\approx 0,h_t\approx 1$, which gives that $s^e(t,x)\approx -x$.
 For small $t$, $h_t\approx 0$, so in the neighborhood of $a_tx_i$, $P_t^e$ is dominated by the $i^{th}$ term in~\eqref{eqn:marginal_prob_empirical}. Therefore, in the vicinity of $a_tx_i$, we have $s^e(t,x)\approx -\frac{x-a_tx_i}{h_t}$. In the backward process, this translates to a component in the drift pointing towards $a_tx_i$. Consequently, the trajectories will tend to move toward training samples, causing the output of the backward process to resemble one of them. This behavior is referred to as \textit{memorization}, occurring when the learned score closely approximates the empirical optimal one. With this in mind, we now qualitatively discuss the curves in Fig.~\ref{fig:lc_minf_rp} for different values of $t$.
\begin{itemize}[leftmargin=*]
    %\item \textbf{$t=\infty$}: we have $a_t=0,h_t=1$ and $s^e(t,y)=-y$. Substituting these, we get
    %    $\cE^\infty_{\text{train}}(\hat{A}_t) = \cM_t = \shortexpect_z\norm{\rat{\hat{A}_t}{p}\act{\rat{W_t}{d}z}+z}^2.$
    %Hence, the train and test errors are equal and depend only on how well the RFNN can approximate a linear function. As $\psi_p$ increases, the approximation power of the RFNN increases, and thus the train and test errors decrease. 
    \item \textbf{$t=\infty$}:  we have $a_t=0,h_t=1$ and $s^e(t,y)=-y$, leading to $\cE^\infty_{\text{train}}(\hat{A}_t) = \cM_t = \shortexpect_z\norm{\rat{\hat{A}_t}{p}\act{\rat{W_t}{d}z}+z}^2$. The train and test errors are then equal and depend only on how well the RFNN can approximate a linear function. As $\psi_p$ increases, the approximation power of the RFNN increases, and thus the train and test errors decrease. 

    %\item \textbf{$t\gg1$}: we still have $s^e(t,y)\approx -y$, leading to $\cV_t\approx a_t^2$. Hence, $\cE^\infty_{\text{train}}(\hat{A}_t)\approx h_t\cM_t+a_t^2$. When $\psi_p$ is large, $\cM_t$ is small, and the train error is dominated by the $a_t^2$ term. Therefore, we see that train error increases rapidly as $t$ decreases. However, the test error remains constant, since it is approximately equal to $\cM_t$.
    \item \textbf{$t\gg1$}: again $s^e(t,y)\approx -y$, leading to $\cV_t\approx a_t^2$. Hence, $\cE^\infty_{\text{train}}(\hat{A}_t)\approx h_t\cM_t+a_t^2$. When $\psi_p$ is large, $\cM_t$ is small, and the train error is dominated by the $a_t^2$ term. Therefore, the train error increases rapidly as $t$ decreases. However, the test error remains constant, since it is approximately equal to $\cM_t$.

    %\item \textbf{$t\ll1$}: the empirical optimal score satisfies $s^e(t,a_tx_i+\sqrt{h_t}z)\approx - \sfrac{z}{\sqrt{h_t}}$ for any data point $x_i$. Consequently, $\cV_t\approx 0$. This leads to the result $\cE^\infty_{\text{train}}(\hat{A}_t)\approx h_t\cM_t$. In this regime, the train error depends on how well the RFNN can approximate the empirical optimal score $s^e$. With an increase in $\psi_p$, the approximation power increases and the train error decreases. However, since the actual score function significantly deviates from the empirical score in this regime, the test error increases rapidly with $\psi_p$. In the neighborhood of $a_tx_i$, learning the empirical optimal score instead of the actual score makes the drift in the backward process to pull the trajectories towards $a_tx_i$. Thus, if the backward trajectory comes in the vicinity of $a_tx_i$ at time $t$, the diffusion model tends to recover the sample $x_i$.
    \item \textbf{$t\ll1$}: the empirical optimal score satisfies $s^e(t,a_tx_i+\sqrt{h_t}z)\approx - \sfrac{z}{\sqrt{h_t}}$ for any data point $x_i$. Consequently, $\cV_t\approx 0$. This leads to $\cE^\infty_{\text{train}}(\hat{A}_t)\approx h_t\cM_t$. In this regime, train error depends on how well the RFNN approximates $s^e$, decreasing as $\psi_p$ grows. By contrast, since the actual score significantly deviates from the empirical one, test error rises quickly with $\psi_p$. In the neighborhood of $a_tx_i$, learning $s^e$ instead of the true score makes the drift in the backward process pull the trajectories towards $a_tx_i$, so if a trajectory enters the vicinity of $a_tx_i$ at time $t$, the model tends to recover the sample $x_i$.
    
    %This is referred to as memorization.
    % \item \textbf{$t\approx 0$}: In this regime we still have $s^e(t,a_tx_i+\sqrt{h_t}z)\approx -\frac{z}{\sqrt{h_t}}$. However, due to regularization, $\hat{A}_t$ cannot take very large values. So, $\cM_t$ scales as $\frac{1}{h_t}$. This makes $\cE^\infty_{\text{train}}(\hat{A}_t)\approx 1$ for very small values of $t$. When $\psi_p$ is large, the test error remains high due to the memorization phenomenon.
\end{itemize}
\begin{figure}[ht]
    \centering
\includegraphics[width=0.48\textwidth]{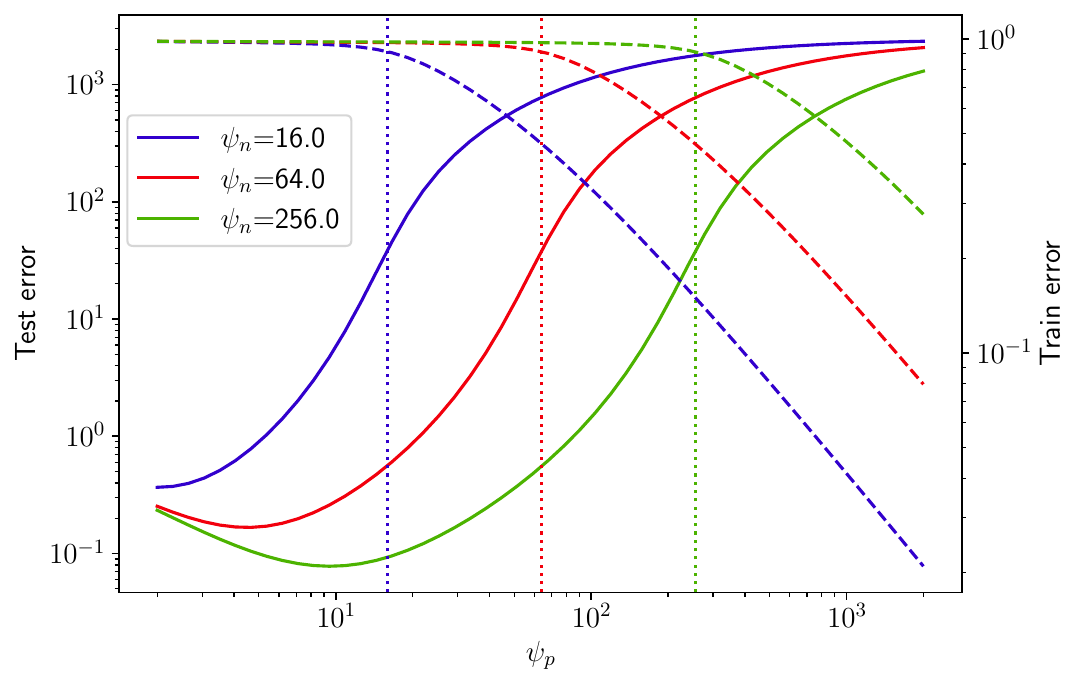}
    \caption{Learning curves for $m=\infty$, with $t=0.01,\lambda=10^{-3}$, $\varrho\equiv $ReLU. Solid (dashed) lines: test (train) error. Dotted vertical lines indicate $\psi_p=\psi_n$.}
    \label{fig:lc_minf_tfixed}
    \vspace{-10pt}
\end{figure}

The analysis of Fig.~\ref{fig:lc_minf_rp} hints that the RFNN starts to show memorization behavior for $t<1$ and large $\psi_p$. To explore this further, we plot in Fig.~\ref{fig:lc_minf_tfixed} the learning curves as a function of $\psi_p$ for different $\psi_n$, with $t$ kept  fixed and small.
We observe: 1) For $\psi_p\ll\psi_n$, train error remains constant and test error is small, indicating generalization. 2) For $\psi_p\gg\psi_n$, train error is small and test error high, indicating the presence of memorization. 3) When $\psi_p\approx\psi_n$, test error rises rapidly, signaling the onset of memorization.

Therefore, $\psi_p=\psi_n$ acts as a crossover point at which the behavior of the score transitions from generalization phase to memorization phase. This is depicted in Fig.~\ref{fig:phase_diagram} as a phase diagram. %We highlight that this shift in behavior is not a sharp transition but rather a gradual change.   

More curves illustrating the effects of $\lambda$ and $\varrho$ are found in Appendices~\ref{appndx:lc_lambda} and \ref{appndx:lc_activation}. Our theory is consistent with previous findings on the roles of $n$ and $p$ on generalization and memorization in practical settings~\citep{zhang_emergence_2024,yoon_diffusion_2023,gu_memorization_2023}.

%Additionally, we highlight that the same trends predicted by our theory under a simplified setting have been observed by~\cite{zhang_emergence_2024} on score parametrized by U-Net architectures trained with CIFAR-10 data. In Fig. 2(b) of this reference, they observe generalization increasing with the number of samples, while in their Fig.3(b) they observe that the train loss decreases with $p$ for a fixed $n$, as depicted in Fig.~\ref{fig:lc_minf_tfixed}.

\subsection{Single Noise Sample per Data Sample}
%We now focus on the $m=1$ case. With $y^t_i = a_tx_i+\sqrt{h_t}z^t_i$, we can write the denoising score matching loss as
For $m=1$, the DSM loss~\eqref{eqn:dsm_loss_rfm} reduces to
\begin{align}\label{eqn:dsm_loss_rfm_m1}
    \cL^1(A_t) &= \frac{1}{dn}\sum_{i=1}^{n}{\norm{\sqrt{h_t}\rat{A_t}{p}\act{\rat{W_t}{d}y^t_i}+z^t_i}^2}\nonumber\\
    &\qquad +\frac{h_t\lambda}{dp}\norm{A_t}_F^2 \;.
\end{align}
%\begin{equation}\label{eqn:dsm_loss_rfm_m1}
%    \cL^1(A_t) = \frac{1}{dn}\sum_{i=1}^{n}{\norm{\sqrt{h_t}\rat{A_t}{p}\act{\rat{W_t}{d}(a_tx_i+\sqrt{h_t}z^t_{i})}+z^t_i}^2} +\frac{h_t\lambda}{dp}\norm{A_t}_F^2 \;.
%\end{equation}
The train and test errors of its minimizer for $\cC=I_d$ are characterized below by Theorem~\ref{thm:lc_m1}. The more general case, where $P_0$ is supported on a $D$-dimensional subspace in $\R^d$, is deferred to Appendix~\ref{appndx:lc_m1_proof} .
%The following theorem characterizes the test and train errors of the minimizer of the loss given in (\ref{eqn:dsm_loss_rfm_m1}).

\begin{theorem}\label{thm:lc_m1}
    %Let $P_0\equiv\cN{0,I_d}$, and $\varrho$ satisfies Assumption~\ref{assmptn:activation_fn}. Let $v^2 = \norm{\varrho}^2-\mu_1^2$,\; $\psi_n = \frac{n}{d}$, and $\psi_p = \frac{p}{d}$. Let $\zeta_1,\zeta_2,\zeta_3,\zeta_4$ be the solution of the following set of algebraic equations as a function of $q$ and $z$:
    Let $P_0\equiv\cN{0,I_d}$, and let $\varrho$ satisfy Assumption~\ref{assmptn:activation_fn}. Define $v^2 = \norm{\varrho}^2-\mu_1^2$,\; $\psi_n = \frac{n}{d}$, and $\psi_p = \frac{p}{d}$. Let $\zeta_1,\zeta_2,\zeta_3,\zeta_4$ be the solution to the following system of algebraic equations in $q$ and $z$:
    %\begin{equation*}
    %    \begin{aligned}[c]
    %    \zeta_1(-z+q\zeta_2+v^2\zeta_4+\mu_1^2\zeta_2\zeta_4) -1&= 0\;,\\
    %    \zeta_2(1+q\psi_p \zeta_1) +\mu_1^2\psi_p\zeta_1\zeta_2\zeta_4 -1&= 0\;,\\
    %    \zeta_3(1+q\psi_p \zeta_1) +\mu_1\psi_p\zeta_1(1+\mu_1\zeta_3\zeta_4) &= 0\;,\\
    %    \end{aligned}
    %    \quad
    %    \begin{aligned}[c]
    %    \zeta_4(\psi_n+v^2\psi_p\zeta_1-\mu_1\zeta_3)-\psi_n &= 0\;,\\
    %    \zeta_4(-h_t\mu_1\zeta_3\zeta_4-1)+\psi_n\zeta_5 &= 0\;. 
    %    \end{aligned}
    %\end{equation*}
    \begin{equation}
        \begin{aligned}
    \zeta_1(-z+(q+\mu_1^2\zeta_4)\zeta_3+v^2\zeta_4) -1&= 0 \;,\\
    \zeta_2(1+q\psi_p \zeta_1) +\mu_1^2\psi_p\zeta_1\zeta_2\zeta_4 +\psi_pa_t\mu_1\zeta_1&= 0 \;,\\
    \zeta_3(1+q\psi_p \zeta_1) +\mu_1^2\psi_p\zeta_1\zeta_3\zeta_4 -1&= 0 \;,\\
    \zeta_4(\psi_n+\psi_pv^2\zeta_1 -\mu_1\zeta_2/a_t)-\psi_n &= 0 \;.
        \end{aligned}
    \end{equation}   
    Define the function $\cK(q,z) = (1-\zeta_4(q,z)(1+\mu_1h_t\zeta_4(q,z)\zeta_2(q,z)/a_t))$. 
    Let $\varepsilon^1_{\text{test}} = 1+2\mu_1\zeta_2(0,-\lambda)\zeta_4(0,-\lambda)/a_t-\frac{\mu_1^2}{h_t}\frac{\partial \cK}{\partial q}(0,-\lambda)+\frac{v^2}{h_t}\frac{\partial \cK}{\partial z}(0,-\lambda)$, and $\varepsilon^1_{\text{train}} = 1-\cK(0,-\lambda)-\lambda \frac{\partial \cK}{\partial z}(0,-\lambda)$. Then, for the minimizer of (\ref{eqn:dsm_loss_rfm_m1}) $\hat{A}_t$, as $d,n,p\to\infty$:
    %\begin{align*}
    %    \lim_{d,n,p\to\infty}\bE{}{\cE^1_{\text{test}}(\hat{A}_t)} =  \varepsilon^1_{\text{test}}\;,\quad
    %\lim_{d,n,p\to\infty}\bE{}{\cE^1_{\text{train}}(\hat{A}_t)} &= \varepsilon^1_{\text{train}} \;.
    %\end{align*}
    \begin{equation}
        \begin{aligned}
             \lim_{d,n,p\to\infty}\bE{}{\cE^1_{\text{test}}(\hat{A}_t)} &= \varepsilon^1_{\text{test}} \;, \nonumber \\
             \lim_{d,n,p\to\infty}\bE{}{\cE^1_{\text{train}}(\hat{A}_t)}  &=\varepsilon^1_{\text{train}} \;.
        \end{aligned}
    \end{equation}
    
    % Let $e_1 = -\sqrt{h_t}\zeta_3(0,-\lambda)\zeta_4(0,-\lambda),\; e_2 = -\frac{\partial \cK}{\partial q}(0,-\lambda)$, and $e_3 = \frac{\partial \cK}{\partial z}(0,-\lambda)$. Then, for
    % the minimizer of (\ref{eqn:dsm_loss_rfm_m1}) $\hat{A}_t$, as $d,n,p\to\infty$:
    % \begin{align*}
    %     \lim_{d,n,p\to\infty}\bE{}{\cE^1_{\text{test}}(\hat{A}_t)} &= 1-\frac{2\mu_1}{\sqrt{h_t}}e_1+\frac{\mu_1^2}{h_t}e_2+\frac{v^2}{h_t}e_3\;,\\
    % \lim_{d,n,p\to\infty}\bE{}{\cE^1_{\text{train}}(\hat{A}_t)} &= 1-\cK(0,-\lambda)-\lambda e_3\;.
    % \end{align*}
\end{theorem}
The proof is given in Appendix~\ref{appndx:lc_m1_proof}, and Remarks~\ref{rem:concentration}-\ref{rem:covariance} apply here as well. Using Theorem~\ref{thm:lc_m1}, we compute test and train errors as functions of $t,\psi_n$ and $\psi_p$, as shown in Fig.~\ref{fig:lc_m1} in Appendix~\ref{appndx:learning_curves_m1}. In Appendix~\ref{appendix:lowdim}, we also illustrate the case where $P_0$ is supported on the $D$ dimensional subspace with $\frac{D}{d}=0.2$.
% shows the plots thus obtained.
%we first derive in Lemma \ref{lemma:lc_m1} general expressions for non-averaged errors when $P_0=\mathcal{N}(0, \mathcal{C})$, and then specialize to the isotropic case to compute expectation values in closed form. 

%Fig.~\ref{fig:lc_m1_nfixed} shows learning curves as a function of $t$ for different $\psi_p$ with fixed $\psi_n$. Several notable trends emerge. Test error increases as $t$ decreases; however, it shows a non-monotonic behavior with $\psi_p$. Train error decreases monotonically with increasing $\psi_p$ for all $t$, indicating the model's progressive capacity to interpolate the training data. 

Fig.~\ref{fig:lc_m1_nfixed} shows learning curves as a function of $t$ for different $\psi_p$ with fixed $\psi_n$. Several notable trends emerge. Test error increases as $t$ decreases but varies non-monotonically with $\psi_p$. Train error decreases monotonically with $\psi_p$ for all $t$, indicating the model's progressive capacity to interpolate the training data.

%Note that for small $t$, the test error remains at least two orders of magnitude smaller than in the $m=\infty$ case. Further, the test error decreases as $\psi_p$ increases beyond $\psi_n$. These observations suggest the model does not display memorization behavior when $m$ equals $1$. This contrasts the $m=\infty$ case, where memorization significantly impacts the test error. These findings indicate that larger values of $m$ increase the propensity to memorize the training data, justifying the illustration in Fig.~\ref{fig:phase_diagram}. 

Note that for small $t$, the test error remains at least two orders of magnitude smaller than in the $m=\infty$ case. Further, it decreases as $\psi_p$ increases beyond $\psi_n$. These observations suggest the model does not display memorization behavior when $m$ equals $1$. This contrasts the $m=\infty$ case, where memorization significantly impacts the test error. These findings indicate that larger values of $m$ increase the propensity to memorize, justifying the illustration in Fig.~\ref{fig:phase_diagram}.

The DSM loss (\ref{eqn:dsm_loss_rfm_m1}) shares similarities with the loss used for RFNN in regression settings. Several works such as \cite{mei_generalization_2022, bodin_model_2021, hu_asymptotics_2024}, have analyzed the learning curves of RFNN in regression contexts. Notably, they predict the presence of a double descent phenomenon: test error peaks at $\psi_p=\psi_n$ and decreases for $\psi_p<\psi_n$ or $\psi_p>\psi_n$. The point $\psi_p=\psi_n$ is called the interpolation threshold. We also observe double descent in the DSM setting with $m=1$, as shown in Fig.~\ref{fig:lc_m1_tfixed}.

%In our study, we also observe the double descent phenomenon in the DSM setting with $m=1$. This is depicted in Fig.~\ref{fig:lc_m1_tfixed}. 

\paragraph{Intermediate values of $m$} Simulations of learning curves for intermediate $m$ are presented in Appendix~\ref{appndx:lc_numerical_comparison}. They also confirm the predictions of Theorems~\ref{thm:lc_minf} and \ref{thm:lc_m1}. When $\psi_p>\psi_n$, the increase in memorization with a larger $m$ can be explained as follows: For large $m$ and small $t$, the learned score $\hat{s}(x)$ effectively approximates $\frac{x_1-x}{h_t}$ in the vicinity of the training sample $x_1$. During the sampling phase, the drift term is given by $Y_t+2\hat{s}(Y_t)$, implying that whenever the sampling trajectory is near a training sample, there is a drift towards it, leading to memorization. However, for $\psi_p<\psi_n$, the model's limited expressibility helps elude memorization even if $m$ is large.

%In Appendix~\ref{appndx:lc_numerical_comparison}, we compare the predictions of Theorems~\ref{thm:lc_minf} and \ref{thm:lc_m1} to numerical simulations. Additionally, we conduct simulations of the leaning curves for intermediate values of $m$. 

\section{NUMERICAL EXPERIMENTS ON MEMORIZATION}\label{sec:numerical_exps}
\begin{figure}
     \centering
     \begin{subfigure}{0.9\linewidth}
         \centering
         \includegraphics[width=0.8\textwidth]{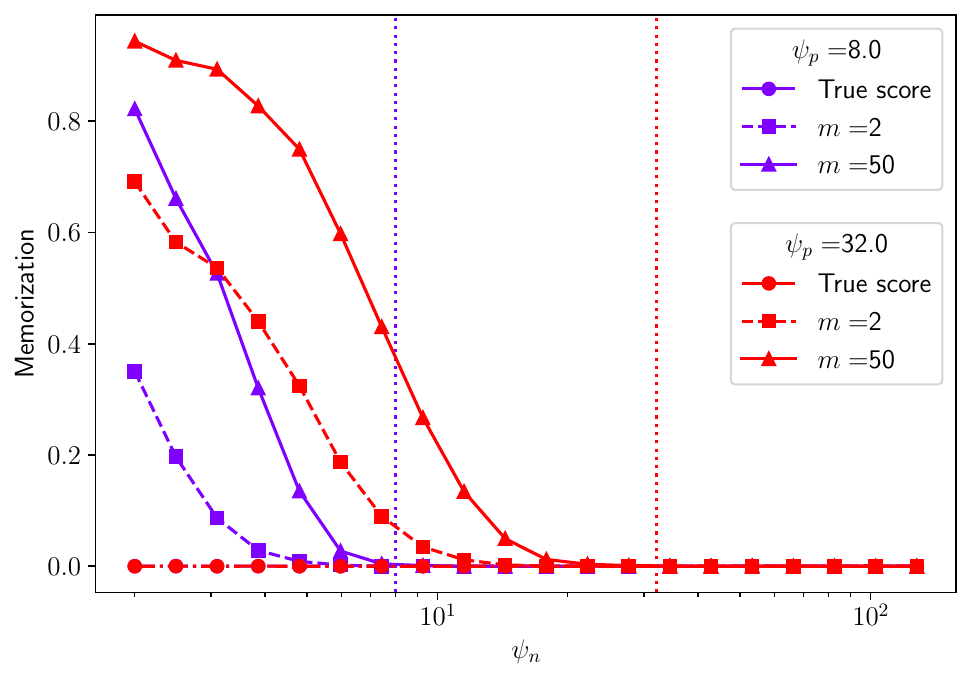}
    \caption{Gaussian data and RFNN score ($N=5000$ and $\delta=\sfrac{1}{3}$). The dotted vertical lines are for $\psi_n=\psi_p$.}
    \label{fig:mem_sim_rfnn}
     \end{subfigure}
     \medskip
     \begin{subfigure}{0.9\linewidth}
         \centering
        \includegraphics[width=0.8\textwidth]{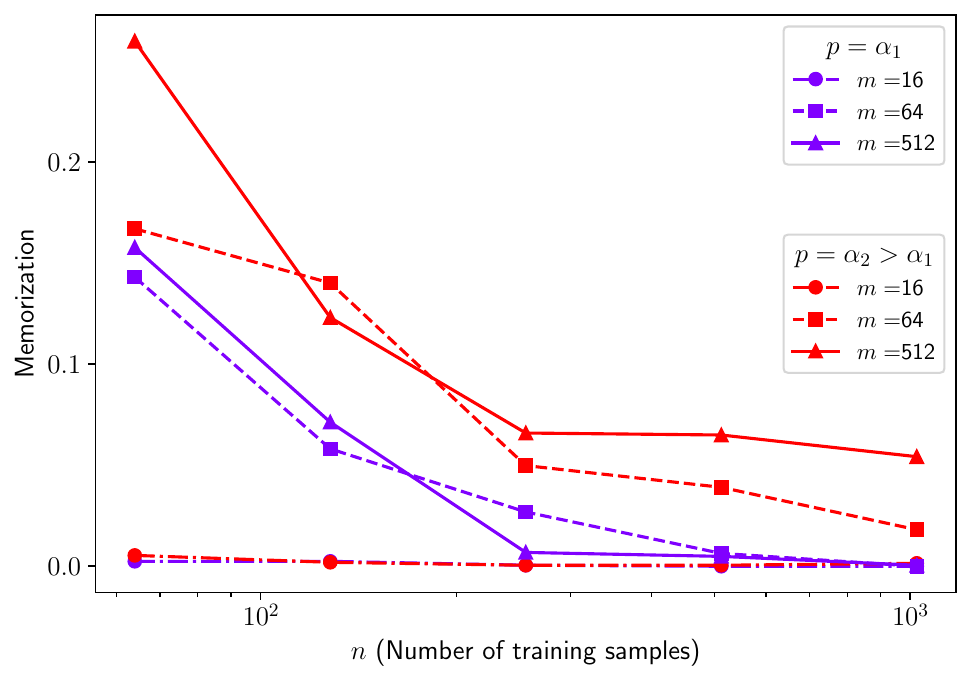}
         \caption{Fashion-MNIST dataset and U-Net score ($N=2000$ and $\delta=\sfrac{1}{2}$).}
    \label{fig:mem_sim_fmnist}
     \end{subfigure}
        \caption{Results of experiments on memorization.}
        \label{fig:mem_sim}
\end{figure}

We now conduct numerical experiments to illustrate the effects of $n,p$ and $m$ on memorization when a learned score function $s_{\hat{\theta}}$ is used in the backward diffusion \eqref{eqn:backward_sde}.  
%In this section, through experiments, we illustrate the effect of $n,p$ and $m$ on memorization when a learned score function $s_{\hat{\theta}}$ is used in the backward diffusion (\ref{eqn:backward_sde}). 
Specifically, we simulate $N$ instances of $-\dd Y_t = \left(Y_t+2s_{\hat{\theta}}(t,Y_t)\right) \;  \dd t+\sqrt{2} \; \dd \tilde{B}_t \;$.
We stop at $t_0=10^{-5}$ and check whether $Y_{t_0}$ is closer to one of the training samples compared to the others~\citep{yoon_diffusion_2023}. Let $\text{NN}_i(x)$ denote the $i^{th}$ nearest neighbor of $x$ in $\{x_1,x_2,\cdots,x_n\}$. We say the backward diffusion retrieves a data sample if $\norm{Y_{t_0}-\text{NN}_1(Y_{t_0})}<\delta \norm{Y_{t_0}-\text{NN}_2(Y_{t_0})}$,
% \begin{equation}\label{eqn:mem_nn_condition}
% \frac{\norm{Y_{t_0}-\text{NN}_1(Y_{t_0})}}{\norm{Y_{t_0}-\text{NN}_2(Y_{t_0})}}<\delta,
% \end{equation}
with $\delta >0$. We measure \textit{memorization} as the fraction of $N$ backward diffusion instances that retrieves one of the data samples.
\paragraph{Gaussian data and RFNN score}\label{sec:expmnt_rfnn}
First, we measure memorization when the $P_0\equiv\cN{0,I_d}$ and $s_{\hat{\theta}}(t,y) =\rat{\hat{A}_t}{p}\act{\rat{W_t}{d}y}$ with $\hat{A}_t$ being the minimizer of (\ref{eqn:dsm_loss_rfm}). We demonstrate that, when $\psi_p>\psi_n$, once the backward diffusion process enters the vicinity of a data sample, it exhibits a tendency to remain within that neighborhood.
%Specifically, we start the backward diffusion at time $t_1=0.1$ in the vicinity of one of the training samples $x_l$.
More precisely, with $t_1 = 0.1$, we start the backward diffusion at $Y_{t_1} = a_{t_1}x_l+\sqrt{h_{t_1}}z$, where $l$ is selected uniformly at random from the set $\{1,2,\cdots,n\}$, and $z\sim\cN{0,I_d}$. Fig.~\ref{fig:mem_sim_rfnn} shows the measure of memorization described in the previous paragraph as a function of $\psi_n$ for different values of $\psi_p$ and $m$. %We observe that the result of this experiment is in line with the predictions made in the previous sections. In particular, 
We note the following: 1) For $m=50$ and $\psi_n<\psi_p$, the memorization is high. 2) The memorization decreases as $m$ decreases. 3) Memorization is zero when $\psi_n>\psi_p$.
\paragraph{Real datasets and U-Net score}
Next, we measure memorization for Fashion-MNIST and MNIST datasets when $s_{\hat{\theta}}$ is a learned neural network with U-Net \citep{ronneberger_u-net_2015} based architecture. 
% \begin{wrapfigure}{l}{0.4\textwidth}
%     \centering
%     \includegraphics[width=0.98\linewidth]{figs/mem_plot_fmnist.pdf}
%          \caption{Fashion-MNIST dataset and U-Net score. We use $N=2000$ and $\delta=1/2$.}
%     \label{fig:mem_sim_fmnist}
%     \vspace{-8pt}
% \end{wrapfigure}
We use a subset of MNIST/Fashion-MNIST dataset of size $n$ for training. For a fixed $n,m,$ and U-Net neural network, we train the U-Net for a large number of epochs (see Appendix~\ref{appndx:real_data_exp} for more details). During the sampling phase, we use the learned score in the backward diffusion. In Fig.~\ref{fig:mem_sim_fmnist}, we plot the memorization for the Fashion-MNIST dataset as a function of $n$ for different $m$, and for two U-Net neural networks with different number of parameters ($p$). Corresponding results for the MNIST dataset can be found in Fig.~\ref{appndx:mem_sim_mnist} in Appendix~\ref{appndx:real_data_exp}. We draw the following conclusions from this experiment: 1) Memorization increases as $\frac{n}{p}$ decreases and 2) Memorization increases as $m$ increases in the overparametrized regime. 

%We remark that the theoretical setting we studied in the previous sections can indeed capture the effect of $n,p$ and $m$ on memorization in practical situations, and we believe this is quite remarkable. 
%Another example is the observation by~\cite{yoon_diffusion_2023} on U-Net for the CIFAR-10 dataset that overparametrization leads to increased memorization (see Fig. 1 in this reference), which is predict within our theory.

Remarkably, our theoretical setting captures the effect of $n,p$ and $m$ on memorization in practical situations.  

\section{DISCUSSION AND FUTURE WORK}
%In this work, we provided theoretical insights into the mechanisms underlying generalization and memorization in diffusion models.
%-- a phenomenon whereby the output of diffusion models resemble one of the training sample.
We studied the mechanisms underlying generalization and memorization in diffusion models by analyzing train and test errors of DSM with random features and Gaussian data. %To the best of our knowledge, this work is the first to provide a theoretical study of memorization when a parametric score is used.
%Memorization is typically observed in practice when highly overparameterized neural networks are trained long enough. In line with these empirical observations, our findings indicate that the model complexity ($p$) and the number of noise samples per data sample ($m$) used during DSM play a significant role in memorization. In practical implementations, $m$ increases as the number of training epochs increases.
Memorization is typically observed in practice when highly overparameterized neural networks are trained long enough. Consistent with this, our findings indicate that model complexity ($p$) and the number of noise samples per data sample ($m$) used during DSM play a significant role in memorization. In practical implementations, $m$ increases as the number of training epochs increases.

As the complexity of the model increases, it can approximate more complex functions, leading to a better approximation of the empirical optimal score. Furthermore, as $m$ increases, an optimizer of DSM loss (\ref{eqn:dsm_mfinite}) tends to be close to the empirical optimal score.  These effects that lead to memorization are captured in our results and are illustrated as a phase diagram in Fig.~\ref{fig:phase_diagram}. As expected, generalization occurs when the complexity of the model is limited, i.e., $p<n$.

%Although our analysis presents simplifications for analytical tractability: RFNNs parametrizing the score, unstructured data and an independent estimator for each time $t$, it is a first step to unveiling the generalization and memorization of DSM in diffusion models. Moreover, the effect of $m$ in memorization has never been theoretically explored to the best of our knowledge, and the behavior predicted by our theory under a simplified setting can be indeed seen for real-world data under a model employed in practice, as shown in Fig.~\ref{fig:mem_sim_fmnist}.

Although our analysis relies on simplifications for analytical tractability: RFNNs for the score, unstructured data and an independent estimator for each time $t$, it is a first step toward understanding the generalization and memorization of DSM in diffusion models. In particular, we provide, to the best of our knowledge, the first theoretical study of the role of $m$ in memorization. Furthermore, the behavior predicted by our theory under this simplified setting can indeed be seen on real data with practical models, as shown in Fig.~\ref{fig:mem_sim_fmnist}.

In our analysis, we adopt an Ornstein–Uhlenbeck (OU) stochastic differential equation (SDE) as the forward process, corresponding to the so-called \textit{variance-preserving} (VP) framework in the diffusion literature. An alternative and widely used choice is the \textit{variance-exploding} (VE) framework, in which the forward process is given by a Brownian motion with increasing variance. Our analysis can be extended to this setting in a straightforward manner by appropriately choosing $a_t$ and $h_t$. We leave a detailed investigation of this case for future work.

Several additional directions could further refine and generalize our analysis. For instance, one could incorporate explicit time dependence in the score function, consider richer and more expressive classes of score models, or relax the Gaussian assumption on the data distribution. Addressing these challenges would bring the theoretical framework closer to practical implementations of diffusion-based generative models and may shed further light on their empirical success.

% In our analysis, we chose an Ornstein-Uhlenbeck SDE for the forward process. In diffusion literature, this is known as the variance preserving framework. Another popular choice for the forward process is a simple Brownian motion known as the variance exploding framework. Our analysis can be easily adapted to the latter framework by using $a_t=1$ and $h_t=\sqrt{t}$. We leave it as a future work the detailed analysis and interpretation of results in that setting. 

% More challenging future directions attempting to refine our analysis could incorporate time as a parameter in the score, consider more expressive score models, or consider more general data distributions. Progress in these aspects would bring the studied setting closer to the ones used in practice.

%%Our work represents a preliminary step in understanding generalization and memorization in diffusion models. Future directions could focus on addressing some of the simplifications inherent in our analysis. Specifically, can we use a time-parameterized model for the score and derive its learning errors? Another direction is to analyze the entire backward diffusion using the learned score and demonstrate the generalization and memorization behaviors. These advancements could further improve our understanding of diffusion models.

\subsubsection*{Acknowledgements}
The work of A. J. G. and R. V. has been supported by Swiss National Science Foundation grant number 200021-204119. 
%\subsubsection*{Acknowledgements}
%All acknowledgments go at the end of the paper, including thanks to reviewers who gave useful comments, to colleagues who contributed to the ideas, and to funding agencies and corporate sponsors that provided financial support. 
%To preserve the anonymity, please include acknowledgments \emph{only} in the camera-ready papers. The acknowledgements do not count against the 9-page page limit in the camera-ready.

%\subsubsection*{References}
%\section*{References}
\bibliography{references}
\clearpage
\appendix
\thispagestyle{empty}

% Supplementary material: To improve readability, you must use a single-column format for the supplementary material.
\onecolumn

\aistatstitle{Supplementary Materials}
%%%%%%%%%%%%%%%%%%%%%%%%%%%%%%%%%%%%%%%%%%%%%%%%%%%%%%%%%%%%%%%%%%%%%%%%%%%%%%%
%%%%%%%%%%%%%%%%%%%%%%%%%%%%%%%%%%%%%%%%%%%%%%%%%%%%%%%%%%%%%%%%%%%%%%%%%%%%%%%
% APPENDIX
%%%%%%%%%%%%%%%%%%%%%%%%%%%%%%%%%%%%%%%%%%%%%%%%%%%%%%%%%%%%%%%%%%%%%%%%%%%%%%%
%%%%%%%%%%%%%%%%%%%%%%%%%%%%%%%%%%%%%%%%%%%%%%%%%%%%%%%%%%%%%%%%%%%%%%%%%%%%%%%
%\newpage
%\appendix
\section{SCORE MATCHING}
\subsection{Proof that minimizers of $\cL_{\text{DSM}}$ and $\cL_{\text{SM}}$ are the same}\label{appndx:dsm_sm_equiv}
From the $L_2$ minimization property of the conditional expectation, the minimizer of $\cL_{\text{DSM}}$ is given by $\hat{s}(t,x) = \expectCond{\nabla\log P_t(x_t|x_0)}{x_t=x}$. We have 
%\begin{align*}
%    &\expectCond{\nabla\log P_t(x_t|x_0)}{x_t=x} \\
%    &\qquad= \int \dd x_0 \frac{\nabla P_t(x_t=x|x_0)}{P_t(x_t=x|x_0)} P_t(x_0|x_t=x)\\
%    &= \int \dd x_0 \nabla P_t(x_t=x|x_0)\frac{P_0(x_0)}{P_t(x)}\\
%    &= \frac{1}{P_t(x)}\nabla\int \dd x_0 P_t(x_t=x|x_0)P_0(x_0)\\
%    &= \nabla \log P_t(x) \;.
%\end{align*}
\begin{align*}
    \expectCond{\nabla\log P_t(x_t|x_0)}{x_t=x} =& \int \dd x_0 \frac{\nabla P_t(x_t=x|x_0)}{P_t(x_t=x|x_0)} P_t(x_0|x_t=x)\\
    =& \int \dd x_0 \nabla P_t(x_t=x|x_0)\frac{P_0(x_0)}{P_t(x)}\\
    =& \frac{1}{P_t(x)}\nabla\int \dd x_0 P_t(x_t=x|x_0)P_0(x_0)\\
    =& \nabla \log P_t(x) \;.
\end{align*}
Hence, we have shown that the minimizer of $\cL_{\text{DSM}}$ is same as the minimizer of $\cL_{\text{SM}}$.

\subsection{Proof of Lemma~\ref{lem:bias_variance_minf}}\label{appndx:bias_var_proof}
Let $P^e_t$ denote the joint probability distribution of $(y_t,z)$, where $y_t= a_tx+\sqrt{h_t}z$, with $x\sim\frac{1}{n}\sum_{i=1}^n\delta_{x_i}$ and $z\sim\cN{0,I_d}$. We have 
\begin{align*}
    \cE^\infty_{\text{train}}(\hat{A}_t) &= \frac{1}{dn}\sum_{i=1}^{n}{\shortexpect_z\norm{\sqrt{h_t}\rat{\hat{A}_t}{p}\act{\rat{W_t}{d}(a_tx_i+\sqrt{h_t}z)}+z}^2}\\
    &= \frac{1}{d}{\bE{P_t^e}{\norm{\sqrt{h_t}\rat{\hat{A}_t}{p}\act{\rat{W_t}{d}y_t}+z}^2}}\\
    &\stackrel{(a)}{=} \frac{1}{d}\bE{P_t^e}{\norm{\sqrt{h_t}\rat{\hat{A}_t}{p}\act{\rat{W_t}{d}y_t}-\sqrt{h_t}s^e(t,y_t)}^2+\norm{\sqrt{h_t}s^e(t,y_t)+z}^2}\\
    &= h_t\cM_t+\cV_t \;.
\end{align*}
The equality $(a)$ follows from:
\begin{align*}
    &\bE{P_t^e}{\left\langle \rat{\hat{A}_t}{p}\act{\rat{W_t}{d}y}-s^e(t,y),\sqrt{h_t}s^e(t,y)+z \right\rangle}\\ 
    &\qquad=\bE{P_t^e}{\expectCond{\left\langle \rat{\hat{A}_t}{p}\act{\rat{W_t}{d}y}-s^e(t,y),\sqrt{h_t}s^e(t,y)+z \right\rangle}{y}}\\
    &\qquad=\bE{P_t^e}{\left\langle \rat{\hat{A}_t}{p}\act{\rat{W_t}{d}y}-s^e(t,y),\sqrt{h_t}s^e(t,y)+\expectCond{z}{y} \right\rangle}\\
    &\qquad= \bE{P_t^e}{\left\langle \rat{\hat{A}_t}{p}\act{\rat{W_t}{d}y}-s^e(t,y),\sqrt{h_t}s^e(t,y)-\sqrt{h_t}\nabla\log P_t^e(y) \right\rangle}\\
    &\qquad= 0 \;,
\end{align*}
since $\expectCond{z}{y} = -\sqrt{h_t}\nabla\log P_t^e(y)$.

%\section{Linear Pencils}\label{appndx:linear_pencils}

% \input{sections/proofs_general_covariance}

\section{PROOF OF THEOREM~\ref{thm:lc_minf}}\label{appndx:lc_minf_proof}
 In this section, we present a more general version of Theorem~\ref{thm:lc_minf}, where $P_0$ is supported on a $D$-dimensional subspace in $\R^d$, and $\psi_D = \frac{D}{d}$ is kept fixed as $D,d\to\infty$. First we define generalization along a subspace as follows: Let $\cM_\parallel$ be a subspace in $\R^d$ and $\cM_\perp$ its orthogonal complement. Let $\Pi_\alpha$ denote a projection matrix that project onto $\cM_\alpha$, for $\alpha\in\{\parallel,\perp\}$. Then, we define the test error along $\cM_\alpha$ to be
\begin{align}
    \cE^m_{\text{test},\alpha}(\hat{A}_t) &= \frac{1}{d}\shortexpect_{x\sim P_t}{\norm{\Pi_\alpha\rat{\hat{A}_t}{p}\act{\rat{W_t}{d}x}-\Pi_\alpha\nabla\log P_t(x)}^2} \label{eq:At_test_inf_man}.
\end{align}
Note that the overall test error is given by $\cE^m_{\text{test}}(\hat{A}_t) = \cE^m_{\text{test},\parallel
}(\hat{A}_t)+\cE^m_{\text{test},\perp}(\hat{A}_t)$. Lemma~\ref{lemma:lc_minf} expresses test and train errors as traces of rational functions of random matrices. For $\kappa>0$, we define the Gaussian expectations: $\mu_0(\kappa) = \bE{g}{\varrho(\kappa g)},\; \mu_1(\kappa) = \bE{g}{\varrho(\kappa g)g},\; \norm{\varrho(\kappa\;\cdot)}^2 = \bE{g}{\varrho(\kappa g)^2}$, $v^2(\kappa) = \norm{\rho(\kappa\;\cdot)}^2-\mu_0^2(\kappa)-\mu_1^2(\kappa)$,  $c(\gamma,\kappa) = \bE{u,v\sim P^\gamma}{\varrho(\kappa u)\varrho(\kappa v)}$ where $g\sim\cN{0,1}$ and $P^\gamma$ denotes the bivariate standard Gaussian distribution with correlation coefficient $\gamma$. Explicitly,
\begin{equation}\label{eqn:bivariate_gaussian_pdf}
    P^\gamma(x,y) = \frac{1}{2\pi\sqrt{1-\gamma^2}}e^{-\frac{x^2+y^2-2\gamma xy}{2(1-\gamma^2)}}\;.
\end{equation}

\begin{lemma}\label{lemma:lc_minf}
    Let $\cM$ be a subspace in $\R^D$. Let $\Pi_\parallel$ denote a projection matrix that project onto $\cM$ and let $\Pi_\perp = I-\Pi_\parallel$. Suppose $P_0\equiv\cN{0,\cC}$, and let $X=[x_1,x_2,\cdots,x_n]$ be the data matrix with columns drawn i.i.d. from $P_0$. Consider the quantities $a_t, h_t, \lambda, \varrho$, and $W_t$ as defined in~\eqref{eqn:dsm_loss_rfm_minf}, with $\varrho$ satisfying Assumption~\ref{assmptn:activation_fn}. Define $\Sigma_t = a_t^2\cC+h_tI_d$, $\sigma_t^2 = \frac{1}{d}\tr{\Sigma_t}$, and $\sigma_0^2 = \frac{1}{d}\tr{\cC}$. Assume that $\sigma_t$ converges to a finite limit as $d\to\infty$. Then, for the minimizer $\hat{A}_t$ of \eqref{eqn:dsm_loss_rfm_minf}, for $\alpha\in\{\parallel,\perp\}$, in the limit $d,n,p\to\infty$, the test and train errors can be written as:
    %Suppose $P_0\equiv\cN{0,\cC}$. Let $X=[x_1,x_2,\cdots,x_n]$ be the data matrix with its columns drawn i.i.d. from $P_0$. Let $a_t, h_t, \lambda, \varrho, W_t$ be the entities is (\ref{eqn:dsm_loss_rfm_minf}). Let $\varrho$ satisfies Assumption~\ref{assmptn:activation_fn}. Let $\Sigma_t = a_t^2\cC+h_tI_d$, $\sigma_t^2 = \frac{1}{d}\tr{\Sigma_t}$, and $\nu_t^2 = a_t^2\sigma_t^2+h_t$. Then, for the minimizer of (\ref{eqn:dsm_loss_rfm_minf}) $\hat{A}_t$, as $d,n,p\to\infty$, we can represent the test and trains errors as:
    \begin{align*}
    \cE^{\infty}_{\text{test},\alpha}(\hat{A}_t) &= E_{0,\alpha}-\frac{2\mu_1(\sigma_t)^2}{\sigma_t^2}E_{1,\alpha}+\frac{\mu_1(\sigma_t)^4}{\sigma_t^4}E_{2,\alpha}\\ &\qquad\qquad+\frac{\mu_1(\sigma_t)^2v(\sigma_t)^2}{\sigma_t^2}E_{3,\alpha} +\frac{\mu_1(\sigma_t)^2\mu_0(\sigma_t)^2}{\sigma_t^2}E_{4,\alpha} \;,\\
    \cE^\infty_{\text{train}}(\hat{A}_t)
    &=-\frac{h_t\mu_1^2(\sigma_t)}{\sigma_t^2}(E_{1,\parallel}+E_{1,\perp})-h_t\lambda\frac{\mu_1^2(\sigma_t)}{\sigma_t^2}(E_{3,\parallel}+E_{3,\perp})+1 \;,
\end{align*}
where
    \begin{align*}
    E_{0,\alpha} &= \frac{1}{d}\tr{\Pi_\alpha\Sigma^{-1}}\;,\\
    E_{1,\alpha} &= \frac{1}{d}\tr{\Pi_\alpha\rat{W_t^T}{d}(U+\lambda I_p)^{-1}\rat{W_t}{d}\Pi_\alpha}\;,\\
    E_{2,\alpha} &= \frac{1}{d}\tr{\Pi_\alpha\rat{W_t^T}{d}(U+\lambda I_p)^{-1}\rat{W_t}{d}\Sigma\rat{W_t^T}{d}(U+\lambda I_p)^{-1}\rat{W_t}{d}\Pi_\alpha}\;,\\
    E_{3,\alpha} &= \frac{1}{d}\tr{\Pi_\alpha\rat{W_t^T}{d}(U+\lambda I_p)^{-2}\rat{W_t}{d}\Pi_\alpha}\;,\\
    E_{4,\alpha} &= \frac{1}{d}\bone_p^T (U+\lambda I_p)^{-1}\rat{W_t}{d}\Pi_\alpha\rat{W_t^T}{d}(U+\lambda I_p)^{-1}\bone_p\;,
\end{align*}
with 
\begin{equation*}
    U = \rat{G}{n}\rat{G^T}{n}+\frac{h_t}{\sigma_t^2}\mu_1^2(\sigma_t)\rat{W_t}{d}\rat{W_t^T}{d}+s^2I_p \;,
\end{equation*}
\begin{equation*}
    G = \mu_0(\sigma_t)\bone_p\bone_n^T+\frac{a_t}{\sigma_t}\mu_1(\sigma_t)\rat{W_t}{d}X+v_0\Omega\;,
\end{equation*}
\begin{align*}
s^2 = \norm{\varrho(\sigma_t\;\cdot)}^2-c(a_t^2\sigma_0^2/\sigma_t^2,\sigma_t)-\frac{h_t}{\sigma_t^2}\mu_1^2(\sigma_t) \;,
\end{align*}
\begin{align*}
    v_0^2 = c(a_t^2\sigma_0^2/\sigma_t^2,\sigma_t)-\mu_0(\sigma_t)^2-\left(\frac{a_t\sigma_0}{\sigma_t}\mu_1(\sigma_t)\right)^2\;.
\end{align*}

\end{lemma}
\begin{proof}
    We first find the minimizer of (\ref{eqn:dsm_loss_rfm_minf}). When $m=\infty$, the denoising score matching loss is given by 
\begin{align*}
    \cL^\infty_t(A_t) &=\frac{1}{dn}\sum_{i=1}^{n}{\bE{z}{\norm{\sqrt{h_t}\rat{A_t}{p}\act{\rat{W_t}{p}(a_tx_i+\sqrt{h_t}z)}+z}^2}}+\frac{h_t\lambda}{dp}\norm{A_t}_F^2\\
    &=\frac{h_t}{d}\tr{\rat{A_t}{p}^T\rat{A_t}{p} U}+\frac{2\sqrt{h_t}}{d}\tr{\rat{A_t}{p} V}+1+\frac{h_t\lambda}{d}\tr{\rat{A_t^T}{p}\rat{A_t}{p}}\;,\\
\end{align*}
where 
$$U = \frac{1}{n}\sum_{i=1}^n\bE{z}{\act{\rat{W_t}{d}(a_tx_i+\sqrt{h_t}z)}\act{\rat{W_t}{d}(a_tx_i+\sqrt{h_t}z)}^T}\;,$$
and 
$$V = \frac{1}{n}\sum_{i=1}^n\bE{z}{\act{\rat{W_t}{d}(a_tx_i+\sqrt{h_t}z)}z^T}\;.$$ Thus we get the optimal $A_t$ as
\begin{equation}
    \rat{\hat{A}_t}{p} = -\frac{1}{\sqrt{h_t}}V^T(U+\lambda I_p)^{-1}\;.
\end{equation}
We now proceed to the computation of test and train error for $P_0\equiv\cN{0,\cC}$. In this case, $P_t$ is $\cN{0,\Sigma_t}$ and $\nabla\log P_t(x)=-\Sigma_t^{-1}x$, where $\Sigma_t = a_t^2\cC+h_tI_d$.\\\\
%Now, we compute test error (generalization error) and train error when $P_0\equiv\cN{0,\cC}$. We note that in this case, $P_t$ is $\cN{0,\Sigma_t}$ and $\nabla\log P_t(x)=\Sigma_t^{-1}x$, where $\Sigma_t = a_t^2\cC+h_tI_d$.\\\\
\textbf{Test Error:}

For $\alpha\in\{\parallel,\perp\}$, rewriting Eq.~\eqref{eq:At_test_inf_man} for the particular case, we have:
\begin{align*}
    \cE^\infty_{\text{test},\alpha}(\hat{A}_t,\cM) &= \frac{1}{d}\bE{x\sim P_t}{\norm{\Pi_\alpha\rat{\hat{A}_t}{p}\act{\rat{W_t}{d}x}-\Pi_\alpha\nabla\log P_t(x)}^2}\\
    &= \frac{1}{d}\bE{x\sim P_t}{\norm{\Pi_\alpha\rat{\hat{A}_t}{p}\act{\rat{W_t}{d}x}+\Pi_\alpha\Sigma_t^{-1}x}^2}\\
    &=\frac{1}{d}\tr{\Pi_\alpha\Sigma_t^{-1}} - \frac{2}{d}\tr{\frac{1}{\sqrt{h_t}}\Sigma_t^{-1}\Pi_\alpha V^T (U+\lambda I_p)^{-1}\underbrace{\bE{x}{\act{\rat{W_t}{d}x}x^T}}_{:=\Tilde{V}}}\\
    &\qquad+\frac{1}{d}\tr{\frac{1}{h_t}(U+\lambda I_p)^{-1}V\Pi_\alpha V^T(U+\lambda I_p)^{-1}\underbrace{\bE{x}{\act{\rat{W_t}{d}x}\act{\rat{W_t}{d}x}^T}}_{:=\Tilde{U}}}\;.
\end{align*}
Since we focus on a single time instant, we drop the subscript $t$ in the above expressions. However, it is important to note that $a$ and $h$ depend on $t$ and the relation $a^2+h=1$ is valid for all times. Additionally, define $\sigma_j^2 = \Sigma_{jj}$.

We need to compute $V,U,\Tilde{V},\Tilde{U}$ in order to get an expression for $\cE^\infty_{\text{test}}$. We will first consider $\Tilde{V}$. Explicitly:
\begin{equation*}
    \Tilde{V} = \bE{x}{\act{\rat{W}{d}x}x^T}\;.
\end{equation*}
Let $w_i$ denote the $i^{\text{th}}$ row of $W$ and let $x(j)$ denote the $j^{th}$ component of $x$. For large $d$, $\frac{\norm{w_i}^2}{d}$ concentrates to $1$. Note that, conditioned on $W$, $\frac{w_i^Tx}{\sqrt{d}}$ is a zero-mean Gaussian random variable with variance $\bE{x}{\frac{(w_i^Tx)^2}{d}}=\frac{1}{d}w_i^T\Sigma w_i$. By the concentration property of quadratic forms, the variance concentrates to $\sigma^2 = \frac{1}{d}\tr{\Sigma}$ as $d\to\infty$. Thus we can write $(u_0,u_1)=\left(\frac{1}{\sigma}\frac{w_i^Tx}{\sqrt{d}},\frac{x(j)}{\sigma_j}\right)\sim P^{\gamma_{ij}}$, with $\gamma_{ij} = \frac{1}{\sigma\sigma_j}\rat{(W\Sigma)_{ij}}{d}$. 
%Let $w_i$ denote the $i^{\text{th}}$ row of $W$ and let $x(j)$ denote the $j^{th}$ component of $x$. For large $d$, $\frac{\norm{w_i}^2}{d}$ concentrates to $1$. Note that, conditioned on $W$, $\frac{w_i^Tx}{\sqrt{d}}$ is a zero-mean Gaussian random variable with variance $\bE{x}{\frac{(w_i^Tx)^2}{d}}=\frac{1}{d}w_i^T\Sigma w_i$. By the concentration property of quadratic forms, the variance concentrates to $\sigma^2 = \frac{1}{d}\tr{\Sigma}$ as $d\to\infty$. Thus we can write $(u_0,u_1)=\left(\frac{1}{\sigma}\frac{w_i^Tx}{\sqrt{d}},\frac{x(j)}{\sigma_j}\right)\sim P^\gamma$, with $\gamma = \frac{1}{\sigma\sigma_j}\rat{(W\Sigma)_{ij}}{d}$.  Then:
\begin{align*}
    \Tilde{V}_{ij} &= \bE{x}{\act{\rat{w_i^Tx}{d}}x(j)}\\
    &= \bE{(u_0,u_1)\sim P^{\gamma_{ij}}}{\act{\sigma u_0}\sigma_j u_1}\\
    &\stackrel{(a)}{=} \sum_{k=0}^{\infty} \frac{\gamma_{ij}^k}{k!}\bE{u_0}{\act{\sigma u_0}\text{He}_k(u_0)}\sigma_j\bE{u_1}{u_1\text{He}_k(u_1)}\\
    &= \frac{\mu_1(\sigma)}{\sigma}\rat{(W\Sigma)_{ij}}{d} \;,
\end{align*}
%\begin{align*}
%    \Tilde{V}_{ij} &= \bE{x}{\act{\rat{w_i^Tx}{d}}x(j)}\\
%    &= \bE{(u_0,u_1)\sim P^{\frac{1}{\sigma\sigma_j}\rat{(W\Sigma)_{ij}}{d}}}{\act{\sigma u_0}\sigma_j u_1}\\
%    &\stackrel{(a)}{=} \sum_{k=0}^{\infty} \frac{(\frac{1}{\sigma\sigma_j}\rat{(W\Sigma)_{ij}}{d})^k}{k!}\bE{u_0}{\act{\sigma u_0}\text{He}_k(u_0)}\sigma_j\bE{u_1}{u_1\text{He}_k(u_1)}\\
%    &= \frac{\mu_1(\sigma)}{\sigma}\rat{(W\Sigma)_{ij}}{d} \;,
%\end{align*}
where in $(a)$ we used the Mehler Kernel formula \cite{kibble_extension_1945}.  
Hence, $\Tilde{V} = \frac{\mu_1(\sigma)}{\sigma}\rat{W}{d}\Sigma$.

When considering $\Tilde{U}$, a similar argument allows to write $(\tilde{u}_0,\tilde{u}_1)=(\frac{1}{\sigma}\frac{w_i^Tx}{\sqrt{d}},\frac{1}{\sigma}\frac{w_j^Tx}{\sqrt{d}})\sim P^{\tilde{\gamma}_{ij}}$ with $\tilde{\gamma}_{ij} = \frac{1}{\sigma^2}\rat{w_i^T\Sigma w_j}{d}$, leading to 
%Now, we consider $\Tilde{U}$. Proceeding similar to the arguments before, we see that $(u_0,u_1)=(\frac{1}{\sigma}\frac{w_i^Tx}{\sqrt{d}},\frac{1}{\sigma}\frac{w_j^Tx}{\sqrt{d}})\sim P^\gamma$, with $\gamma = \frac{1}{\sigma^2}\rat{w_i^T\Sigma w_j}{d}$. Thus, we get
\begin{align*}
    \Tilde{U}_{ij} &= \bE{x}{\act{\rat{w_i^Tx}{d}}\act{\rat{w_j^Tx}{d}}}\\
    &= \bE{(\tilde{u}_0,\tilde{u}_1)\sim P^{\tilde{\gamma}_{ij}}}{\varrho(\sigma\tilde{u}_0)\varrho(\sigma\tilde{u}_1)}\\
    &= \sum_{k=0}^{\infty} \frac{\tilde{\gamma}_{ij}^k}{k!} \;\bE{\tilde{u}_0}{\varrho(\sigma \tilde{u}_0)\text{He}_k(u_0)}\bE{u_1}{\varrho(\sigma \tilde{u}_1)\text{He}_k(\tilde{u}_1)}\\
    &= \sum_{k=0}^{\infty} \frac{\tilde{\gamma}_{ij}^k}{k!} \; \bE{\tilde{u}_0}{\varrho(\sigma\tilde{u}_0)\text{He}_k(\tilde{u}_0)}^2 \;.
\end{align*}
In short:
\begin{equation*}
    \Tilde{U}_{ij} = \begin{cases}
        \mu_0^2(\sigma)+\frac{\mu_1^2(\sigma)}{\sigma^2} \frac{w_i^T\Sigma w_j}{d} + O(1/d)\quad &\text{if } i\neq j \;,\\
        \norm{\varrho(\sigma\;\cdot)}^2\quad &\text{if } i=j \;.
    \end{cases}
\end{equation*}
The $\bigO{1/d}$ contribution cannot give rise to a $\bigO{1}$ change in the asymptotic spectrum. Hence, we neglect it. We have
\begin{equation*}
    \Tilde{U} = \mu_0^2(\sigma) \bone_p\bone_p^T + \frac{\mu_1^2(\sigma)}{\sigma^2} \rat{W}{d}\Sigma\rat{W^T}{d} + v^2(\sigma) I_p \;,
\end{equation*}
where $v^2(\kappa) = \norm{\rho(\kappa\;\cdot)}^2-\mu_0^2(\kappa)-\mu_1^2(\kappa)$.
We now proceed to $V$. For the $l^{\text{th}}$ data sample, we consider the matrix: 
\begin{equation*}
    V^l = \bE{z}{\act{\rat{W}{d}(ax_l+\sqrt{h}z)}z^T}\;,
\end{equation*}
and write for each of its components:
\begin{align*}
    V^l_{ij} &= \bE{z}{\act{\rat{w_i^T(ax_l+\sqrt{h}z)}{d}}z_j}\\
    &= \bE{(u,v)\sim P^{\rat{w_{ij}}{d}}}{\act{\rat{a w_i^Tx_l}{d}+\sqrt{h}u}v}\\
    &= \sum_{k=0}^{\infty} \frac{(\rat{w_{ij}}{d})^k}{k!}\bE{u}{\act{\rat{a w_i^Tx_l}{d}+\sqrt{h}u}\text{He}_k(u)}\bE{v}{v\text{He}_k(v)}\\
    &= \rat{w_{ij}}{d}\bE{u}{\act{\rat{a w_i^Tx_l}{d}+\sqrt{h}u}u}\\
    &= \rat{w_{ij}}{d}\varrho_1\left(\rat{a w_i^Tx_l}{d}\right)\;,
\end{align*}
where $\varrho_1(y) = \bE{u}{\varrho(y+\sqrt{h}u)u}$. Summing over the $n$ data samples:
\begin{align*}
    V_{ij} &= \frac{1}{n}\sum_{l=1}^{n} V^l_{ij}\\
    &= \rat{w_{ij}}{d} \frac{1}{n}\sum_{l=1}^{n} \varrho_1\left(\rat{a w_i^Tx_l}{d}\right)\\
    &= \rat{w_{ij}}{d} \bE{x}{\varrho_1\left(\rat{a w_i^Tx}{d}\right)} + O(1/d)\\
    &= \rat{w_{ij}}{d} \bE{g}{\varrho_1(a\sigma_0 g)} + O(1/d)\\
    &= \rat{w_{ij}}{d} \bE{g,u}{\varrho(a\sigma_0 g+\sqrt{h}u)u} + O(1/d)\\
    &= \frac{\sqrt{h}}{\sigma}\mu_1(\sigma)\rat{w_{ij}}{d} + O(1/d)\;,
\end{align*}
where we recall that $\sigma_0^2 = a^2\frac{1}{d}\tr{\cC}+h$.
Neglecting $O(1/d)$ terms, we have $V = \frac{\sqrt{h}}{\sigma}\mu_1(\sigma)\rat{W}{d}$.

Going forward to $U$, we again consider the matrix relative to the $l^{\text{th}}$ data sample:
\begin{align*}
    U^l &= \bE{z}{\act{\rat{W}{d}(ax_l+\sqrt{h}z)}\act{\rat{W}{d}(ax_l+\sqrt{h}z)}^T}\;.
\end{align*}
For the components with $i\neq j $, we have:
\begin{align*}
    U^l_{ij} &= \bE{z}{\act{\rat{w_i^T(ax_l+\sqrt{h}z)}{d}}\act{\rat{w_j^T(ax_l+\sqrt{h}z)}{d}}}\\
    &= \bE{(u,v)\sim P^{\frac{w_i^Tw_j}{d}}}{\act{a\rat{w_i^Tx_l}{d} + \sqrt{h}u}\act{a\rat{w_j^Tx_l}{d} + \sqrt{h}v}}\\
    &= \sum_{k=0}^{\infty} \frac{(\frac{w_i^Tw_j}{d})^k}{k!}\bE{u}{\act{a\rat{w_i^Tx_l}{d} + \sqrt{h}u}\text{He}_k(u)}\bE{v}{\act{a\rat{w_j^Tx_l}{d} + \sqrt{h}v}\text{He}_k(v)}\\
    &= \varrho_0\left(a\rat{w_i^Tx_l}{d}\right)\varrho_0\left(a\rat{w_j^Tx_l}{d}\right)+\frac{w_i^Tw_j}{d}\varrho_1\left(a\rat{w_i^Tx_l}{d}\right)\varrho_1\left(a\rat{w_j^Tx_l}{d}\right)+O(1/d)\;,
\end{align*}
%where $\varrho_0(y) = \bE{u}{\varrho(y+\sqrt{h}u)}$ and $\varrho_1(y) = \bE{u}{\varrho(y+\sqrt{h}u)u}$. Let $X=[x_1, x_2,\cdots,x_n]\in\R^{d\times n}$:
where $\varrho_0(y) = \bE{u}{\varrho(y+\sqrt{h}u)}$ and $\varrho_1(y) = \bE{u}{\varrho(y+\sqrt{h}u)u}$. Summing over the $n$ data samples:
\begin{align*}
    U_{ij} &= \frac{1}{n}\sum_{l=1}^{n} U^l_{ij}\\
    &= \frac{1}{n}\sum_{l=1}^{n}\varrho_0\left(a\rat{w_i^Tx_l}{d}\right)\varrho_0\left(a\rat{w_j^Tx_l}{d}\right)+\frac{w_i^Tw_j}{d}\frac{1}{n}\sum_{l=1}^{n}\varrho_1\left(a\rat{w_i^Tx_l}{d}\right)\varrho_1\left(a\rat{w_j^Tx_l}{d}\right) + O(1/d)\\
    &= \frac{1}{n}\sum_{l=1}^{n}\varrho_0\left(a\rat{w_i^Tx_l}{d}\right)\varrho_0\left(a\rat{w_j^Tx_l}{d}\right)+\frac{w_i^Tw_j}{d}\bE{x}{\varrho_1\left(a\rat{w_i^Tx}{d}\right)\varrho_1\left(a\rat{w_j^Tx}{d}\right)} + O(1/d)\\
    &= \frac{1}{n}\sum_{l=1}^{n}\varrho_0\left(a\rat{w_i^Tx_l}{d}\right)\varrho_0\left(a\rat{w_j^Tx_l}{d}\right)+\frac{w_i^Tw_j}{d}\bE{g}{\varrho_1(a\sigma_0 g)}^2 + O(1/d)\\
    &= \frac{1}{n}\sum_{l=1}^{n}\varrho_0\left(a\rat{w_i^Tx_l}{d}\right)\varrho_0\left(a\rat{w_j^Tx_l}{d}\right)+\frac{h}{\sigma^2}\mu_1^2(\sigma)\frac{w_i^Tw_j}{d} + O(1/d)\;.\\
\end{align*}
For $i=j$, we have:
\begin{align*}
    U^l_{ii} &= \bE{z}{\left(\act{\rat{w_i^T(ax_l+\sqrt{h}z)}{d}}\right)^2} \;,\\
\end{align*}
and
\begin{align*}
    U_{ii} &= \frac{1}{n}\sum_{l=1}^n\bE{z}{\left(\act{\rat{w_i^T(ax_l+\sqrt{h}z)}{d}}\right)^2}\\
    &= \bE{z,x}{\left(\act{\rat{w_i^T(ax+\sqrt{h}z)}{d}}\right)^2}+O(1/\sqrt{d})\\
    &= \norm{\varrho(\sigma\;\cdot)}^2+O(1/\sqrt{d}) \;.
\end{align*}
The $\bigO{1/\sqrt{d}}$ term in the above equation can be neglected, since there are only $\bigO{d}$ terms on the diagonal. Let $X=[x_1,x_2,\cdots,x_n]\in\R^{d\times n}$. We can write $U$ as:
\begin{equation*}
    U = \rat{\varrho_0\left(a\rat{W}{d}X\right)}{n}\rat{\varrho_0\left(a\rat{W}{d}X\right)^T}{n}+\frac{h}{\sigma^2}\mu_1^2(\sigma)\rat{W}{d}\rat{W^T}{d}+s^2I_p \;,
\end{equation*}
where
\begin{align*}
s^2 &= \norm{\varrho(\sigma\;\cdot)}^2-\bE{g}{\varrho_0(a\sigma_0 g)^2}-\frac{h}{\sigma^2}\mu_1^2(\sigma)\\
&= \norm{\varrho(\sigma\;\cdot)}^2-\bE{g}{\bE{u}{\varrho(a\sigma_0 g+\sqrt{h}u)}^2}-\frac{h}{\sigma^2}\mu_1^2(\sigma)\\
&= \norm{\varrho(\sigma\;\cdot)}^2-c(a^2\sigma_0^2/\sigma^2,\sigma)-\frac{h}{\sigma^2}\mu_1^2(\sigma) \;,
\end{align*}
with $c(\gamma,\kappa) = \bE{u,v\sim P^\gamma}{\varrho(\kappa u)\varrho(\kappa v)}$.

We can use the Gaussian equivalence principle to replace the nonlinear term in $U$. A Gaussian equivalent for $\varrho_0(a\rat{W}{d}X)$ is given by
\begin{align*}
    G &= \bE{g}{\varrho_0(a\sigma_0 g)}\bone_p\bone_n^T+\bE{g}{\varrho_0(a\sigma _0g)g}\rat{W}{d}\frac{X}{\sigma_0}\\
    &\qquad\qquad+\left(\bE{g}{\varrho_0(a\sigma_0 g)^2}-\bE{g}{\varrho_0(a\sigma_0 g)}^2-\bE{g}{\varrho_0(a\sigma _0g)g}^2\right)^{1/2}\Omega\\
    &= \mu_0(\sigma)\bone_p\bone_n^T+a\frac{\mu_1(\sigma)}{\sigma}\rat{W}{d}X+\left(\underbrace{c(a^2\sigma_0^2/\sigma^2,\sigma)-\mu_0(\sigma)^2-\left(\frac{a\sigma_0}{\sigma}\mu_1(\sigma)\right)^2}_{:=v_0^2}\right)^{1/2}\Omega\;,
\end{align*}
where $\Omega\in\R^{p\times n}$ is a random matrix with standard Gaussian entries.
Hence we have
\begin{equation*}
    U = \rat{G}{n}\rat{G^T}{n}+\frac{h}{\sigma^2}\mu_1^2(\sigma)\rat{W}{d}\rat{W^T}{d}+s^2I_p \;,
\end{equation*}
with
\begin{equation*}
    G = \mu_0(\sigma)\bone_p\bone_n^T+a\frac{\mu_1(\sigma)}{\sigma}\rat{W}{d}X+v_0\Omega\;.
\end{equation*}
We now have expressions for all terms terms contributing to the test error:
\begin{align*}
    \cE^\infty_{\text{test},\alpha}(\hat{A}_t) &=\frac{1}{d}\tr{\Pi_\alpha\Sigma^{-1}} - \frac{2}{d}\tr{\frac{1}{\sqrt{h}}\Sigma^{-1}\Pi_\alpha V^T (U+\lambda I_p)^{-1}\Tilde{V}}\;\\&\qquad\qquad+\frac{1}{d}\tr{\frac{1}{h_t}(U+\lambda I_p)^{-1}V \Pi_\alpha V^T(U+\lambda I_p)^{-1}\Tilde{U}} \\
    &= \frac{1}{d}\tr{\Pi_\alpha\Sigma^{-1}} -\frac{2\mu^2_1(\sigma)}{\sigma^2}\frac{1}{d}\tr{\Pi_\alpha\rat{W^T}{d}(U+\lambda I_p)^{-1}\rat{W}{d}}\\&\qquad+\frac{\mu_1^2(\sigma)}{\sigma^2}\frac{1}{d}\text{tr}\bigg\{(U+\lambda I_p)^{-1}\rat{W}{d}\Pi_\alpha\rat{W^T}{d}(U+\lambda I_p)^{-1}\\&\qquad\qquad\qquad\left(\mu_0(\sigma)^2 \bone_p\bone_p^T + \frac{\mu_1(\sigma)^2}{\sigma^2} \rat{W}{d}\Sigma\rat{W^T}{d} + v(\sigma)^2 I_p\right)\bigg\}\;.
\end{align*}
We can then write
\begin{align*}
    \cE^\infty_{\text{test},\alpha}(\hat{A}_t) &= E_{0,\alpha}-\frac{2\mu_1^2(\sigma)}{\sigma^2}E_{1,\alpha}+\frac{\mu_1^4(\sigma)}{\sigma^4}E_{2,\alpha}+\frac{\mu_1^2(\sigma)v(\sigma)^2}{\sigma^2}E_{3,\alpha} +\frac{\mu_1(\sigma)^2\mu_0(\sigma)^2}{\sigma^2}E_{4,\alpha} \ \;,
\end{align*}
with
\begin{align*}
    E_{0,\alpha} &= \frac{1}{d}\tr{\Pi_\alpha\Sigma^{-1}}\;,\\
    E_{1,\alpha} &= \frac{1}{d}\tr{\Pi_\alpha\rat{W^T}{d}(U+\lambda I_p)^{-1}\rat{W}{d}\Pi_\alpha}\;,\\
    E_{2,\alpha} &= \frac{1}{d}\tr{\Pi_\alpha\rat{W^T}{d}(U+\lambda I_p)^{-1}\rat{W}{d}\Sigma\rat{W^T}{d}(U+\lambda I_p)^{-1}\rat{W}{d}\Pi_\alpha}\;,\\
    E_{3,\alpha} &= \frac{1}{d}\tr{\Pi_\alpha\rat{W^T}{d}(U+\lambda I_p)^{-2}\rat{W}{d}\Pi_\alpha}\;,\\
    E_{4,\alpha} &= \frac{1}{d}\bone_p^T (U+\lambda I_p)^{-1}\rat{W}{d}\Pi_\alpha\rat{W^T}{d}(U+\lambda I_p)^{-1}\bone_p\;,
\end{align*}

Next, we look at the train error.\\\\
\textbf{Train error:}
\begin{align*}
    \cE^\infty_{\text{train}}(\hat{A}_t) &= \cL(\hat{A}_t)-\frac{h_t\lambda}{pd}\norm{\hat{A}_t}^2_F\\
    &=\frac{h}{d}\tr{\rat{\hat{A}_t^T}{p}\rat{\hat{A}_t}{p} (U+\lambda I_p)}+\frac{2\sqrt{h}}{d}\tr{\rat{\hat{A}_t}{p} V}+1-\frac{h\lambda}{d}\tr{\rat{\hat{A}_t^T}{p}\rat{\hat{A}_t}{p}}\\
    &=\frac{1}{d}\tr{(U+\lambda I_p)^{-1}VV^T}-\frac{2}{d}\tr{V^T(U+\lambda I_p)^{-1}V}+1-\frac{\lambda}{d}\tr{V^T(U+\lambda I_p)^{-2}V}\\
    &=-\frac{h\mu_1^2(\sigma)}{d\sigma^2}\tr{(U+\lambda I_p)^{-1}\rat{W}{d}\rat{W^T}{d}}+1-\frac{h\mu_1^2(\sigma)\lambda}{d\sigma^2}\tr{(U+\lambda I_p)^{-2}\rat{W}{d}\rat{W^T}{d}}\\
    &=-\frac{h\mu_1^2(\sigma)}{\sigma^2}(E_{1,\parallel}+E_{1,\perp})-h\lambda\frac{\mu_1^2(\sigma)}{\sigma^2}(E_{3,\parallel}+E_{3,\perp})+1 \;.\\
\end{align*}
This completes the proof of lemma \ref{lemma:lc_minf}.
\end{proof}

We have the following theorem characterizing test and train errors in the $m=\infty$ case:
\begin{theorem}\label{appendix:thm:lc_minf_man} Let $\cM_\parallel$ be a $D$-dimensional subspace in $\R^D$, $\Pi_\parallel$ a projection matrix onto it, and $P_0\equiv\cN{0,\Pi_\parallel}$ with $\varrho$ satisfying Assumption~\ref{assmptn:activation_fn}. Define $\sigma_0^2 = \frac{1}{d}\tr{\Pi_\parallel}$, $\sigma_t^2 = a_t^2\sigma_0^2+h_t$, and $\mu_{1,t} = \mu_1(\sigma_t)/\sigma_t$. Set $s^2 = \norm{\varrho(\sigma_t\cdot)}^2-c(a_t^2\sigma_0^2/\sigma_t^2,\sigma_t)-h_t\mu_{1,t}^2$,\; $v_0^2=c(a_t^2\sigma_0^2/\sigma_t^2,\sigma)-a_t^2\sigma_0^2\mu_{1,t}^2$ and $v^2 = \norm{\varrho(\sigma_t\cdot)}^2-\mu_1(\sigma_t)^2$. Let $\psi_D = \frac{D}{d}, \psi_n = \frac{n}{d}$, $\psi_p = \frac{p}{d}$ and $\zeta_1,\zeta_2,\zeta_3,\zeta_4,\zeta_5$ be the solution of the following system of algebraic equations in $q$ and $z$:
    \begin{align*}
    \zeta_1(s^2-z+(1-\psi_D)h_t(\mu_{1,t}^2+q)\zeta_2+\psi_D(h_t\mu_{1,t}^2+q)\zeta_{33}+\psi_Da_t^2\mu_{1,t}^2\zeta_3\zeta_4+v_0^2\zeta_4) -1 &= 0 \;,\nonumber\\
    \zeta_2(1+\psi_ph_t(\mu_{1,t}^2+q)\zeta_1) - 1  &= 0 \;,\nonumber\\
    \zeta_3(1+\psi_p(h_t\mu_{1,t}^2+q)\zeta_1)+\psi_pa_t^2\mu_{1,t}^2\zeta_1\zeta_3\zeta_4 -1 &= 0 \;,\nonumber\\
    \zeta_5(1+\psi_p(h_t\mu_{1,t}^2+q)\zeta_1)+(1+a\mu_{1,t}\zeta_4\zeta_5)\psi_pa_t\mu_{1,t}\zeta_1 &= 0 \;,\nonumber\\
    \zeta_4\left(1+\frac{\psi_p}{\psi_n}v_0^2\zeta_1-\frac{\psi_D}{\psi_n}a_t\mu_{1,t}\zeta_5\right) -1 &= 0\;.\nonumber
    \end{align*}
    Let $e_{0,\parallel} = \psi_D$ and $e_{0,\perp} = \frac{1-\psi_D}{h_t}$. Define functions $\cK_\parallel(q,z)=-\frac{\psi_D\zeta_5(q,z)}{a_t\mu_{1,t}}$ and $\cK_\perp(q,z)=\frac{(1-\psi_D)(1-\zeta_2(q,z))}{h_t(\mu_{1,t}^2+q)}$. For $\alpha\in\{\parallel,\perp\}$,
    let $\varepsilon^{\infty}_{\text{test},\alpha} =e_{0,\alpha}-2\mu_{1,t}^2\cK_\alpha(0,-\lambda)-\mu_{1,t}^4\frac{\partial \cK_\alpha}{\partial q}(0,-\lambda)+\mu_{1,t}^2v^2\frac{\partial \cK_{\alpha}}{\partial z}(0,-\lambda)$, and $\varepsilon^\infty_{\text{train}} = 1-\mu_{1,t}^2h_t(\cK_\parallel(0,-\lambda)+\cK_\perp(0,-\lambda))-\mu_{1,t}^2\lambda h_t(\frac{\partial \cK_\parallel}{\partial z}(0,-\lambda)+\frac{\partial \cK_\perp}{\partial z}(0,-\lambda))$. Then, for the minimizer of (\ref{eqn:dsm_loss_rfm_minf}) $\hat{A}_t$, as $d,n,p\to\infty$:
    \begin{align*}
        \lim_{d,n,p\to\infty}\bE{}{\cE^\infty_{\text{test},\parallel}(\hat{A}_t)} =  \varepsilon^\infty_{\text{test},\parallel}\;,\quad \lim_{d,n,p\to\infty}\bE{}{\cE^\infty_{\text{test},\perp}(\hat{A}_t)} =  \varepsilon^\infty_{\text{test},\perp}\;,\quad
    \lim_{d,n,p\to\infty}\bE{}{\cE^\infty_{\text{train}}(\hat{A}_t)} &= \varepsilon^\infty_{\text{train}} \;.
    \end{align*}
\end{theorem}
\begin{remark}
    Theorem~\ref{thm:lc_minf} can be recovered by substituting $\psi_D=1$.
\end{remark}
\begin{proof}
First, for $P_0 \equiv \cN{0,\cC}$, we use the following Lemma~\ref{lemma:lc_minf} to express the test and train errors for $m=\infty$ as sums of traces of rational functions of random matrices. This will facilitate the use of linear pencils to compute the asymptotic test and train errors. 

To prove Theorem~\ref{thm:lc_minf}, we begin by noting that, that without loss of generality, we may assume that $\cC = \Pi_\parallel$ is a diagonal matrix with its first $D$ diagonal entries equal to $1$ and the remaining entries equal to $0$. That is, 
\begin{equation}
    \cC = \Pi_\parallel = \text{diag}([\underbrace{1,1,\cdots,1}_{D},0,0\cdots,0]) \;.
\end{equation}
This simplification follows from the rotational invariance of the problem. We now invoke Lemma~\ref{lemma:lc_minf} for this value of $\cC$. As in the proof of the Lemma, since we focus on a single time instant, we drop the subscript $t$ in the above expressions. For the $\cC$ used here, note that $\sigma^2 = a^2\psi_D+h, \sigma_0^2 =  \psi_D$. Also, let $\mu_0 = \mu_0(\sigma), \mu_1 = \mu_1(\sigma)/\sigma, \norm{\varrho(\sigma\;\cdot)} = \norm{\varrho}$.

Let the first $D$ columns of $W$ be denoted by $W_\parallel$ and $W_\perp$ the remaining columns, i.e., $W = [W_\parallel, W_\perp]$. For simplicity in presentation, we assume $\mu_0=0$ and write:
\begin{align*}
    \cE^\infty_{\text{test},\alpha}(\hat{A}_t) &= E_{0,\alpha}-2\mu_1^2E_{1,\alpha}+\mu_1^4E_{2,\alpha}+\mu_1^2v^2E_{3,\alpha} \;,\\
    \cE^\infty_{\text{train}}(\hat{A}_t)
    &=-h\mu_1^2(E_{1,\parallel}+E_{1,\perp})-h\lambda\mu_1^2(E_{3,\parallel}+E_{3,\perp})+1 \;,
\end{align*}
with
\begin{align*}
    E_{0,\parallel} &= \psi_D, \quad E_{0,\perp} = \frac{1-\psi_D}{h}\;, \\
    E_{1,\alpha} &= \frac{1}{d}\tr{\rat{W_\alpha^T}{d}(U+\lambda I_p)^{-1}\rat{W_\alpha}{d}}\;, \\
    E_{2,\alpha} &= \frac{1}{d}\tr{\rat{W_\alpha^T}{d}(U+\lambda I_p)^{-1}\rat{W}{d}\Sigma\rat{W^T}{d}(U+\lambda I_p)^{-1}\rat{W_\alpha}{d}}\;,\\
    E_{3,\alpha} &= \frac{1}{d}\tr{\rat{W_\alpha^T}{d}(U+\lambda I_p)^{-2}\rat{W_\alpha}{d}}\;,
\end{align*}
for $\alpha\in\{\parallel,\perp\}$. We now define the matrix:
\begin{align*}
    U(q) &:=  \rat{G}{n}\rat{G^T}{n}+h\mu_1^2\rat{W}{d}\rat{W^T}{d}+q\rat{W}{d}\Sigma\rat{W^T}{d}+s^2I_p \\
    &= \rat{G}{n}\rat{G^T}{n}+(h\mu_1^2+q)\rat{W_\parallel}{d}\rat{W_\parallel^T}{d}+(h\mu_1^2+hq)\rat{W_\perp}{d}\rat{W_\perp^T}{d}+s^2I_p \;,
\end{align*}
and consider its resolvent:
\begin{equation*}
    R(q,z) = (U(q)-zI_p)^{-1}\;.
\end{equation*}
Note that for our special choice of $\cC$, the matrix $G$ takes the form:
%\begin{align*}
%    G &= a\mu_1\rat{W}{d}X+v_0\Omega\;,\\
%    &= a\mu_1\rat{W_\parallel}{d}X_\parallel+v_0\Omega\;,
%\end{align*}
\begin{equation*}
      G = a\mu_1\rat{W}{d}X+v_0\Omega = a\mu_1\rat{W_\parallel}{d}X_\parallel+v_0\Omega\;,
\end{equation*}
where $X_\parallel$ denotes the first $D$ rows of $X$. Define
\begin{equation*}
    K_\alpha(q,z) = \frac{1}{d}\tr{\rat{W_\alpha^T}{d}R(q,z)\rat{W_\alpha}{d}}\;.
\end{equation*}
Using the matrix derivative identities $\frac{\partial R}{\partial q} = -R(q,z)\frac{\dd U}{\dd q}R(q,z)$ and $\frac{\partial R}{\partial z} = R(q,z)^2$, we observe that 
\begin{align*}
    E_{1,\alpha} &= K_\alpha(0,-\lambda) \;,\\
    E_{2,\alpha}  &= -\frac{\partial K_\alpha}{\partial q}(0,-\lambda)\;,\\
    E_{3,\alpha} &= \frac{\partial K_\alpha}{\partial z}(0,-\lambda)\;.
\end{align*}
Therefore, assuming we can take limit of the expectation inside the derivative, it suffices to consider the function $\cK_\alpha(q,z):=\lim_{d\to\infty}\bE{}{K_\alpha(q,z)}$ to have an expression for $\lim_{d\to\infty}\bE{}{\cE^\infty_{\text{test},\alpha}(\hat{A}_t)}$. We then have
\begin{align}\label{eqn:Etest_minf}
    \lim_{d\to\infty}\bE{}{\cE^\infty_{\text{test},\alpha}(\hat{A}_t)} &= E_{0,\alpha}-2\mu_1^2\lim_{d\to\infty}\bE{}{E_{1,\alpha}}+\mu_1^4\lim_{d\to\infty}\bE{}{E_{2,\alpha}}+\mu_1^2v^2\lim_{d\to\infty}\bE{}{E_{3,\alpha}} \nonumber\\
    &= E_{0,\alpha}-2\mu_1^2\cK_\alpha(0,-\lambda)+\mu_1^4\frac{\partial \cK_\alpha}{\partial q}(0,-\lambda)+\mu_1^2v^2\frac{\partial \cK_\alpha}{\partial z}(0,-\lambda) \nonumber\\
    &= E_{0,\alpha}-2\mu_1^2e_{1,\alpha}+\mu_1^4e_{2,\alpha}+\mu_1^2v^2e_{3,\alpha} \;,
\end{align}
where $e_{1,\alpha} = \cK_\alpha(0,-\lambda),\; e_{2,\alpha} = -\frac{\partial \cK_\alpha}{\partial q}(0,-\lambda),\; e_{3,\alpha} = \frac{\partial \cK_\alpha}{\partial z}(0,-\lambda)$. The above computations can also be carried out without interchanging derivatives and expectations. This can be done by first differentiating $K_\alpha(q,z)$ (instead of $\mathcal{K_\alpha}(q, z)$), and then constructing a linear pencil matrix that is nearly twice the size of the one used here. Since this approach is more tedious, we opt to work with the smaller linear pencil below. 

We now derive an expression for $\cK_\alpha$ using the Linear Pencils method. To begin, we construct the following $5\times 5$ block matrix:
\begin{equation*}
    L = \mleft[
        \begin{array}{c|cccc}
        (s^2-z)I_p & (h\mu_1^2+hq)\rat{W_2}{d} & (h\mu_1^2+q)\rat{W_1}{d} & v_0\rat{\Omega}{n} & a\mu_1\rat{W_1}{d}\\
        \hline
        -\rat{W_2^T}{d} & I_{d-D} & 0 & 0 & 0\\
        -\rat{W_1^T}{d} & 0 & I_D & 0 & 0\\
        -v_0\rat{\Omega^T}{n} & 0 & -a\mu_1\rat{X_1^T}{n} & I_n & 0\\
        0 & 0 & 0 & -\rat{X_1}{d} & I_D
    \end{array}
    \mright] = \mleft[\begin{array}{cc}
        L_{11} & L_{12} \\
        L_{21} & L_{22}
    \end{array}\mright] \;.
\end{equation*}
First, we can invert $L$ and verify that $K_\alpha$ can be obtained from the trace of one of the blocks in $L^{-1}$. In particular, we observe that $(L^{-1})^{2,2} = I - (h\mu_1^2+hq)\rat{W_\perp^T}{d}R(q,z)\rat{W_\perp}{d}$ and $(L^{-1})^{3,5} = -a\mu_1\rat{W_\parallel^T}{d}R(q,z)\rat{W_\parallel}{d}$, which yield the desired terms. We use the linear pencil formalism to derive the traces of the square blocks in $L^{-1}$. Let $g$ be the matrix of traces of square blocks in $L^{-1}$ divided by the block size. For example, if $L^{i,j}$ is a square matrix of dimension $N$, then $g_{ij}=\frac{1}{N}\tr{(L^{-1})^{i,j}}$. If $L^{i,j}$ is not a square matrix, we set $g_{ij}=0$. In our setting, this gives:
\begin{equation*}
    g = \mleft[
    \begin{array}{ccccc}
        g_{11} & 0 & 0 & 0 & 0  \\
        0 & g_{22} & 0 & 0 & 0 \\
        0 & 0 & g_{33} & 0 & g_{35}  \\
        0 & 0 & 0 & g_{44} & 0\\
        0 & 0 & g_{53} & 0 & g_{55}  \\
    \end{array}
    \mright]\;.
\end{equation*}
Notice that all constant matrices appearing in $L$ are multiples of identity. We can therefore encode their coefficients in a matrix $B$, given by
\begin{equation*}
    B = \mleft[
    \begin{array}{ccccc}
        s^2-z & 0 & 0 & 0 & 0  \\
        0 & 1 & 0 & 0 & 0  \\
        0 & 0 & 1 & 0 & 0 \\
        0 & 0 & 0 & 1 & 0 \\
        0 & 0 & 0 & 0 & 1
    \end{array}
    \mright]\;.
\end{equation*}
If $L^{il}$ and $L^{jk}$ are square matrices, let $\sigma_{ij}^{kl}$ denote the covariance between an element of $L^{ij}$ and an element of $L^{kl}$ multiplied by the block size of $L^{jk}$. Let $L^{ij}$ be of dimension $N_i\times N_j$ and $M$ denote the non-constant part of $L$. Then,
\begin{equation*}
    \sigma_{ij}^{kl} = N_j\bE{}{M^{ij}_{uv}M^{kl}_{vu}}\;.
\end{equation*}
Let $S = \{(i,j): N_i=N_j\}$ be the set of indices corresponding to square blocks in $L$. Then, a mapping $\eta_L$ is defined such that
\begin{equation*}
    \eta_L(G)_{il} = \sum_{(jk) \in S} \sigma_{ij}^{kl}g_{jk}\;,
\end{equation*}
for $(il)$ in $S$. In our case, this yields:
\begin{equation*}
    \eta_L(g) = \mleft[
    \begin{array}{ccccc}
        \sigma_{12}^{21}g_{22}+\sigma_{13}^{31}g_{33}+\sigma_{15}^{31}g_{53}+\sigma_{14}^{41}g_{44} & 0 & 0 & 0 & 0 \\
        0 & \sigma_{12}^{21}g_{11} & 0 & 0 & 0\\
        0 & 0 & \sigma_{31}^{13}g_{11} & 0 & \sigma_{31}^{15}g_{11}  \\
        0 & 0 & 0 & \sigma_{41}^{14}g_{11}+\sigma_{43}^{54}g_{35} & 0 \\ 
        0 & 0 & \sigma_{54}^{43}g_{44} & 0 & 0  \\
    \end{array}
    \mright]\;.
\end{equation*}
The block dimensions are $N_1=p, N_2=d-D, N_3=D, N_4=n, N_5=D$, and the non-zero covariances are
%\begin{align}
%    & \sigma_{12}^{41} = -a\mu_1 \;,\qquad\qquad\;\;\sigma_{41}^{12}= -a\mu_1\psi_p \;,\\
%    & \sigma_{13}^{31} = -v_0^2 \;, \qquad\qquad\;\;\;\;\sigma_{31}^{13} = -v_0^2\psi_p/\psi_n\;,\\
%    & \sigma_{14}^{41} = -(h\mu_1^2+q) \;,\qquad \sigma_{41}^{14} = -(h\mu_1^2+q)\psi_p\;,\\
%    & \sigma_{23}^{34} = a\mu_1 \;, \qquad\qquad\;\;\;\;\;\sigma_{34}^{23}=a\mu_1/\psi_n \;.
%\end{align}
\begin{alignat}{2}
  &\sigma_{12}^{21} = -(1-\psi_D)(h\mu_1^2+hq)       \;,&&\quad\quad \sigma_{21}^{12} = -\psi_p(h\mu_1^2+hq)          \;,\\
  &\sigma_{13}^{31} = -\psi_D(h\mu_1^2+q)        \;,&&\quad\quad \sigma_{31}^{13} = -\psi_p(h\mu_1^2+q)  \;,\\
  &\sigma_{14}^{41} = -v_0^2 \;,&&\quad\quad \sigma_{41}^{14} = -\frac{\psi_p}{\psi_n}v_0^2   \;,\\
  &\sigma_{15}^{31} = -\psi_Da\mu_1        \;,&& \quad\quad \sigma_{31}^{15} = -\psi_pa\mu_1         \;,\\
  &\sigma_{43}^{54} = \frac{\psi_D}{\psi_n}a\mu_1        \;,&& \quad\quad \sigma_{54}^{43} =a\mu_1         \;.
\end{alignat}
Finally, the matrix $g$ satisfies the fixed point equation
\begin{equation*}
    (B-\eta_L(g))g = I \;,
\end{equation*}
which explicitly reads:
\begin{multline*}
    \mleft[
    \begin{array}{ccccc}
        \begin{smallmatrix}s^2-z+\\(1-\psi_D)(h\mu_1^2+hq)g_{22}+\\\psi_D(h\mu_1^2+q)g_{33}+\psi_Da\mu_1g_{53}+\\v_0^2g_{44}\end{smallmatrix} & 0 & 0 & 0 & 0  \\
        0 & 1+\psi_p(h\mu_1^2+hq)g_{11} & 0 & 0 & 0  \\
        0 & 0 & 1+\psi_p(h\mu_1^2+q)g_{11} & 0  & \psi_pa\mu_1g_{11} \\
        0 & 0 & 0 & \begin{smallmatrix}
            1+\frac{\psi_p}{\psi_n}v_0^2g_{11}-\\\frac{\psi_D}{\psi_n}a\mu_1g_{35}
        \end{smallmatrix} & 0 \\
        0 & 0 & -a\mu_1g_{44} & 0 & 1
    \end{array}
    \mright]\\
    \times
    \mleft[
    \begin{array}{ccccc}
        g_{11} & 0 & 0 & 0 & 0  \\
        0 & g_{22} & 0 & 0 & 0 \\
        0 & 0 & g_{33} & 0 & g_{35}  \\
        0 & 0 & 0 & g_{44} & 0\\
        0 & 0 & g_{53} & 0 & g_{55}  \\
    \end{array}
    \mright] = I \;.
\end{multline*}
This leads to the following set of equations:
\begin{align*}
    g_{11}(s^2-z+(1-\psi_D)(h\mu_1^2+hq)g_{22}+\psi_D(h\mu_1^2+q)g_{33}+\psi_Da\mu_1g_{53}+v_0^2g_{44}) &= 1 \;,\\
    g_{22}(1+\psi_p(h\mu_1^2+hq)g_{11}) &= 1 \;,\\
    g_{33}(1+\psi_p(h\mu_1^2+q)g_{11})+g_{53}\psi_pa\mu_1g_{11} &= 1 \;,\\
    g_{35}(1+\psi_p(h\mu_1^2+q)g_{11})+g_{55}\psi_pa\mu_1g_{11} &= 0 \;,\\
    g_{44}(1+\frac{\psi_p}{\psi_n}v_0^2g_{11}-\frac{\psi_D}{\psi_n}a\mu_1g_{35}) &= 1 \;,\\
    -g_{33}a\mu_1g_{44}+g_{53} &= 0 \;,\\
    -g_{35}a\mu_1g_{44}+g_{55} &= 1 \;.
\end{align*}
This system of seven equations can be reduced to the following five equations:
\begin{align*}
    \zeta_1(s^2-z+(1-\psi_D)h(\mu_1^2+q)\zeta_5+\psi_D(h\mu_1^2+q)\zeta_{4}+\psi_Da^2\mu_1^2\zeta_4\zeta_2+v_0^2\zeta_2) -1 &= 0 \;,\nonumber\\
    \zeta_4(1+\psi_p(h\mu_1^2+q)\zeta_1)+\psi_pa^2\mu_1^2\zeta_1\zeta_4\zeta_2 -1 &= 0 \;,\nonumber\\
    \zeta_3(1+\psi_p(h\mu_1^2+q)\zeta_1)+(1+a\mu_1\zeta_2\zeta_3)\psi_pa\mu_1\zeta_1 &= 0 \;,\nonumber\\
    \zeta_2(1+\frac{\psi_p}{\psi_n}v_0^2\zeta_1-\frac{\psi_D}{\psi_n}a\mu_1\zeta_3) -1 &= 0\;,\nonumber\\
    \zeta_5(1+\psi_ph(\mu_1^2+q)\zeta_1) - 1  &= 0 \;,\nonumber\\
    \end{align*}
where $\zeta_1=g_{11},\; \zeta_2=g_{44},\; \zeta_3=g_{35},\; \zeta_4=g_{33},\; \zeta_5=g_{22}$. We solve this system numerically to obtain $\zeta_1,\zeta_2,\zeta_3,\zeta_4,$ and $\zeta_5$. We compute $\cK_\alpha = \lim_{d\to\infty}\bE{}{K_\alpha}$ by using the expressions $\cK_\parallel(q,z)=-\frac{\psi_D\zeta_3(q,z)}{a\mu_1}$ and $\cK_\perp(q,z)=\frac{(1-\psi_D)(1-\zeta_5(q,z))}{h\mu_1^2+hq}$. Finally, we use \eqref{eqn:Etest_minf} to obtain $\lim_{d\to\infty}\bE{}{\cE^\infty_{\text{test},\alpha}(\hat{A}_t)}$.

Next, we compute the train error. From the above:\\\\
%\textbf{Train error:}
\begin{align*}
    \cE^\infty_{\text{train}}(\hat{A}_t) &= -h\mu_1^2(E_{1,\parallel}+E_{1,\perp})-h\lambda \mu_1^2(E_{3,\parallel}+E_{3,\perp})+1 \;,\\
    &=-h\mu_1^2(K_\parallel(0,-\lambda)+K_\perp(0,-\lambda))-h\lambda\mu_1^2(\frac{\partial K_\parallel}{\partial z}(0,-\lambda)+\frac{\partial K_\perp}{\partial z}(0,-\lambda))+1 \;.\\
\end{align*}
Thus,
\begin{align}\label{eqn:Etrain_minf}
    \lim_{d\to\infty}\bE{}{\cE^\infty_{\text{train}}(\hat{A}_t)} &= -h\mu_1^2(\cK_\parallel(0,-\lambda)+\cK_\perp(0,-\lambda))-h\lambda\mu_1^2(\frac{\partial \cK_\parallel}{\partial z}(0,-\lambda)+\frac{\partial \cK_\perp}{\partial z}(0,-\lambda))+1 \nonumber\\
    &= 1-h\mu_1^2(e_{1,\parallel}+e_{1,\perp})-h\lambda\mu_1^2(e_{3,\parallel}+e_{3,\perp}) \;,
\end{align}
where $e_{1,\alpha} = \cK_\alpha(0,-\lambda),\; e_{3,\alpha} = \frac{\partial \cK_\alpha}{\partial z}(0,-\lambda)$.
This concludes the proof of Theorem~\ref{thm:lc_minf}.
\end{proof}
\section{PROOF OF THEOREM~\ref{thm:lc_m1}}\label{appndx:lc_m1_proof}

We use Lemma~\ref{lemma:lc_m1} to express the test and train errors for $m=1$ case as sums of traces of rational functions of random matrices when $P_0 \equiv \cN{0,\cC}$. We recall the definition of the following entities: $\mu_0(\kappa) = \bE{g}{\varrho(\kappa g)},\; \mu_1(\kappa) = \bE{g}{\varrho(\kappa g)g},\; \norm{\varrho(\kappa\;\cdot)}^2 = \bE{g}{\varrho(\kappa g)^2}$, $v^2(\kappa) = \norm{\rho(\kappa\;\cdot)}^2-\mu_0^2(\kappa)-\mu_1^2(\kappa)$,  $c(\gamma,\kappa) = \bE{u,v\sim P^\gamma}{\varrho(\kappa u)\varrho(\kappa v)}$, where $g\sim\cN{0,1}$ and $P^\gamma$ denote the bivariate standard Gaussian distribution with correlation coefficient $\gamma$. Explicitly,
\begin{equation*}
    P^\gamma(x,y) = \frac{1}{2\pi\sqrt{1-\gamma^2}}e^{-\frac{x^2+y^2-2\gamma xy}{2(1-\gamma^2)}}\;.
\end{equation*}
\begin{lemma}\label{lemma:lc_m1}
    Let $\cM$ be a subspace in $\R^D$. Let $\Pi_\parallel$ denote a projection matrix onto $\cM$, and let $\Pi_\perp = I-\Pi_\parallel$. Suppose $P_0\equiv\cN{0,\cC}$. Let $a_t, h_t, \lambda, \varrho, W_t$ be as in \eqref{eqn:dsm_loss_rfm_m1}, and define $\Sigma_t = a_t^2\cC+h_tI_d$, $\sigma_t^2 = \frac{1}{d}\tr{\Sigma_t}$, and $\sigma_0^2 = \frac{1}{d}\tr{\cC}$.  Assume that $\sigma_t$ converges to a finite limit as $d\to\infty$. Let $Z=[z_1,z_2,\cdots,z_n]$ with $z_i \sim \cN{0,I_d}$ i.i.d., and let $Y=[y_1,y_2,\cdots,y_n]$ for $y_i = a_t x_i+\sqrt{h_t}z_i$, with $x_i \sim P_0$ i.i.d. Assume $\varrho$ satisfies Assumption~\ref{assmptn:activation_fn}. Then, for the minimizer of (\ref{eqn:dsm_loss_rfm_m1}) $\hat{A}_t$, for $\alpha\in\{\parallel,\perp\}$, as $d,n,p\to\infty$, the test and train errors can be represented as:
    \begin{align*}
    \cE^1_{\text{test},\alpha}(\hat{A}_t) &= E_{0,\alpha}-\frac{2\mu_1(\sigma_t)}{\sigma_t\sqrt{h_t}}E_{1,\alpha}+\frac{\mu_1(\sigma_t)^2}{\sigma_t^2 h_t}E_{2,\alpha}+\frac{v(\sigma_t)^2}{h_t}E_{3,\alpha}+ \frac{\mu_0(\sigma_t)^2}{h_t}E_{4,\alpha}\;,\\
    \cE^1_{\text{train}}(\hat{A}_t) &=-E_5-\lambda (E_{3,\parallel}+E_{3,\perp})+1 \;,
\end{align*}
where
\begin{align*}
    E_{0,\alpha} &= \frac{1}{d}\tr{\Pi_\alpha\Sigma^{-1}}\;,\\
    E_{1,\alpha} &= \frac{1}{d}\tr{\Pi_\alpha\rat{Z}{n}\rat{F^T}{n}(U+\lambda I_p)^{-1}\rat{W_t}{d},\Pi_\alpha}\;,\\
    E_{2,\alpha} &= \frac{1}{d}\tr{\Pi_\alpha\rat{Z}{n}\rat{F^T}{n}(U+\lambda I_p)^{-1}\rat{W_t}{d}\rat{W_t^T}{d}(U+\lambda I_p)^{-1}\rat{F}{n}\rat{Z^T}{n}\Pi_\alpha}\;,\\
    E_{3,\alpha} &= \frac{1}{d}\tr{\Pi_\alpha\rat{Z}{n}\rat{F^T}{n}(U+\lambda I_p)^{-2}\rat{F}{n}\rat{Z^T}{n}\Pi_\alpha}\;,\\
    E_{4,\alpha} &= \frac{1}{d}\bone_p^T (U+\lambda I_p)^{-1}\rat{F}{n}\rat{Z^T}{n}\Pi_\alpha\rat{Z}{n}\rat{F^T}{n}(U+\lambda I_p)^{-1}\bone_p\;,\\
    E_5 &= \frac{1}{d}\tr{\rat{Z}{n}\rat{F^T}{n}(U+\lambda I_p)^{-1}\rat{F}{n}\rat{Z^T}{n}}\;,
\end{align*}
with 
\begin{equation*}
    U = \rat{F}{n}\rat{F^T}{n}\;,
\end{equation*}
and
\begin{equation*}
    F = \mu_0(\nu_t)\bone_p\bone_n^T+ \frac{\mu_1(\sigma_t)}{\sigma_t}\rat{W_t}{d}Y+v(\sigma_t)\Omega\;.
\end{equation*}
\end{lemma}
\begin{proof}
When $m=1$, the loss function reduces to:
\begin{align}
    \cL^1_t(A_t) &=\frac{1}{dn}\sum_{i=1}^{n}{{\norm{\sqrt{h_t}\rat{A_t}{p}\act{\rat{W_t}{p}(a_tx_i+\sqrt{h_t}z_i)}+z_i}^2}}+\frac{h_t\lambda}{dp}\norm{A_t}_F^2 \;.
\end{align}

Let $X=[x_1,x_2,\cdots,x_n]$, $Z=[z_1,z_2,\cdots,z_n]$, and let $Y$ be the matrix whose $i^{th}$ column is $y_i = a_tx_i+\sqrt{h_t}z_i$. Also let $F=\varrho(\rat{W}{d}Y)$. We then write:
\begin{align*}
    \cL^1_t(A_t) &=\frac{1}{dn}\sum_{i=1}^{n}{{\norm{\sqrt{h_t}\rat{A_t}{p}\act{\rat{W_t}{p}(a_tx_i+\sqrt{h_t}z_i)}+z_i}^2}}+\frac{h_t\lambda}{dp}\norm{A_t}_F^2 \\
    &=\frac{h_t}{d}\tr{\rat{A_t}{p}^T\rat{A_t}{p} U}+\frac{2\sqrt{h_t}}{d}\tr{\rat{A_t}{p} V}+1+\frac{h_t\lambda}{d}\tr{\rat{A_t^T}{p}\rat{A_t}{p}} \;,\\
\end{align*}
where 
$$U = \frac{1}{n}\act{\rat{W_t}{d}Y}\act{\rat{W_t}{d}Y}^T = \rat{F}{n}\rat{F^T}{n} \;,$$
and 
$$V = \frac{1}{n}{\act{\rat{W_t}{d}Y}Z^T} = \rat{F}{n}\rat{Z^T}{n} \;.$$ Thus, the optimal $A_t$ is written as 
\begin{equation}
    \rat{\hat{A}_t}{p} = -\frac{1}{\sqrt{h_t}}V^T(U+\lambda I_p)^{-1} = -\frac{1}{\sqrt{h_t}}\rat{Z}{n}\rat{F^T}{n}\left(\rat{F}{n}\rat{F^T}{n}+\lambda I_p\right)^{-1}.
\end{equation}
\textbf{Test error:}
\begin{align*}
    \cE^1_{\text{test},\alpha}(\hat{A}_t) &= \frac{1}{d}\bE{x\sim P_t}{\norm{\Pi_\alpha\rat{\hat{A}_t}{p}\act{\rat{W_t}{d}x}-\Pi_\alpha\nabla\log P_t(x)}^2}\\
    &= \frac{1}{d}\bE{x\sim P_t}{\norm{\rat{F}{n}\rat{Z^T}{n}\Pi_\alpha\rat{\hat{A}_t}{p}\act{\rat{W_t}{d}x}+\Pi_\alpha\Sigma_t^{-1}x}^2}\\
    &=\frac{1}{d}\tr{\Pi_\alpha\Sigma_t^{-1}} + \frac{2}{d}\tr{\Sigma_t^{-1}\Pi_\alpha\rat{\hat{A}_t}{p}\underbrace{\bE{x}{\act{\rat{W_t}{d}x}x^T}}_{:=\Tilde{V}}}\\&\qquad\qquad +\frac{1}{d}\tr{\rat{\hat{A}^T_t}{p}\Pi_\alpha\rat{\hat{A}_t}{p}\underbrace{\bE{x}{\act{\rat{W_t}{d}x}\act{\rat{W_t}{d}x}^T}}_{:=\Tilde{U}}}\;.
\end{align*}

We have already derived expressions for $\Tilde{V}$ and $\Tilde{U}$ in Section~\ref{appndx:lc_minf_proof}. Namely, $\Tilde{V} = \frac{\mu_1(\sigma_t)}{\sigma_t}\rat{W_t}{d}\Sigma_t$,\; $\Tilde{U} = \mu_0(\sigma_t)^2\bone_p\bone_p^T+ \frac{\mu_1(\sigma_t)^2}{\sigma_t^2}\rat{W_t}{d}\Sigma_t\rat{W_t^T}{d}+v(\sigma_t)^2I_p$. Since we focus on a single time instant, we drop the subscript $t$ henceforth. However, it is important to keep the time-dependence of $a$ and $h$, and the relation $a^2+h=1$. Thus:
\begin{align*}
    \cE^1_{\text{test},\alpha}(\hat{A}_t)  &= \frac{1}{d}\tr{\Pi_\alpha\Sigma^{-1}} - \frac{2\mu_1(\sigma)}{\sqrt{h}\sigma d}\tr{\Pi_\alpha\rat{Z}{n}\rat{F^T}{n}\left(\rat{F}{n}\rat{F^T}{n}+\lambda I_p\right)^{-1}\rat{W}{d}}\\
    &\qquad+\frac{\mu_1(\sigma)^2}{h\sigma^2d}\tr{\left(\rat{F}{n}\rat{F^T}{n}+\lambda I_p\right)^{-1}\rat{F}{n}\rat{Z^T}{n}\Pi_\alpha\rat{Z}{n}\rat{F^T}{n}\left(\rat{F}{n}\rat{F^T}{n}+\lambda I_p\right)^{-1}\rat{W}{d}\Sigma\rat{W^T}{d}}\\
    &\qquad+\frac{v(\sigma)^2}{hd}\tr{\left(\rat{F}{n}\rat{F^T}{n}+\lambda I_p\right)^{-1}\rat{F}{n}\rat{Z^T}{n}\Pi_\alpha\rat{Z}{n}\rat{F^T}{n}\left(\rat{F}{n}\rat{F^T}{n}+\lambda I_p\right)^{-1}}\;,\\
    &\qquad+\frac{\mu_0(\sigma)^2}{hd}\bone_p^T\left(\rat{F}{n}\rat{F^T}{n}+\lambda I_p\right)^{-1}\rat{F}{n}\rat{Z^T}{n}\Pi_\alpha\rat{Z}{n}\rat{F^T}{n}\left(\rat{F}{n}\rat{F^T}{n}+\lambda I_p\right)^{-1}\bone_p \;.
\end{align*}
To handle the non-linearity in $F$, we apply the Gaussian equivalence principle. In particular:
\begin{equation*}
    F = \mu_0(\sigma)\bone_p\bone_n^T+ \frac{\mu_1(\sigma)}{\sigma}\rat{W}{d}Y+v(\sigma)\Omega\;.
\end{equation*}
Let 
\begin{align*}
    E_{0,\alpha} &= \frac{1}{d}\tr{\Pi_\alpha\Sigma^{-1}},\\
    E_{1,\alpha} &= \frac{1}{d}\tr{\Pi_\alpha\rat{Z}{n}\rat{F^T}{n}\left(\rat{F}{n}\rat{F^T}{n}+\lambda I_p\right)^{-1}\rat{W}{d}\Pi_\alpha}\;,\\
    E_{2,\alpha} &= \frac{1}{d}\tr{\left(\rat{F}{n}\rat{F^T}{n}+\lambda I_p\right)^{-1}\rat{F}{n}\rat{Z^T}{n}\Pi_\alpha\rat{Z}{n}\rat{F^T}{n}\left(\rat{F}{n}\rat{F^T}{n}+\lambda I_p\right)^{-1}\rat{W}{d}\Sigma\rat{W^T}{d}}\;,\\
    E_{3,\alpha} &= \frac{1}{d}\tr{\rat{F}{n}\rat{Z^T}{n}\Pi_\alpha\rat{Z}{n}\rat{F^T}{n}\left(\rat{F}{n}\rat{F^T}{n}+\lambda I_p\right)^{-2}}\;,\\
    E_{4,\alpha} &= \frac{1}{d}\bone_p^T \left(\rat{F}{n}\rat{F^T}{n}+\lambda I_p\right)^{-1}\rat{F}{n}\rat{Z^T}{n}\Pi_\alpha\rat{Z}{n}\rat{F^T}{n}\left(\rat{F}{n}\rat{F^T}{n}+\lambda I_p\right)^{-1}\bone_p\;,\\
    E_5 &= \frac{1}{d}\tr{\rat{F}{n}\rat{Z^T}{n}\rat{Z}{n}\rat{F^T}{n}\left(\rat{F}{n}\rat{F^T}{n}+\lambda I_p\right)^{-1}}\;.
\end{align*}
With the above definitions, we can write $\cE^1_{\text{test},\alpha}$ as 
\begin{equation*}
   \cE^1_{\text{test},\alpha}(\hat{A}_t) = E_{0,\alpha}-\frac{2\mu_1(\sigma)}{\sqrt{h}\sigma}E_{1,\alpha}+\frac{\mu_1(\sigma)^2}{h\sigma^2}E_{2,\alpha}+\frac{v(\sigma)^2}{h}E_{3,\alpha}+ \frac{\mu_0(\sigma)^2}{h}E_{4,\alpha}\;.
\end{equation*}
\textbf{Train error:}
To compute the train error, we proceed as follows:
\begin{align*}
    \cE^1_{\text{train}}(\hat{A}_t)    &=\frac{h}{d}\tr{\rat{\hat{A}_t^T}{p}\rat{\hat{A}_t}{p} (U+\lambda I_p)}+\frac{2\sqrt{h}}{d}\tr{\rat{\hat{A}_t}{p} V}+1-\frac{h\lambda}{d}\tr{\rat{\hat{A}_t^T}{p}\rat{\hat{A}_t}{p}}\\
    &=\frac{1}{d}\tr{(U+\lambda I_p)^{-1}VV^T}-\frac{2}{d}\tr{V^T(U+\lambda I_p)^{-1}V}+1-\frac{\lambda}{d}\tr{(U+\lambda I_p)^{-2}VV^T}\\
    &=-\frac{1}{d}\tr{(U+\lambda I_p)^{-1}\rat{F}{n}\rat{Z^T}{n}\rat{Z}{n}\rat{F^T}{n}}+1-\frac{\lambda}{d}\tr{(U+\lambda I_p)^{-2}\rat{F}{n}\rat{Z^T}{n}\rat{Z}{n}\rat{F^T}{n}}\\
    &=-E_5-\lambda (E_{3,\parallel}+E_{3,\perp})+1 \;.
\end{align*}
This completes the proof of lemma \ref{lemma:lc_m1}.
\end{proof}
The following theorem derives the test and train errors when $P_0$ is a Gaussian supported on a $D$-dimensional subspace in $\R^d$.
\begin{theorem}\label{appendix:thm:lc_m1_man}
    Let $\cM_\parallel$ be a $D$-dimensional subspace in $\R^D$, $\Pi_\parallel$ a projection matrix onto it, and $P_0\equiv\cN{0,\Pi_\parallel}$ with $\varrho$ satisfying Assumption~\ref{assmptn:activation_fn}. Define $\sigma_0^2 = \frac{1}{d}\tr{\Pi_\parallel}$, $\sigma_t^2 = a_t^2\sigma_0^2+h_t$, and $\mu_{1,t} = \mu_1(\sigma_t)/\sigma_t$. Set $v^2 = \norm{\varrho(\sigma_t\cdot)}^2-\mu_1(\sigma_t)^2$. Let $\psi_D = \frac{D}{d}, \psi_n = \frac{n}{d}$, $\psi_p = \frac{p}{d}$ and $\zeta_1,\zeta_2,\zeta_3,\zeta_4,\zeta_5, \zeta_6$ be the solution of the following system of algebraic equations in $q$ and $z$:
    \begin{align*}
    \zeta_1(-z+(1-\psi_D)(q+\mu_{1,t}^2\zeta_4)h_t\zeta_6 +\psi_D(q+\mu_{1,t}^2\zeta_4)\zeta_3+v^2\zeta_4) -1&= 0 \;,\\
    \zeta_2(1+q\psi_p \zeta_1) +\mu_{1,t}^2\psi_p\zeta_1\zeta_2\zeta_4 +\psi_pa_t\mu_{1,t}\zeta_1&= 0 \;,\\
    \zeta_5(1+qh_t\psi_p \zeta_1) +\mu_{1,t}\psi_p\sqrt{h_t}\zeta_1(1+\mu_{1,t}\sqrt{h_t}\zeta_4\zeta_5) &= 0 \;,\\
    \zeta_3(1+q\psi_p \zeta_1) +\mu_{1,t}^2\psi_p\zeta_1\zeta_3\zeta_4 -1&= 0 \;,\\
    \zeta_4(\psi_n+\psi_pv^2\zeta_1 -(1-\psi_D)\mu_{1,t}\sqrt{h_t}\zeta_5 - \psi_D\mu_{1,t}\zeta_2/a_t)-\psi_n &= 0 \;,\\
    \zeta_6(1+qh_t\psi_p \zeta_1) +\mu_{1,t}^2\psi_ph_t\zeta_1\zeta_6\zeta_4 -1&= 0 \;.
\end{align*}
    Let $e_{0,\parallel} = \psi_D$ and $e_{0,\perp} = \frac{1-\psi_D}{h_t}$. Define the functions 
%%%%%%%%%%%
%    \begin{align}
%    e_{1,\alpha} &= \begin{cases}
%        -\frac{\psi_D\sqrt{h_t}}{a_t}\zeta_4\zeta_2,\quad \alpha = \parallel,\\
%        -(1-\psi_D)\zeta_4\zeta_5,\quad \alpha = \perp,
%    \end{cases} \;,\\
%    K_\alpha(q,z) & = \begin{cases}
%        \psi_D(1-\zeta_4(1+\mu_{1,t}h_t\zeta_4\zeta_2/a_t)),\quad \alpha = \parallel,\\
%        1-\psi_D-\frac{1}{\mu_{1,t}\sqrt{h_t}}(qh_te_{1,\perp}+(1-\psi_D\mu_{1,t}\sqrt{h_t}\zeta_4\zeta_6)),\quad \alpha = \perp,
%    \end{cases} \;.
%\end{align}
%%%%%%%%%%%%
\begin{align}
     e_{1,\alpha} &= \left\{\begin{alignedat}{2}
    & -\frac{\psi_D\sqrt{h_t}}{a_t}\zeta_4\zeta_2 \;, && \alpha = \parallel \\
    & -(1-\psi_D)\zeta_4\zeta_5  \;, \qquad  && \alpha = \perp 
  \end{alignedat}\;, \right. \\
    K_\alpha(q,z) &= \left\{\begin{alignedat}{2}
    & \psi_D(1-\zeta_4(1+\mu_{1,t}h_t\zeta_4\zeta_2/a_t))  \;, && \alpha = \parallel \\
    & 1-\psi_D-\frac{1}{\mu_{1,t}\sqrt{h_t}}(qh_te_{1,\perp}+(1-\psi_D)\mu_{1,t}\sqrt{h_t}\zeta_4\zeta_6)  \;, \qquad  && \alpha = \perp 
  \end{alignedat}\;.\right. 
\end{align}
%%%%%%%%%%%%%
    For $\alpha\in\{\parallel,\perp\}$,
    let $\varepsilon^{1}_{\text{test},\alpha} =e_{0,\alpha}-\frac{2\mu_{1,t}}{\sqrt{h_t}}e_{1,\alpha}-\frac{\mu_{1,t}^2}{h_t}\frac{\partial \cK_\alpha}{\partial q}(0,-\lambda)+\frac{v^2}{h_t}\frac{\partial \cK_\alpha}{\partial z}(0,-\lambda) \;$ and $\varepsilon^1_{\text{train}} = 1-(\cK_\parallel(0,-\lambda)+\cK_\perp(0,-\lambda))-\lambda (\frac{\partial \cK_\parallel}{\partial z}(0,-\lambda)+\frac{\partial \cK_\perp}{\partial z}(0,-\lambda)).$ Then, for the minimizer of (\ref{eqn:dsm_loss_rfm_minf}) $\hat{A}_t$, as $d,n,p\to\infty$:
    \begin{align*}
        \lim_{d,n,p\to\infty}\bE{}{\cE^1_{\text{test},\parallel}(\hat{A}_t)} =  \varepsilon^1_{\text{test},\parallel}\;,\quad \lim_{d,n,p\to\infty}\bE{}{\cE^1_{\text{test},\perp}(\hat{A}_t)} =  \varepsilon^1_{\text{test},\perp}\;,\quad
    \lim_{d,n,p\to\infty}\bE{}{\cE^1_{\text{train}}(\hat{A}_t)} &= \varepsilon^1_{\text{train}} \;.
    \end{align*}
\end{theorem}
\begin{remark}
    Theorem~\ref{thm:lc_m1} can be obtained by substituting $\psi_D=1$.
\end{remark}
\begin{proof}
To prove Theorem~\ref{thm:lc_m1}, we specialize Lemma~\ref{lemma:lc_m1} to $\cC=\Pi_\parallel$. As in the proof of Lemma~\ref{lemma:lc_m1}, since we focus on a single time instant, we drop the subscript $t$ in the above expressions. For the $\cC$ used here, we note that $\sigma^2 = a^2\psi_D+h, \sigma_0^2 =  \psi_D$. Also, let $\mu_0 = \mu_0(\sigma), \mu_1 = \mu_1(\sigma)/\sigma, \norm{\varrho(\sigma\;\cdot)} = \norm{\varrho}$. Let $W = [W_\parallel,W_\perp]$, $Z = [Z_\parallel^T,Z_\perp^T]^T$, $X = [X_\parallel^T,X_\perp^T]^T$, and $Y = [Y_\parallel^T,Y_\perp^T]^T$

For simplicity in presentation, we assume that $\mu_0=0$ and write
\begin{align*}
   \cE^1_{\text{test},\alpha}(\hat{A}_t) &= E_{0,\alpha}-\frac{2\mu_1}{\sqrt{h}}E_{1,\alpha}+\frac{\mu_1^2}{h}E_{2,\alpha}+\frac{v^2}{h}E_{3,\alpha} \;,\\
   \cE^1_{\text{train}}(\hat{A}_t) &=-E_5-\lambda (E_{3,\parallel}+E_{3,\perp})+1 \;.
\end{align*}
Letting
\begin{align*}
    R(q,z) &= \left(\rat{F}{n}\rat{F^T}{n}+q\rat{W}{d}\Sigma\rat{W^T}{d}-zI_p\right)^{-1} \;,\\
    &= \left(\rat{F}{n}\rat{F^T}{n}+q\rat{W_\parallel}{d}\rat{W_\parallel^T}{d}+qh\rat{W_\perp}{d}\rat{W_\perp^T}{d}-zI_p\right)^{-1} \;,
\end{align*}
and
\begin{align*}
    K_\alpha(q,z) = \frac{1}{d}\tr{\rat{Z_\alpha}{n}\rat{F^T}{n}R(q,z)\rat{F}{n}\rat{Z_\alpha^T}{n}} \;,
\end{align*}
we have
\begin{align*}
    E_{2,\alpha} &= -\frac{\dd K_\alpha}{\dd q}(0,-\lambda) \;,\\
    E_{3,\alpha} &= \frac{\dd K_\alpha}{\dd z}(0,-\lambda) \;,\\
    E_{5} &= K_\parallel(0,-\lambda)+K_\perp(0,-\lambda) \;.
\end{align*}

Therefore, assuming we can interchange limit, derivatives and expectations, it suffices to have $e_{1,\alpha} = \lim_{d\to\infty}\bE{}{E_{1,\alpha}}$ and the function $\cK_{\alpha}(q,z):=\lim_{d\to\infty}\bE{}{K_\alpha(q,z)}$ to have an expression for $\lim_{d\to\infty}\bE{}{\cE^1_{\text{test},\alpha}(\hat{A}_t)}$. We have
\begin{align}\label{eqn:Etest_m1}
    \lim_{d\to\infty}\bE{}{\cE^1_{\text{test},\alpha}(\hat{A}_t)} &= \lim_{d\to\infty}E_{0,\alpha}-\frac{2\mu_1}{\sqrt{h}}\lim_{d\to\infty}\bE{}{E_{1,\alpha}}+\frac{\mu_1^2}{h}\lim_{d\to\infty}\bE{}{E_{2,\alpha}}+\frac{v^2}{h}\lim_{d\to\infty}\bE{}{E_{3,\alpha}} \nonumber\\
    &= e_{0,\alpha}-\frac{2\mu_1}{\sqrt{h}}e_{1,\alpha}+\frac{\mu_1^2}{h}e_{2,\alpha}+\frac{v^2}{h}e_{3,\alpha} \; ,
\end{align}
where $e_{2,\alpha} = -\frac{\partial \cK_\alpha}{\partial q}(0,-\lambda),\; e_{3,\alpha} = \frac{\partial \cK_\alpha}{\partial z}(0,-\lambda)$. As in the previous theorem, one could avoid interchanging limit of expectation and derivatives by differentiating first $K_\alpha(q,z)$ and using a larger linear pencil.

As in Theorem~\ref{thm:lc_minf}, proved in Appendix~\ref{appndx:lc_minf_proof}, we use linear pencils to obtain the desired terms. Explicitly, the following $6\times 6$ linear pencil matrix:
\begin{equation*}
    L = \mleft[
        \begin{array}{c|ccccccc}
        -zI_p & qh\rat{W_\perp}{d} & q\rat{W_\parallel}{d} & v\rat{\Omega}{n} & \mu_1\sqrt{h}\rat{W_\perp}{d} & \mu_1\sqrt{h}\rat{W_\parallel}{d} & \mu_1a\rat{W_\parallel}{d} & 0\\
        \hline
        -\rat{W_\perp^T}{d} & I_d & 0 & 0 & 0 & 0 & 0 & 0\\
        -\rat{W_\parallel^T}{d} & 0 & I_d & 0 & 0 & 0 & 0 & 0\\
        -v\rat{\Omega^T}{n} & -\mu_1\sqrt{h}\rat{Z_\perp^T}{n} & -\mu_1\rat{Y_\parallel^T}{n} & I_n & 0 & 0 & 0 & 0\\
        0 & 0 & 0 & -\rat{Z_\perp}{n} & I_d & 0 & 0 & 0\\
        0 & 0 & 0 & -\rat{Z_\parallel}{n} & 0 & I_d & 0 & 0\\
        0 & 0 & 0 & -\rat{X_\parallel^T}{n} & 0 & 0 & I_n & 0\\
        0 & 0 & 0 & 0 & 0 & -\rat{Z_\parallel^T}{n} & 0 & I_n
    \end{array}
    \mright]\;.
\end{equation*}
By computing $L^{-1}$, we obtain:
\begin{align}
    (L^{-1})^{8,4}(q,z) &= \rat{Z_\parallel^T}{n}\rat{Z_\parallel}{n} -\rat{Z_\parallel^T}{n}\rat{Z_\parallel}{n}\rat{F^T}{n}R(q,z)\rat{F}{n} \;,\\
    (L^{-1})^{6,7}(q,z) &= -\mu_1a\rat{Z_\parallel}{n}\rat{F^T}{n}R(q,z)\rat{W_\parallel}{d} \;,\\
    (L^{-1})^{5,2}(q,z) &= \mu_1\sqrt{h}\rat{Z_\perp}{n}\rat{Z_\perp^T}{n} -\mu_1\sqrt{h}\rat{Z_\perp}{n}\rat{F^T}{n}R(q,z)\rat{F}{n}\rat{Z_\perp^T}{n} -qh\rat{Z_\perp}{n}\rat{F^T}{n}R(q,z)\rat{W_\perp^T}{d} \;,\\
    (L^{-1})^{5,5}(q,z) &= I -\mu_1\sqrt{h}\rat{Z_\perp}{n}\rat{F^T}{n}R(q,z)\rat{W_\perp^T}{d} \;.
\end{align}
Hence,
\begin{align}
    E_{1,\alpha} &= \frac{1}{d}\tr{\rat{Z_\alpha}{n}\rat{F^T}{n}R(0,-\lambda)\rat{W_\alpha}{d}} = \begin{cases}
        -\frac{1}{\mu_1ad}\tr{(L^{-1})^{6,7}(0,-\lambda)},\quad \alpha = \parallel,\\
        \frac{1-\psi_D}{\mu_1\sqrt{h}}(1-\frac{1}{d-D}\tr{(L^{-1})^{5,5}(0,-\lambda)}),\quad \alpha = \perp,
    \end{cases} \;,\\
    K_\alpha(q,z) &= \frac{1}{d}\tr{\rat{Z_\alpha}{n}\rat{F^T}{n}R(q,z)\rat{F}{n}\rat{Z_\alpha^T}{n}} = \begin{cases}
        \psi_D-\frac{1}{d}\tr{(L^{-1})^{8,4}(0,-\lambda)},\quad \alpha = \parallel,\\
        1-\psi_D-\frac{1}{\mu_1\sqrt{h}}(qhE_{1,\perp}+\frac{1}{d}\tr{(L^{-1})^{5,2}(0,-\lambda)}),\quad \alpha = \perp,
    \end{cases} \;.
\end{align}
The traces of blocks in $L^{-1}$ are computed as in the Appendix~\ref{appndx:lc_minf_proof}. We then have: 
%\begin{equation*}
%    g = \mleft[
%    \begin{array}{cccccc}
%        g_{11} & 0 & 0 & 0 & 0 & 0 \\
%        0 & g_{22} & 0 & g_{24} & 0 & 0 \\
%        0 & 0 & g_{33} & 0 & 0 & 0 \\
%        0 & g_{42} & 0 & g_{44} & 0 & 0 \\
%        0 & g_{52} & 0 & g_{54} & 1 & 0 \\
%        0 & 0 & g_{63} & 0 & 0 & 1  \\
%    \end{array}
%    \mright]\;,
%\end{equation*}
%\begin{equation*}
%    B = \mleft[
%    \begin{array}{cccccc}
%        -z & 0 & 0 & 0 & 0 & 0 \\
%        0 & 1 & 0 & 0 & 0 & 0 \\
%        0 & 0 & 1 & 0 & 0 & 0 \\
%        0 & 0 & 0 & 1 & 0 & 0 \\
%        0 & 0 & 0 & 0 & 1 & 0 \\
%        0 & 0 & 0 & 0 & 0 & 1  \\
%    \end{array}
%    \mright] \;,
%\end{equation*}
\begin{alignat*}{2}
  g=&\mleft[
    \begin{array}{cccccccc}
        g_{11} & 0 & 0 & 0 & 0 & 0 & 0 & 0\\
        0 & g_{22} & 0 & 0 & g_{25} & 0 & 0 & 0\\
        0 & 0 & g_{33} & 0 & 0 & g_{36} & g_{37} & 0\\
        0 & 0 & 0 & g_{44} & 0 & 0 & 0 & 0\\
        0 & g_{52} & 0 & 0 & g_{55} & 0 & 0 & 0\\
        0 & 0 & g_{63} & 0 & 0 & g_{66}  & g_{67} & 0\\
        0 & 0 & g_{73} & 0 & 0 & g_{76}  & g_{77} & 0\\
        0 & 0 & 0 & g_{84} & 0 & 0  & 0 & 1\\
    \end{array}
    \mright] \;,  \quad  B=&&  \mleft[
    \begin{array}{cccccccc}
        -z & 0 & 0 & 0 & 0 & 0 & 0 & 0 \\
        0 & 1 & 0 & 0 & 0 & 0 & 0 & 0\\
        0 & 0 & 1 & 0 & 0 & 0 & 0 & 0\\
        0 & 0 & 0 & 1 & 0 & 0 & 0 & 0\\
        0 & 0 & 0 & 0 & 1 & 0 & 0 & 0\\
        0 & 0 & 0 & 0 & 0 & 1 & 0 & 0\\
        0 & 0 & 0 & 0 & 0 & 0 & 1 & 0\\
        0 & 0 & 0 & 0 & 0 & 0 & 0 & 1\\
    \end{array}
    \mright] \;,
\end{alignat*}
\begin{equation*}
    \eta_L(g) = \mleft[
    \begin{array}{cccccccc}
        \begin{smallmatrix}
            \sigma_{12}^{21}g_{22}+\sigma_{15}^{21}g_{52}+\sigma_{13}^{31}g_{33}+\\\sigma_{16}^{31}g_{63}+\sigma_{17}^{31}g_{73}+\sigma_{14}^{41}g_{44}
        \end{smallmatrix} & 0 & 0 & 0 & 0 & 0 & 0 & 0 \\
        0 & \sigma_{21}^{12}g_{11} & 0 & 0 & \sigma_{21}^{15}g_{11} & 0 & 0 & 0\\
        0 & 0 & \sigma_{31}^{13}g_{11} & 0 & 0 & \sigma_{31}^{16}g_{11} & \sigma_{31}^{17}g_{11} & 0\\
        0 & 0 & 0 &
        \begin{smallmatrix}
        \sigma_{41}^{14}g_{11}+\sigma_{42}^{54}g_{25} +\\ \sigma_{43}^{64}g_{36}+\sigma_{43}^{74}g_{37} \end{smallmatrix}  & 0 & 0 & 0 & 0 \\
        0 & \sigma_{54}^{42}g_{44} & 0 & 0 & 0 & 0 & 0 & 0  \\
        0 & 0 & \sigma_{64}^{43}g_{44} & 0 & 0 & \sigma_{64}^{86}g_{48} & 0 & 0  \\
        0 & 0 & \sigma_{74}^{43}g_{44} & 0 & 0 & 0 & 0 & 0  \\
        0 & 0 & 0 &\sigma_{86}^{64}g_{66} & 0 & 0 & 0 & 0  \\
    \end{array}
    \mright] \;,
\end{equation*}
and
\begin{alignat}{4}
  & \sigma_{12}^{21} = -(1-\psi_D)qh \;,&&\quad\quad \sigma_{21}^{12}= -\psi_pqh \;,&&\quad\quad \sigma_{15}^{21}= -(1-\psi_D)\mu_1\sqrt{h} \;,&&\quad\quad \sigma_{21}^{15}= -\psi_p\mu_1\sqrt{h} \\
  & \sigma_{13}^{31} = -\psi_Dq \;,&&\quad\quad \sigma_{31}^{13}= -\psi_pq \;,&&\quad\quad \sigma_{16}^{31}= -\psi_D\mu_1\sqrt{h} \;,&&\quad\quad \sigma_{31}^{16}= -\psi_p\mu_1\sqrt{h} \\
  & \sigma_{17}^{31} = -\psi_Da\mu_1 \;,&&\quad\quad \sigma_{31}^{17}= -\psi_pa\mu_1 \;,&&\quad\quad \sigma_{14}^{41}= -v^2 \;,&&\quad\quad \sigma_{41}^{14}= -\psi_pv^2/\psi_n \\
  & \sigma_{42}^{54} = (1-\psi_D)\mu_1\sqrt{h}/\psi_n \;,&&\quad\quad \sigma_{54}^{42}= \mu_1\sqrt{h} \;,&&\quad\quad \sigma_{43}^{64}= \psi_D\mu_1\sqrt{h}/\psi_n \;,&&\quad\quad \sigma_{64}^{43}= \mu_1\sqrt{h} \\
  & \sigma_{43}^{74} = \psi_Da\mu_1/\psi_n \;,&&\quad\quad \sigma_{74}^{43}= a\mu_1 \;,&&\quad\quad \sigma_{64}^{86}= 1 \;,&&\quad\quad \sigma_{86}^{64}= \psi_D/\psi_n.
\end{alignat}
Finally, the matrix $g$ satisfies the fixed point equation:
\begin{equation*}
    (B-\eta_L(g))g = I \;.
\end{equation*}
This leads, after simplification, the following set of algebraic equations:
\begin{align*}
    \zeta_1(-z+(1-\psi_D)(q+\mu_1^2\zeta_4)h\zeta_6 +\psi_D(q+\mu_1^2\zeta_4)\zeta_3+v^2\zeta_4) -1&= 0 \;,\\
    \zeta_2(1+q\psi_p \zeta_1) +\mu_1^2\psi_p\zeta_1\zeta_2\zeta_4 +\psi_pa\mu_1\zeta_1&= 0 \;,\\
    \zeta_5(1+qh\psi_p \zeta_1) +\mu_1\psi_p\sqrt{h}\zeta_1(1+\mu_1\sqrt{h}\zeta_4\zeta_5) &= 0 \;,\\
    \zeta_3(1+q\psi_p \zeta_1) +\mu_1^2\psi_p\zeta_1\zeta_3\zeta_4 -1&= 0 \;,\\
    \zeta_4(\psi_n+\psi_pv^2\zeta_1 -(1-\psi_D)\mu_1\sqrt{h}\zeta_5 - \psi_D\mu_1\zeta_2/a)-\psi_n &= 0 \;,\\
    \zeta_6(1+qh\psi_p \zeta_1) +\mu_1^2\psi_ph\zeta_1\zeta_6\zeta_4 -1&= 0 \;.
\end{align*}
where $\zeta_1=g_{11},\; \zeta_2=g_{37},\; \zeta_3=g_{33},\; \zeta_4=g_{44},\; \zeta_5=g_{25},\;
\zeta_6=g_{22}$. As before, we solve the system numerically to obtain $\zeta_1,\zeta_2,\zeta_3,\zeta_4,$ $\zeta_5$, and $\zeta_6$. We then obtain $e_{1,\alpha}$ and $\cK_\alpha$ through the expressions: 
%\begin{align}
%    e_{1,\alpha} &= \begin{cases}
%        -\frac{\psi_D\sqrt{h}}{a}\zeta_4\zeta_2 \; , \; \quad \alpha = \parallel,\\
%        -(1-\psi_D)\zeta_4\zeta_5 \; , \; \quad \alpha = \perp,
%    \end{cases} \;,\\
%    K_\alpha(q,z) & = \begin{cases}
%        \psi_D(1-\zeta_4(1+\mu_1h\zeta_4\zeta_2/a)),\quad \alpha = \parallel,\\
%        1-\psi_D-\frac{1}{\mu_1\sqrt{h}}(qhe_{1,\perp}+(1-\psi_D\mu_1\sqrt{h}\zeta_4\zeta_6)),\quad \alpha = \perp,
%    \end{cases} \;.
%\end{align}
\begin{align}
     e_{1,\alpha} &= \left\{\begin{alignedat}{2}
    & -\frac{\psi_D\sqrt{h}}{a}\zeta_4\zeta_2 \;, && \alpha = \parallel \\
    & -(1-\psi_D)\zeta_4\zeta_5 \;, \qquad  && \alpha = \perp 
  \end{alignedat}\;, \right. \\
    K_\alpha(q,z) &= \left\{\begin{alignedat}{2}
    & \psi_D(1-\zeta_4(1+\mu_1h\zeta_4\zeta_2/a)) \;, && \alpha = \parallel \\
    & 1-\psi_D-\frac{1}{\mu_1\sqrt{h}}(qhe_{1,\perp}+(1-\psi_D)\mu_1\sqrt{h}\zeta_4\zeta_6) \;, \qquad  && \alpha = \perp 
  \end{alignedat}\;.\right. 
\end{align}

Finally, we use \eqref{eqn:Etest_m1} to compute $\lim_{d\to\infty}\bE{}{\cE^1_{\text{test},\alpha}(\hat{A}_t)}$.\\\\
\textbf{Train error:}
To compute the train error, we proceed as follows:
\begin{align*}
    \cE^1_{\text{train}}(\hat{A}_t)    &=-E_5-\lambda (E_{3,\parallel}+E_{3,\perp})+1 \;,\\
    &=-(K_\parallel(0,-\lambda)+K_\perp(0,-\lambda))-\lambda(\frac{\partial K_\parallel}{\partial z}(0,-\lambda)+\frac{\partial K_\perp}{\partial z}(0,-\lambda))+1 \;.
\end{align*}
Thus,
\begin{align}\label{eqn:Etrain_m1}
    \lim_{d\to\infty}\bE{}{\cE^1_{\text{train}}(\hat{A}_t)} &= 1-(\cK_\parallel(0,-\lambda)+\cK_\perp(0,-\lambda))-\lambda(\frac{\partial \cK_\parallel}{\partial z}(0,-\lambda)+\frac{\partial \cK_\perp}{\partial z}(0,-\lambda)) \;,\nonumber\\
    &= 1-(\cK_\parallel(0,-\lambda)+\cK_\perp(0,-\lambda))-\lambda (e_{3,\perp}+e_{3,\perp}). \;,
\end{align}
where $e_{3,\alpha} = \frac{\partial \cK_\alpha}{\partial z}(0,-\lambda)$.
This concludes the proof of Theorem~\ref{thm:lc_m1}.

\end{proof}

\section{ILLUSTRATIONS OF ANALYTICALLY COMPUTED TEST AND TRAIN ERRORS}
In this section, we provide additional plots to further illustrate the analytical predictions of test and train errors derived from Theorems~\ref{appendix:thm:lc_minf_man} and \ref{appendix:thm:lc_m1_man}. First, we present the learning curves for the case of $m=1$, which were omitted from the main text due to space constraints. Next we show the learning curves when the data lies on a $D$-dimensional subspace in $\R^d$. Thereafter, we include plots demonstrating the impact of the regularization strength $\lambda$ on the learning curves. Finally, we provide learning curves for cases where different activation functions are employed, highlighting their influence on the model's performance. 

\subsection{Test and train errors for $m=1$}\label{appndx:learning_curves_m1}
\begin{figure}[h]
     \centering
     \begin{subfigure}[b]{0.8\textwidth}
         \centering
         \includegraphics[width=\textwidth]{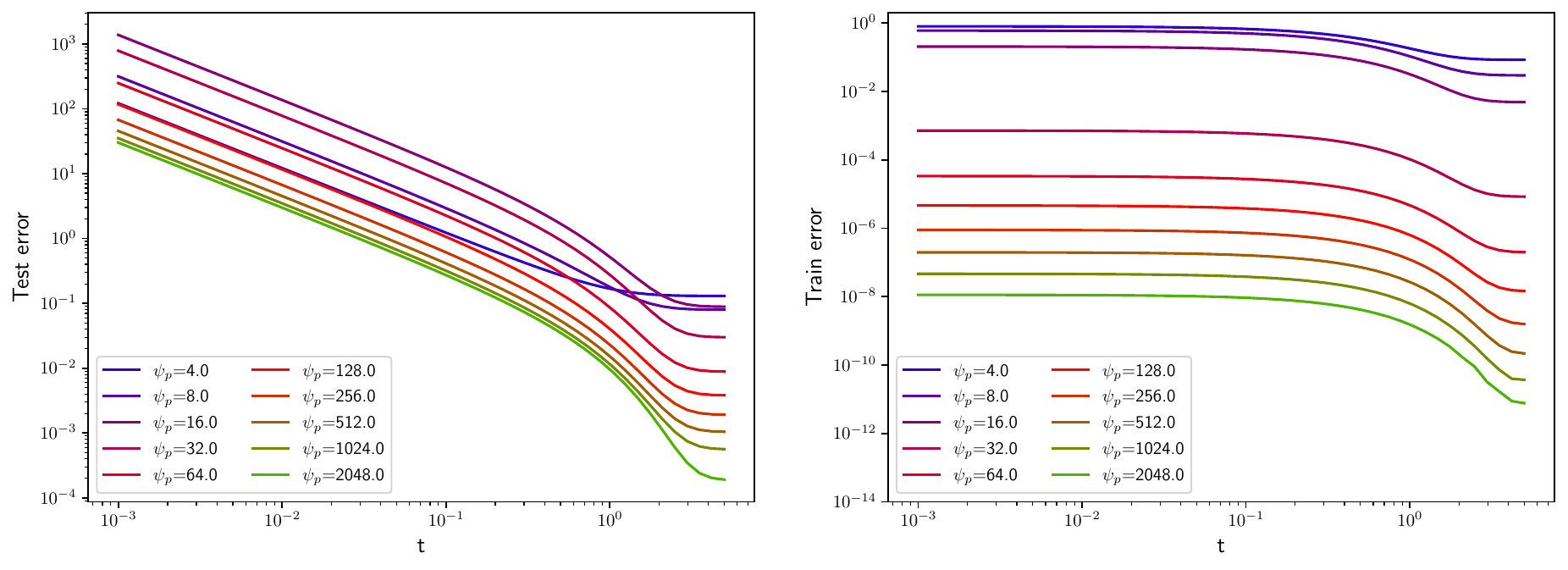}
         \caption{Test and train error as a function of $t$ for different values of $\psi_p$ and a fixed $\psi_n=20.0$.}
    \label{fig:lc_m1_nfixed}
     \end{subfigure}
     \hfill
     
     \begin{subfigure}[b]{0.8\textwidth}
         \centering
         \includegraphics[width=\textwidth]{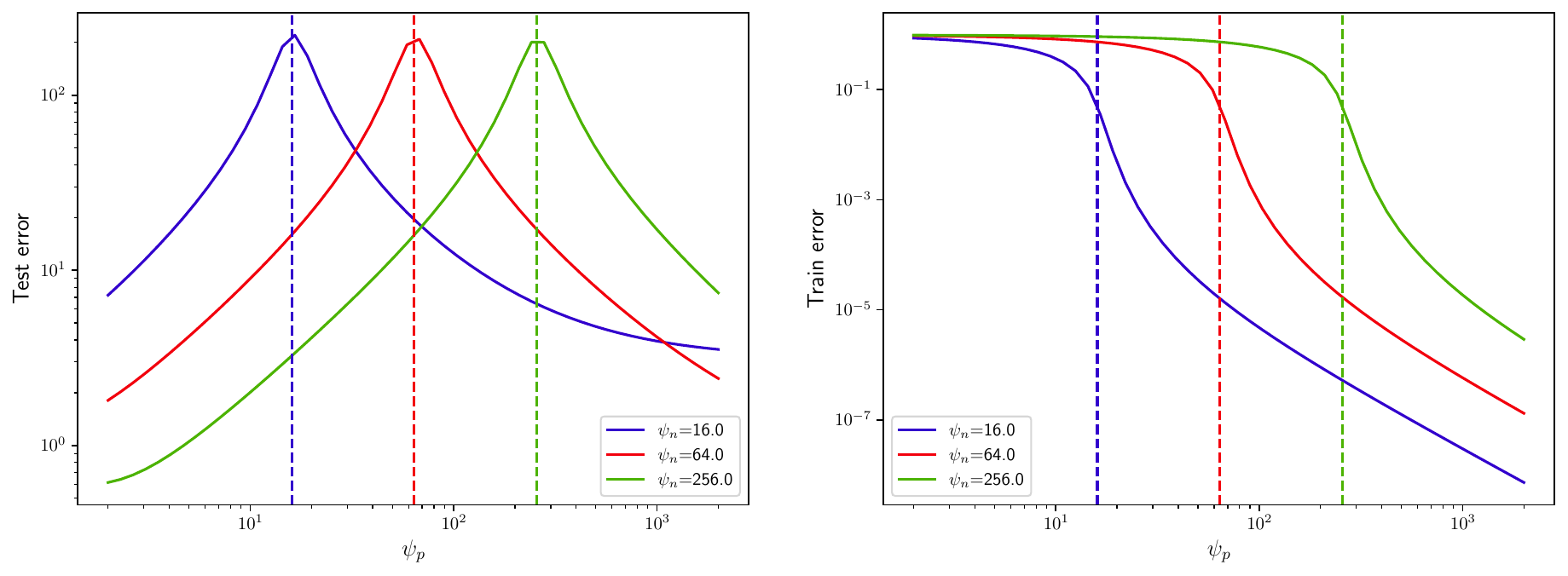}
         \caption{Test and train error as a function of $\psi_p$ for different values of $\psi_n$ and a fixed $t = 0.01$. The dashed vertical lines indicate $\psi_p=\psi_n$.}
    \label{fig:lc_m1_tfixed}
     \end{subfigure}
     \hfill
        \caption{Learning curves for $m=1$.We used $\lambda = 0.001$ and the activation function used is ReLU.}
        \label{fig:lc_m1}
\end{figure}
Fig.~\ref{fig:lc_m1} shows test and train errors for $m=1$ case. Fig.~\ref{fig:lc_m1_nfixed} presents the learning curves as a function of $t$ for various $\psi_p$ with $\psi_n$ fixed. In Fig.~\ref{fig:lc_m1_tfixed} we plot the learning curves as a function of $\psi_p$ for different values of $\psi_n$ and keeping $t$ fixed and small.

As already discussed in the main text, the curves reveal several trends. The train error decreases monotonically with increasing $\psi_p$ for all $t$, reflecting the model's capacity to interpolate the training data. However, the test error shows a non-monotonic behavior with $\psi_p$. The dependence of test error on $t$ is also evident. The test error increases as $t$ decreases, but for small $t$, it remains at least two orders of magnitude lower than in the $m=\infty$ case. Furthermore, the test error decreases as $\psi_p$ increases beyond $\psi_n$, suggesting the model does not display memorization behavior under these conditions. It contrasts with the $m=\infty$ case, where memorization significantly impacts the test error. These findings indicate that larger $m$ increases the tendency of diffusion models to memorize the initial dataset.

Lastly, as observed in earlier works such as~\cite{mei_generalization_2022, bodin_model_2021, hu_asymptotics_2024}, we also detect the double descent phenomenon, characterized by a peak in the test error at $\psi_p=\psi_n$ (see Fig.~\ref{fig:lc_m1_tfixed}).

\subsection{Plots when $D<d$}\label{appendix:lowdim}
\begin{figure}
     \centering
     \begin{subfigure}[b]{0.45\textwidth}
         \centering
         \includegraphics[width=\textwidth]{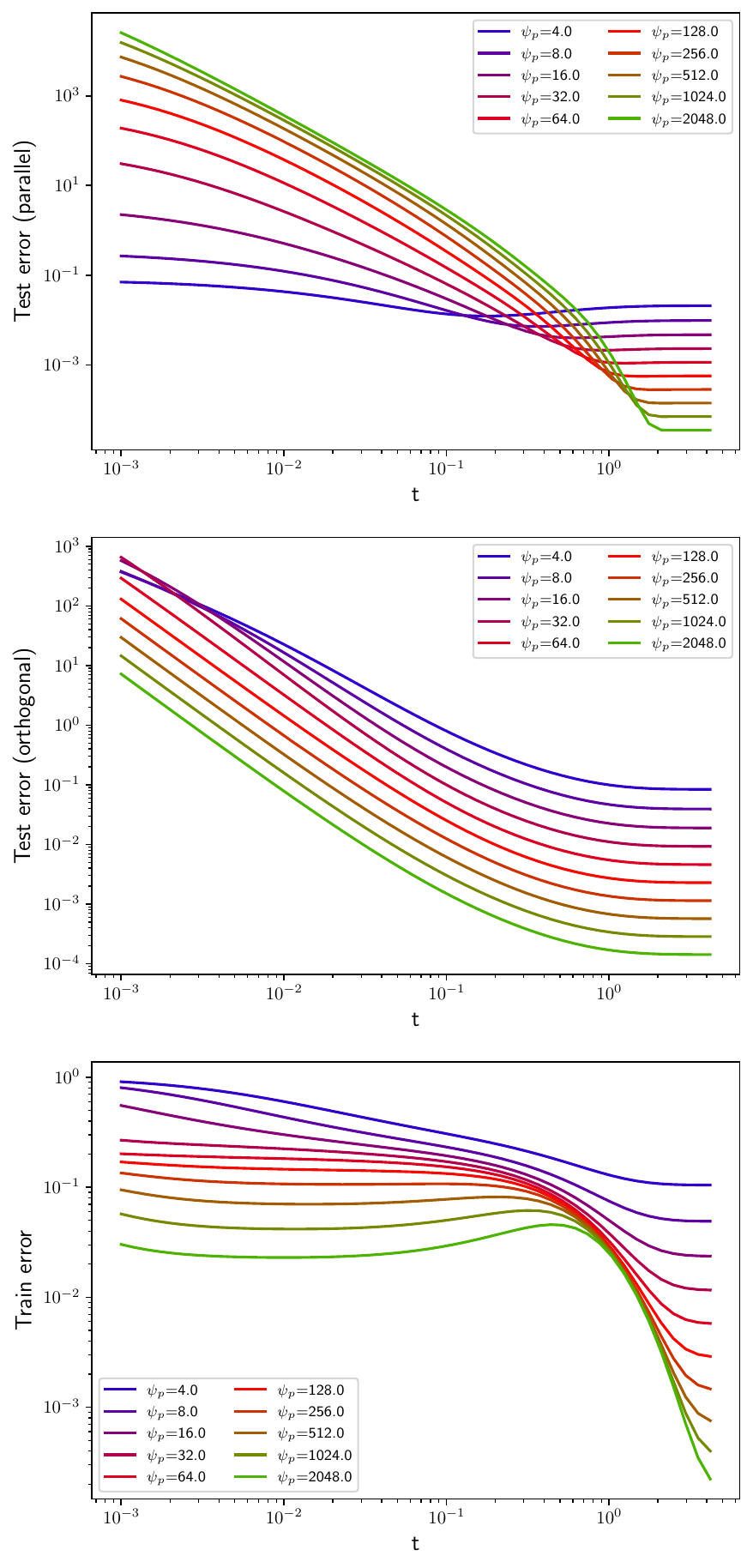}
         \caption{Test and train error for $m=\infty$ as a function of $t$ for different values of $\psi_p$ and a fixed $\psi_n=20.0$.}
    \label{fig:lc_lowD_minf_nfixed}
     \end{subfigure}
     \begin{subfigure}[b]{0.45\textwidth}
         \centering
         \includegraphics[width=\textwidth]{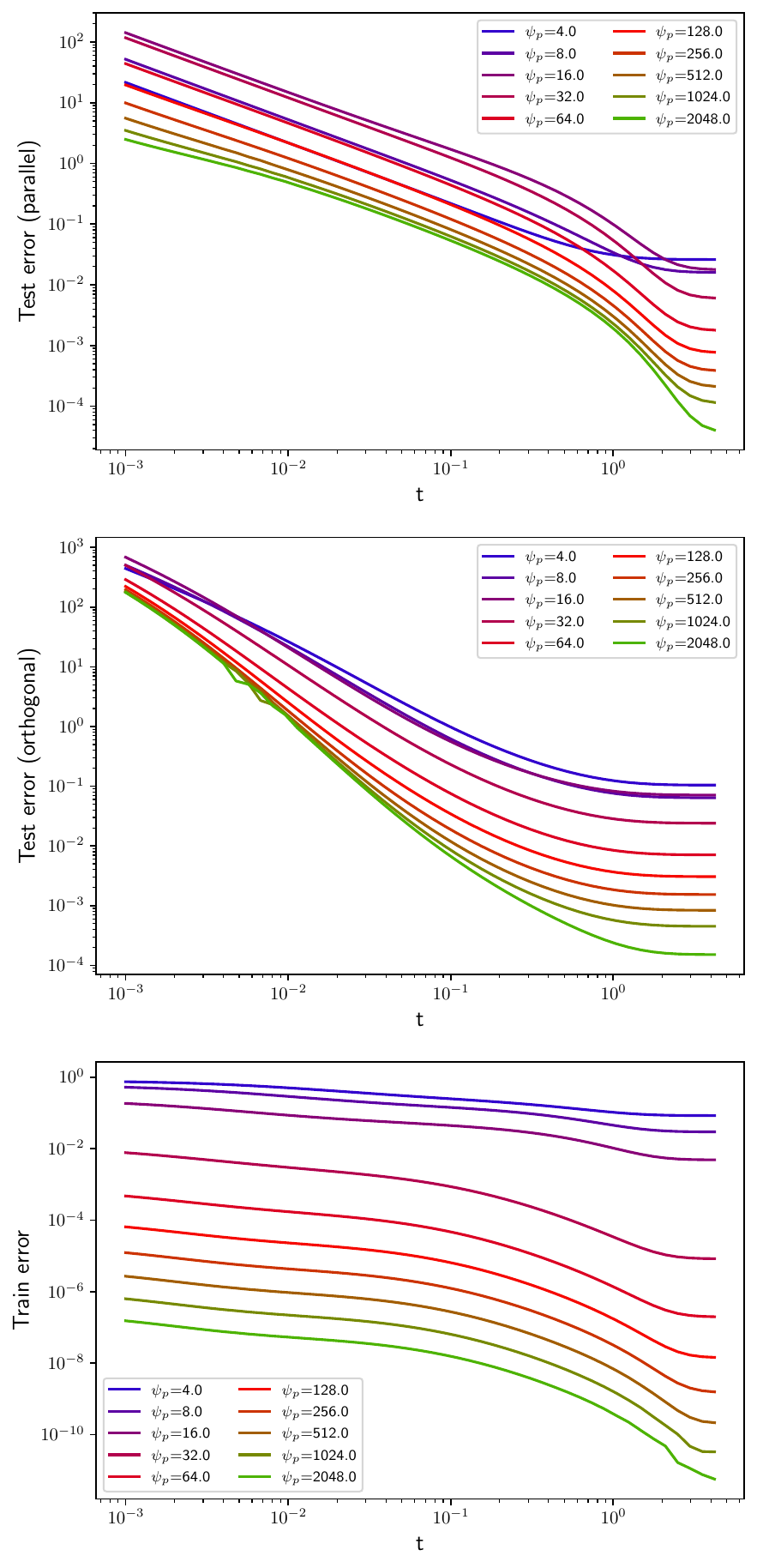}
         \caption{Test and train error for $m=1$ as a function of $t$ for different values of $\psi_p$ and a fixed $\psi_n=20.0$.}
    \label{fig:lc_lowD_m1_nfixed}
     \end{subfigure}
     \hfill
        \caption{Learning curves for $\psi_D = \frac{D}{d} = 0.2$. We used $\lambda = 0.001$ and the activation function used is ReLU.}
        \label{fig:lc_lowD}
\end{figure}

In this section, we present the learning curves for the case $\psi_D = \frac{D}{d} = 0.2$. When the data lies on a subspace, we need to investigate separately the components of the score function that are parallel and orthogonal to subspace, since they exhibit different behavior. As $t$ decreases, the magnitude of the score in the orthogonal direction becomes large. This would dominate the test error and we would not be able to observe the memorization-to-generalization transition along the subspace. This is the reason why we define directional generalization in $\eqref{eq:At_test_inf_man}$. Our Theorems~\ref{appendix:thm:lc_minf_man} and \ref{appendix:thm:lc_m1_man} compute the parallel and orthogonal contributions to the generalization separately. Figure~\ref{fig:lc_lowD} shows the resulting learning curves for $m=\infty$ and $m=1$. It is important to note that, as $\psi_p$ increases, the parallel and orthogonal test errors behave differently for small $t$. This happens because the memorization-to-generalization transition occurs only in the parallel component of the score.

\subsection{Plots for different values of $\lambda$}\label{appndx:lc_lambda}

\begin{figure}
     \centering
     \begin{subfigure}{0.7\textwidth}
         \centering
         \includegraphics[width=\textwidth]{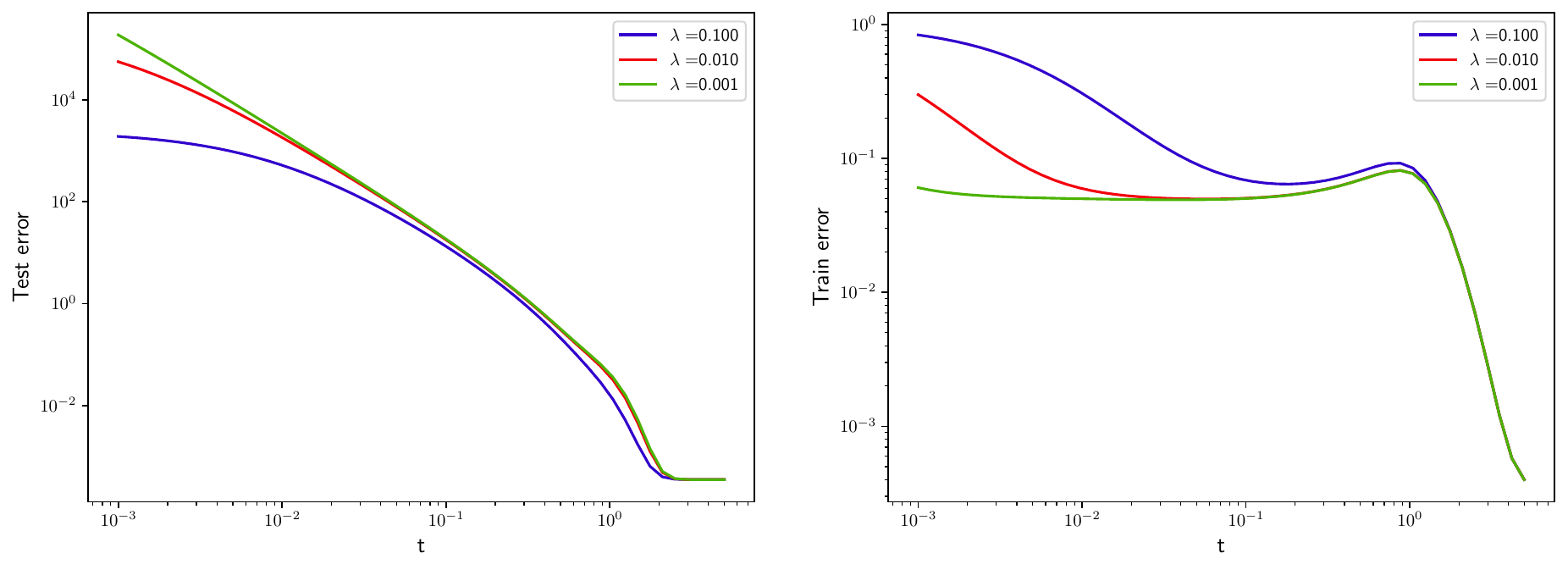}
         \caption{Test and train error as a function of $t$ for different values of $\lambda$ for a fixed $\psi_n=20.0$ and $\psi_p = 1024.0$.}
    \label{fig:lc_lambda_minf_nfixed}
     \end{subfigure}
     \hfill
     
     \begin{subfigure}{0.7\textwidth}
         \centering
         \includegraphics[width=\textwidth]{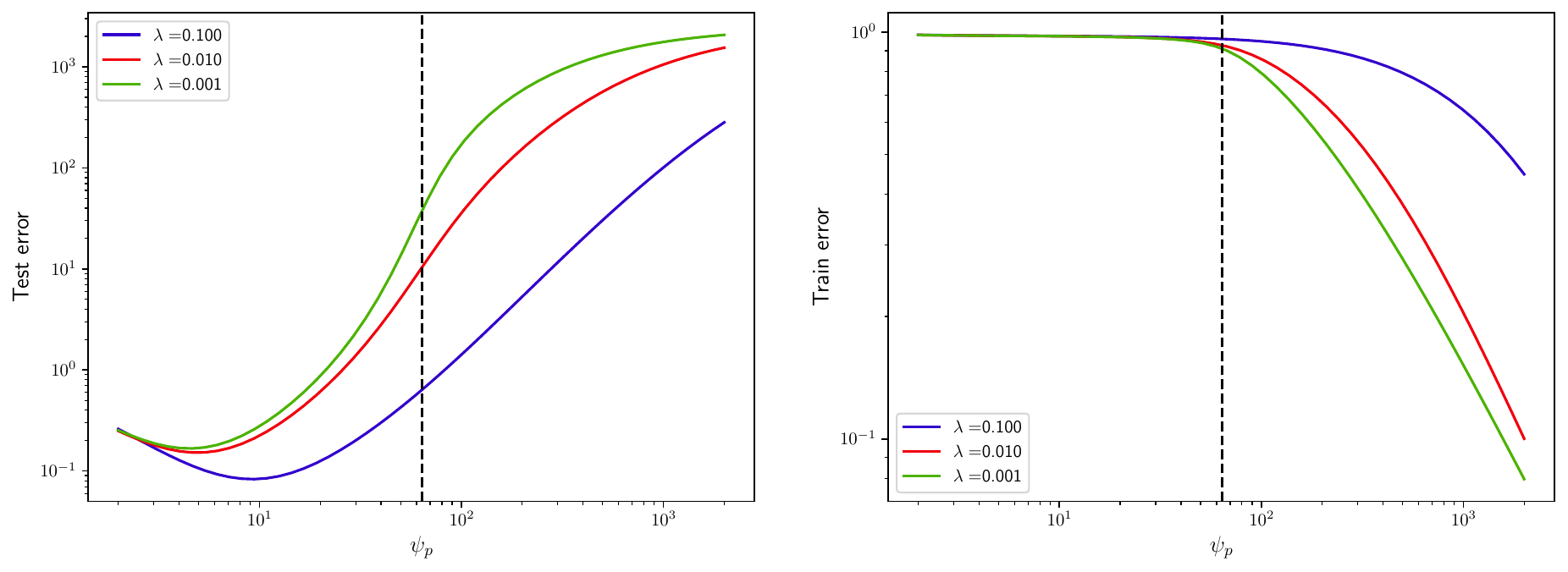}
         \caption{Test and train error as a function of $\psi_p$ for different values of $\lambda$ for a fixed $\psi_n=64.0$ and $t = 0.01$. The dashed vertical line indicates $\psi_p=\psi_n$.}
    \label{fig:lc_lambda_minf_tfixed}
     \end{subfigure}
     \hfill
        \caption{Learning curves for different values of $\lambda$ for $m=\infty$ and with ReLU activation.}
        \label{fig:lc_lambda_minf}
        
    \vspace{\intextsep}

     \centering
     \begin{subfigure}{0.7\textwidth}
         \centering
         \includegraphics[width=\textwidth]{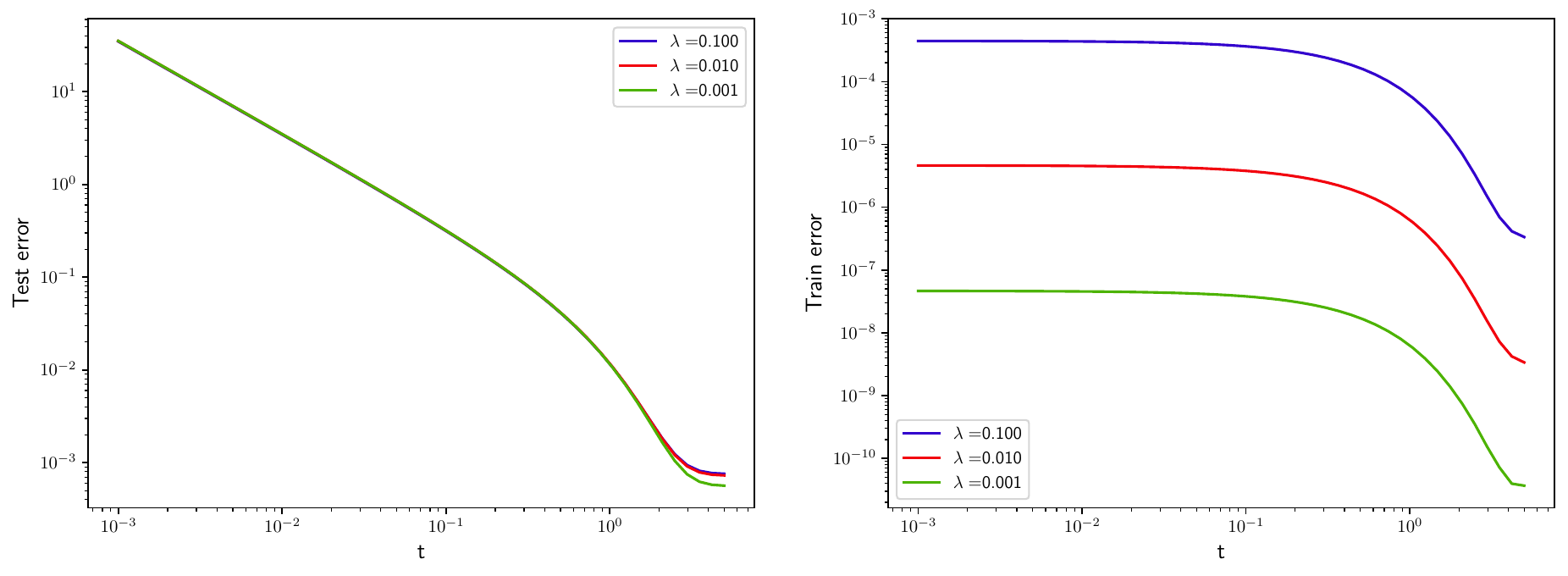}
         \caption{Test and train error as a function of $t$ for different values of $\lambda$ for a fixed $\psi_n=20.0$ and $\psi_p = 1024.0$.}
    \label{fig:lc_lambda_m1_nfixed}
     \end{subfigure}
     \hfill
     
     \begin{subfigure}{0.7\textwidth}
         \centering
         \includegraphics[width=\textwidth]{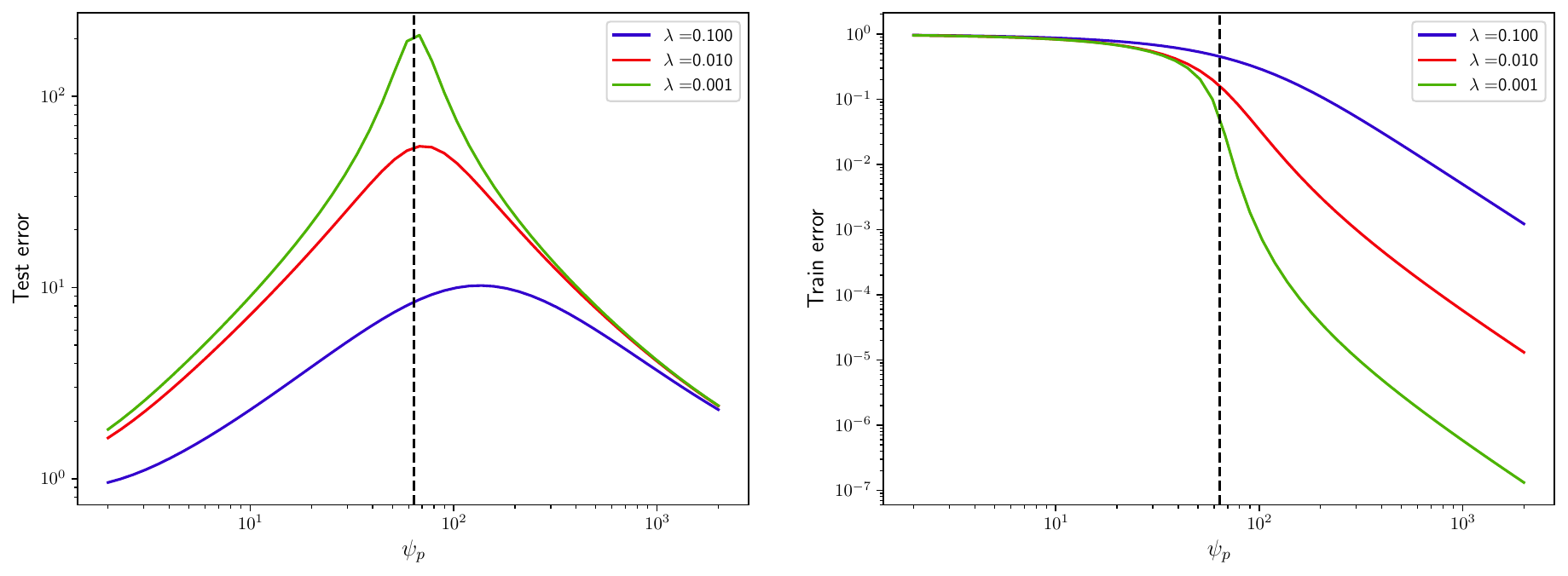}
         \caption{Test and train error as a function of $\psi_p$ for different values of $\lambda$ for a fixed $\psi_n=64.0$ and $t = 0.01$. The dashed vertical line indicates $\psi_p=\psi_n$.}
    \label{fig:lc_lambda_m1_tfixed}
     \end{subfigure}
     \hfill
        \caption{Learning curves for different values of $\lambda$ for $m=1$. The activation function used is ReLU.}
        \label{fig:lc_lambda_m1}
\end{figure}

In this section, we illustrate the behavior of the test and train errors for different values of $\lambda$. Fig.~\ref{fig:lc_lambda_minf} shows the learning errors for $m=\infty$. Note that for small $t$, the train error increases and the test error decreases as $\lambda$ increases. This can be explained as follows: In this regime, we have $s^e(t,a_tx_i+\sqrt{h_t}z)\approx -\frac{z}{\sqrt{h_t}}$. However, as regularization increases, $\hat{A}_t$ cannot take very large values. Therefore, $\cM_t$ scales as $\frac{1}{h_t}$, making $\cE^\infty_{\text{train}}(\hat{A}_t)\approx 1$ for very small values of $t$. This leads to lower memorization as well, as evidenced by the small test error.

Fig.~\ref{fig:lc_lambda_m1} displays the $m=1$ case. The peak at $\psi_p=\psi_n$ is attenuated as $\lambda$ increases, which is expected as regularization helps reducing overfitting. 

%In this case, larger values of $\lambda$ decreases the peak at $\psi_p=\psi_n$ due to double descent. This is expected since the regularization will help in reducing overfitting.

\subsection{Plots for different activation functions}\label{appndx:lc_activation}
\begin{figure}
     \centering
     \begin{subfigure}[b]{0.7\textwidth}
         \centering
         \includegraphics[width=\textwidth]{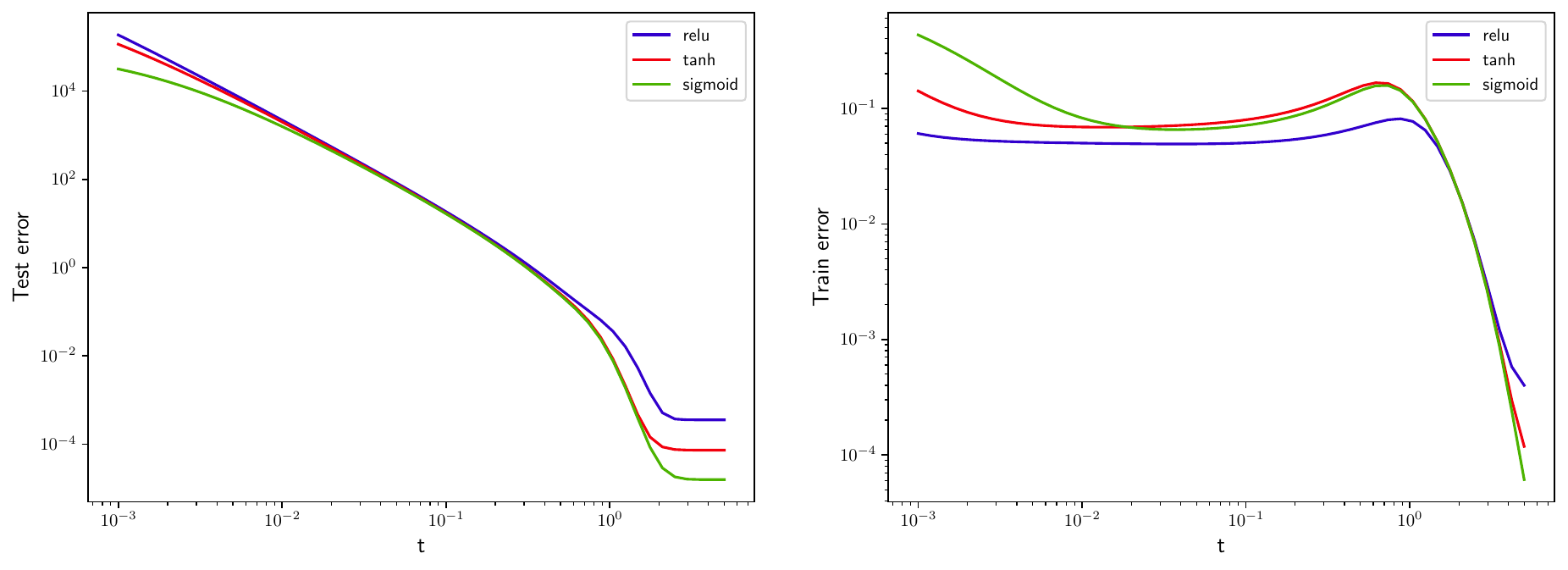}
         \caption{Test and train error as a function of $t$ for different activation functions for a fixed $\psi_n=20.0$ and $\psi_p = 1024.0$.}
    \label{fig:lc_act_minf_nfixed}
     \end{subfigure}
     \hfill
     
     \begin{subfigure}[b]{0.7\textwidth}
         \centering
         \includegraphics[width=\textwidth]{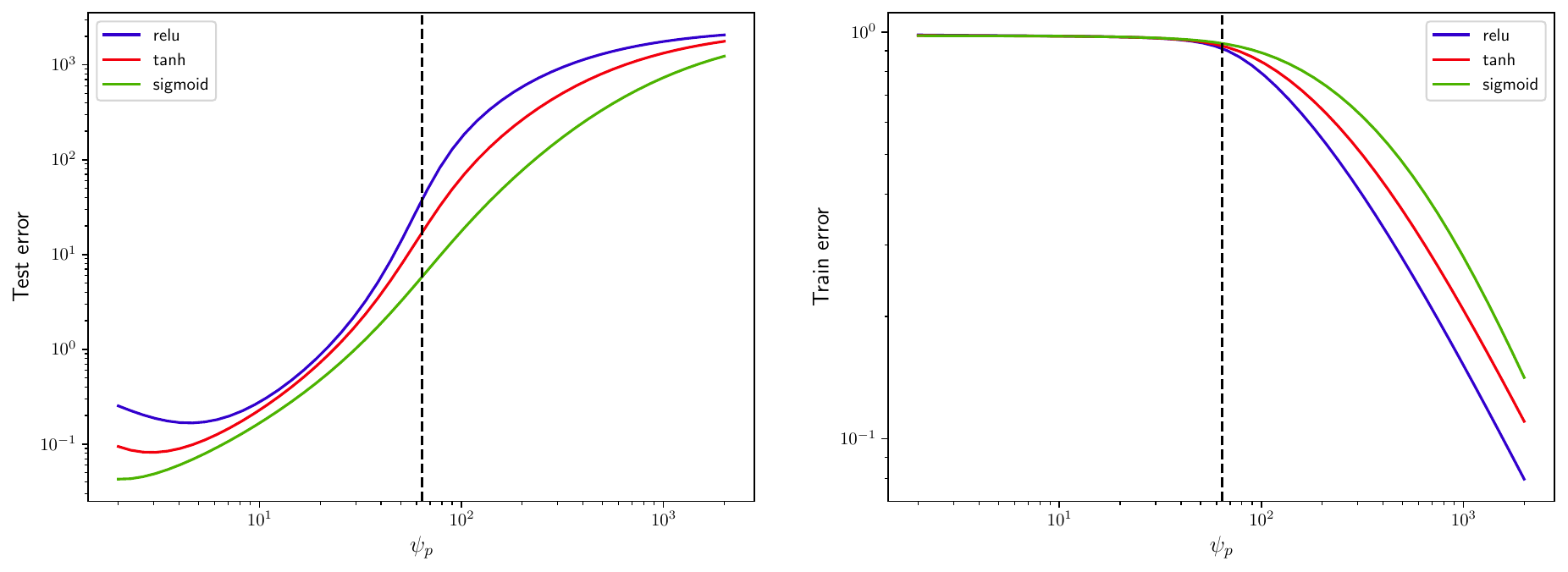}
         \caption{Test and train error as a function of $\psi_p$ for different activation functions for a fixed $\psi_n=64.0$ and $t = 0.01$. The dashed vertical line indicates $\psi_p=\psi_n$.}
    \label{fig:lc_act_minf_tfixed}
     \end{subfigure}
     \hfill
        \caption{Learning curves for different activation functions for $m=\infty$. We used $\lambda=0.001$.}
        \label{fig:lc_act_minf}

    \vspace{\intextsep}

     \centering
     \begin{subfigure}[b]{0.7\textwidth}
         \centering
         \includegraphics[width=\textwidth]{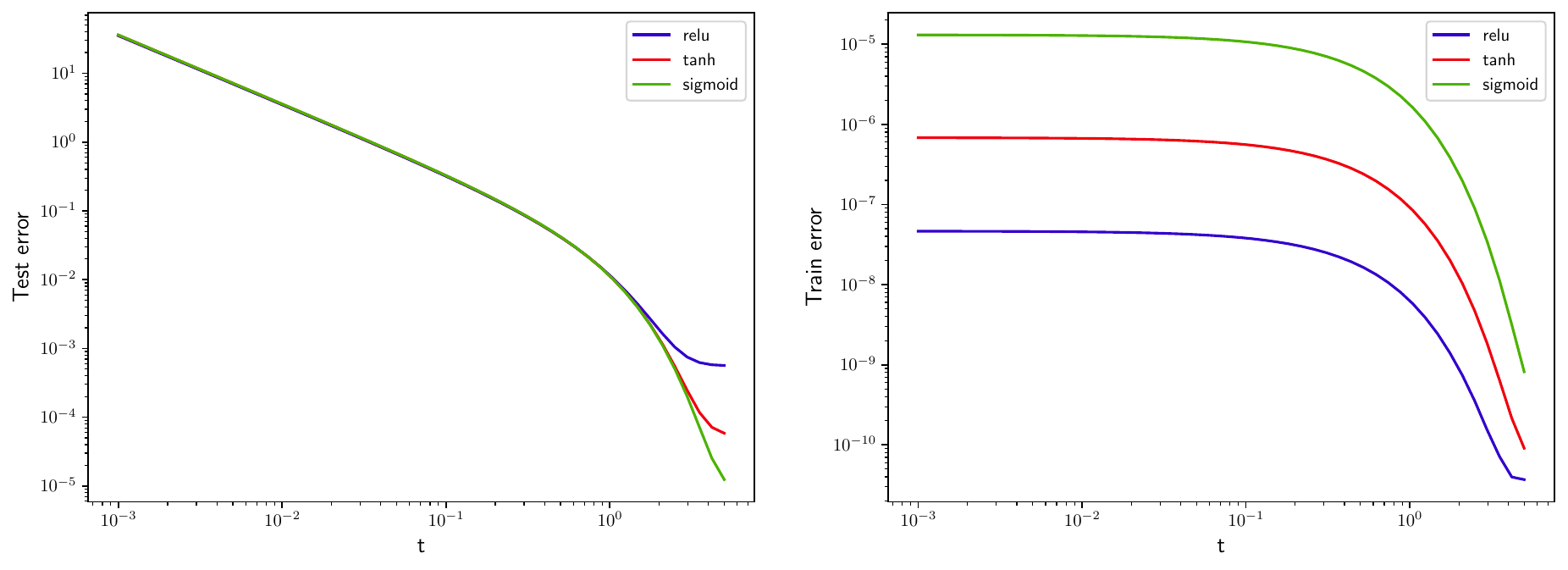}
         \caption{Test and train error as a function of $t$ for different activation functions for a fixed $\psi_n=20.0$ and $\psi_p = 1024.0$.}
    \label{fig:lc_act_m1_nfixed}
     \end{subfigure}
     \hfill
     
     \begin{subfigure}[b]{0.7\textwidth}
         \centering
         \includegraphics[width=\textwidth]{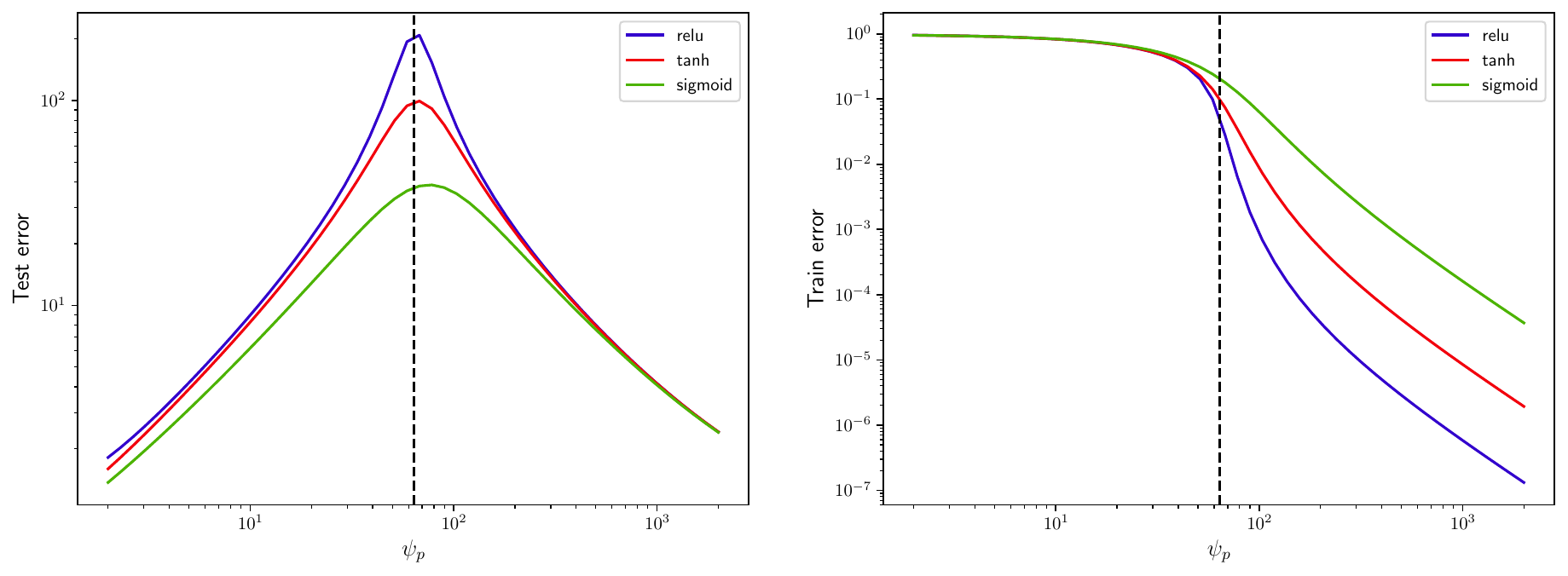}
         \caption{Test and train error as a function of $\psi_p$ for different activation functions for a fixed $\psi_n=64.0$ and $t = 0.01$. The dashed vertical line indicates $\psi_p=\psi_n$.}
    \label{fig:lc_act_m1_tfixed}
     \end{subfigure}
     \hfill
        \caption{Learning curves for different activation functions for $m=1$. We used $\lambda=0.001$.}
        \label{fig:lc_act_m1}
\end{figure}
%In this section, we illustrate the behavior of the learning curves for different activation functions. We compare three different activation functions: relu, tanh, and sigmoid. We introduces a shift and scaling to these functions inorder for them to have $\mu_0=0$ and have same $L_2$ norm. Concretely, these functions are given by:
We now exemplify the behavior of the learning curves for different activation functions: ReLU, tanh, and sigmoid. To make them have the same $L_2$ norm and $\mu_0=0$, we introduce proper shiftings and rescalings. Concretely:
\begin{align*}
    \text{ReLU}(x) &= x\mathds{1}\{x\ge 0\}-\frac{1}{\sqrt{2\pi}} \;, \\
    \text{tanh}(x) &= 0.93\left(\frac{e^x-e^{-x}}{e^x+e^{-x}}\right) \;,\\
    \text{sigmoid}(x) &= \frac{2.8}{1+e^{-x}}-1.4  \;.
\end{align*}
Fig.~\ref{fig:lc_act_minf} displays the results for $m=\infty$. The activation function enters the analysis through the coefficients $\mu_1, v, v_0,$ and $s$. 

Fig.~\ref{fig:lc_act_m1} presents the case $m=1$.
\newpage
\section{COMPARISON WITH NUMERICALLY OBTAINED LEARNING CURVES}\label{appndx:lc_numerical_comparison}
\begin{figure*}
    \centering
    \includegraphics[width=0.9\linewidth]{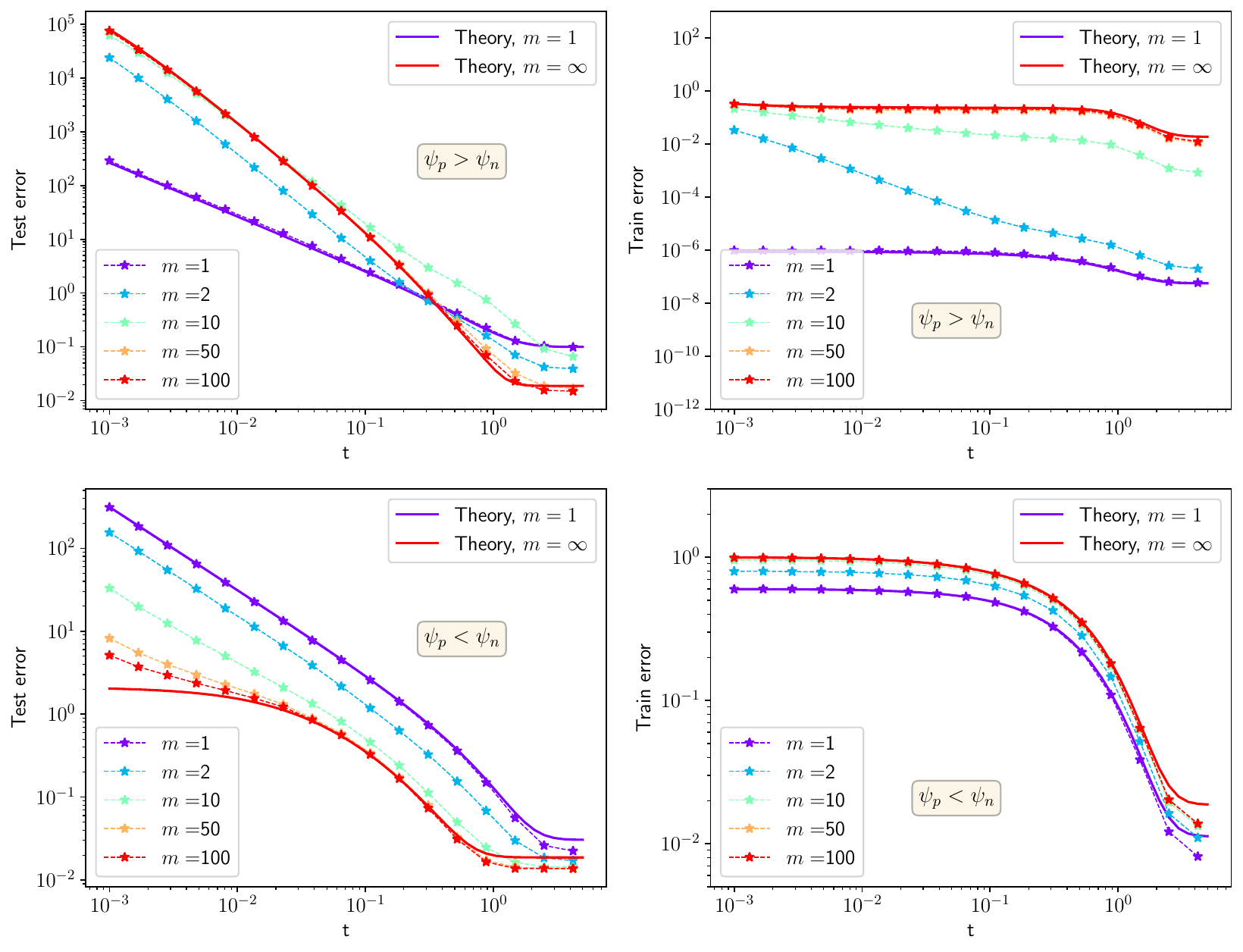}
    \caption{Simulation results ($d=100$) for different values of $m$ and fixed $\psi_p=20$; with $\psi_n=2$ (upper plots) and $\psi_n=50$ (lower plots). Theoretical results for $m=1$ and $m=\infty$ are depicted as continuous lines.}
    \label{fig:lc_sim_comp_appndx}
\end{figure*}
In this section, we discuss the learning curves obtained numerically for different values of $m$ in $\psi_p>\psi_n$ and $\psi_p<\psi_n$ regimes.
Fig.~\ref{fig:lc_sim_comp_appndx} presents the plots of test and train errors thus obtained. The upper plot displays the learning errors for the case $\psi_p>\psi_n$, while the lower plot corresponds to $\psi_p<\psi_n$. Based on the previous discussions, memorization is expected when $\psi_p>\psi_n$, and this behavior is evident in Fig.~\ref{fig:lc_sim_comp_appndx} when $t$ is small. Additionally, the extent of memorization increases with $m$, indicating that large $m$ is detrimental to generalization (small test error) in this regime. This is in contrast to the behavior of test and train errors in the $\psi_p<\psi_n$ regime, where the generalization improves as $m$ increases, evidenced by a decrease in the test error.

The solid lines in Fig.~\ref{fig:lc_sim_comp_appndx} plots the analytical predictions derived in Section~\ref{sec:main_results} for $m=1$ and $m=\infty$. These predictions align closely with the numerical results for $m=1$ and $m=100$ respectively, validating its consistency with the observed data. Minor mismatches between theoretical and numerical curves in the small $t$ and large $t$ regimes can be attributed to finite size effects.

 % The test error was illustrated in Fig.~\ref{fig:lc_sim}. In Fig.~\ref{fig:lc_sim_comp_appndx} we exhibit the corresponding train errors as well.
%\begin{figure*}[ht]
%    \centering
%    \includegraphics[width=1.0\linewidth]{figs/rfm_err_sim_lam0p001.pdf}
%    \caption{Learning curves obtained numerically for $\psi_p=20.0$. We used $d=100$ to get the numerical results. The upper row is for $\psi_n=2.0$ and the lower row is for $\psi_n=50.0$.}
%    \label{fig:lc_sim_comp_appndx}
%\end{figure*}
\section{DETAILS OF REAL DATA EXPERIMENT}\label{appndx:real_data_exp}
\begin{figure}
     \centering
     \begin{subfigure}{0.49\linewidth}
         \centering
         \includegraphics[width=\textwidth]{figs/mem_plot_fmnist.pdf}
         \caption{Results for Fashion-MNIST dataset.}
    \label{appndx:mem_sim_fmnist}
     \end{subfigure}
     \begin{subfigure}{0.49\linewidth}
         \centering
        \includegraphics[width=\textwidth]{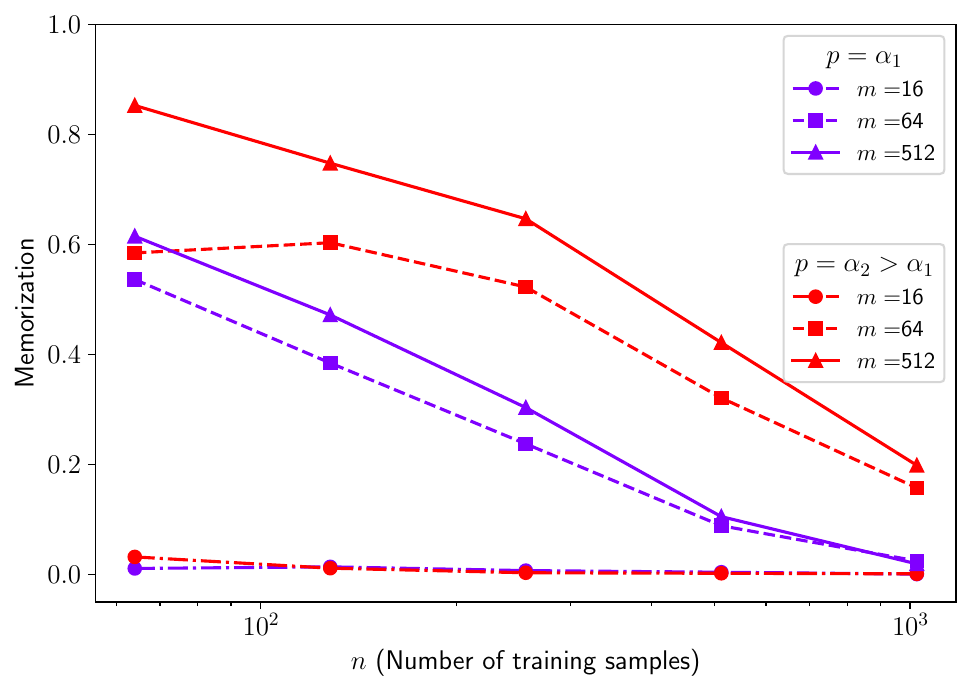}
         \caption{Results for MNIST dataset.}
    \label{appndx:mem_sim_mnist}
     \end{subfigure}
        \caption{Results of real data experiment on memorization with U-Net score function. We use $N=2000$ and $\delta=1/2$.}
        \label{appndx:mem_sim_real}

    \vspace{\intextsep}
    
    \centering
    \includegraphics[width=0.8\linewidth]{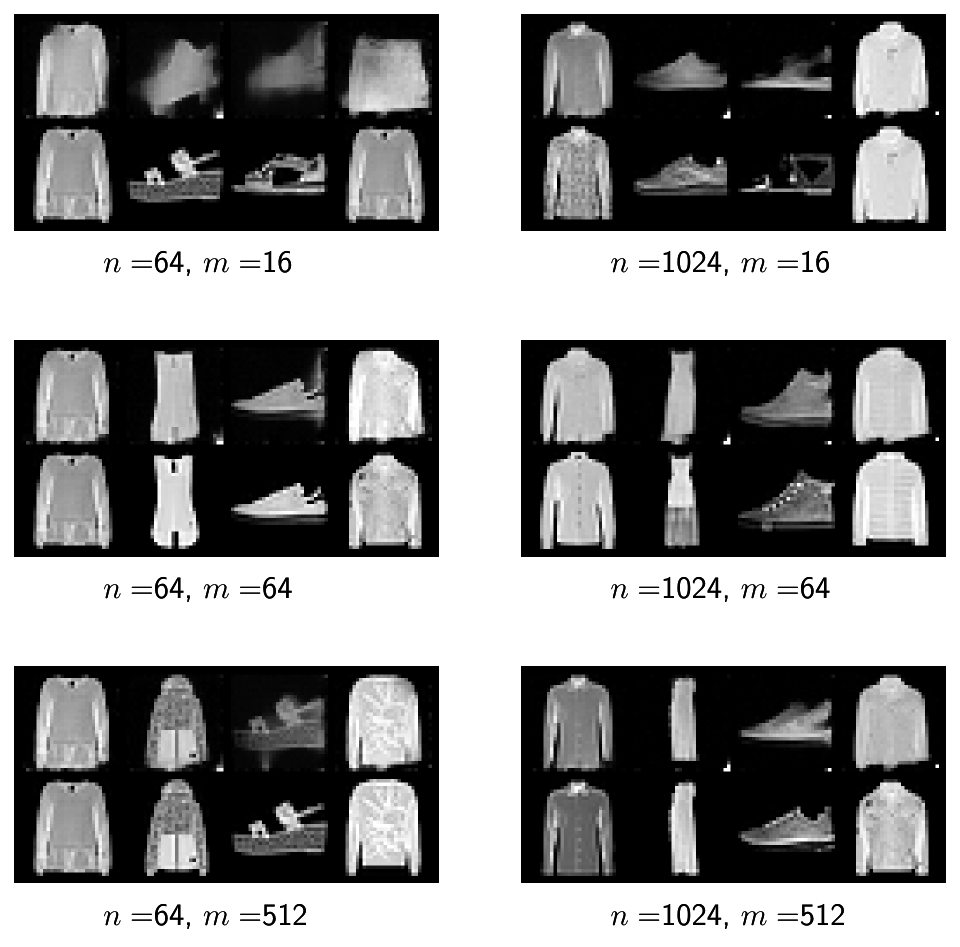}
    \caption{Images generated and their nearest neighbors in the training data set for different values of $n,m$. For each pair $n$ and $m$, we display the samples generated by diffusion model (top row) alongside their first nearest neighbors in the training dataset (bottom row).}
    \label{appndx:mem_sim_img_real}
\end{figure}

In this section, we provide details of the experiments with MNIST and Fashion-MNIST datasets and a U-Net based score function. This setting is used to obtain the results in Figures~\ref{fig:mem_sim_fmnist},~\ref{appndx:mem_sim_real} and~\ref{appndx:mem_sim_img_real}. The U-Net based neural network implementation was adapted from the one by Yang Song available \href{https://colab.research.google.com/drive/1SeXMpILhkJPjXUaesvzEhc3Ke6Zl_zxJ?usp=sharing}{here}. We used ADAM \cite{kingma_adam_2017} for optimizing the U-Net parameters. The hyperparameters used during training are: Epochs = 4000, Batch size = $\min\{64,n\}$, and Learning rate = 0.001. For a fixed $m$, we draw noise samples $\{z_1,z_2,\cdots,z_m\}$ apriori, and randomly sample from this set to do training step for one batch.

In Fig.~\ref{appndx:mem_sim_real}, we plot the measure of memorization for MNIST and Fashion-MNIST datasets as a function of $n$ for different $m$, and for two U-Net neural networks with different number of parameters ($p$). As mentioned in Section~\ref{sec:numerical_exps}, we draw the following conclusions from this experiment: 1) Memorization increases as $\frac{n}{p}$ decreases and 2) Memorization increases as $m$ increases in the overparametrized regime. Fig.~\ref{appndx:mem_sim_img_real} exhibits samples generated by the U-Net based diffusion model trained on Fashion-MNIST dataset in the above experiment. For each pair $n$ and $m$, we display samples generated by the diffusion model (top row) alongside their first nearest neighbors in the training dataset (bottom row).

We emphasize that Fig.~\ref{appndx:mem_sim_img_real} is presented as an illustration, as the phenomenology is not obvious directly from the images. 

\section{DISCUSSION ON \cite{bonnaire_why_2025}}\label{appendix:disc_biroli}
Here, we briefly discuss the related work \cite{bonnaire_why_2025}. The authors study how early stopping during training affects memorization. In their theoretical analysis, the score is also parameterized by a random features model that minimizes the loss \eqref{eqn:dsm_loss_rfm_minf} under an isotropic Gaussian data distribution. Memorization and generalization performances are defined as we do here, with an infinite number of noise samples ($m=+\infty$). Unlike our setting, which seeks the least square estimator, \cite{bonnaire_why_2025} focuses on gradient flow optimization and analyze generalization and memorization as a function of training time. Their approach is based on studying the spectrum of the same feature covariance matrix ($U$ in Lemma~\ref{lemma:lc_minf}). In particular, two spectral bulks are identified, that control two different time scales, leading to a training window where generalization occurs instead of memorization. Moreover, the width of this window grows with the number of data samples. Our present work corresponds to the regime of infinite training time and is consistent with their findings. From a technical perspective, they use the replica method to derive spectral properties of $U$ whereas we use linear pencil techniques, and a sanity check shows that both methods lead to equivalent sets of algebraic equations determining the resolvent of $U$.
%%%%%%%%%%%%%%%%%%%%%%%%%%%%%%%%%%%%%%%%%%%%%%%%%%%%%%%%%%%%%%%%%%%%%%%%%%%%%%%
%%%%%%%%%%%%%%%%%%%%%%%%%%%%%%%%%%%%%%%%%%%%%%%%%%%%%%%%%%%%%%%%%%%%%%%%%%%%%%%
\iffalse
\newpage
\input{sections/denoiser}
\fi

%\begin{figure}[ht]
%    \centering
%    \includegraphics[width=1.0\linewidth]{figs/rfm_err_sim_lam0p001_test.pdf}
%    \caption{Simulation results ($d=100$) for different values of $m$ and fixed $\psi_p=20$; with $\psi_n=2$ (upper plot) and $\psi_n=50$ (lower plot). Theoretical results for $m=1$ and $m=\infty$ are depicted as continuous lines.}
%    \label{fig:lc_sim}
%\end{figure}

\end{document}